\documentclass[11pt]{article}
\pdfoutput=1    
    
\usepackage[a4paper]{geometry}
\usepackage{amsmath}
\usepackage{amsfonts}
\usepackage{amssymb}

\usepackage[T1]{fontenc}

\usepackage{booktabs}
\usepackage{natbib,lscape}
\usepackage{macros}
\usepackage{pdfpages}
\usepackage{multirow,bigdelim}
\usepackage{algpseudocode} 
\usepackage{graphicx,xcolor,mathrsfs}
\usepackage{setspace}
\usepackage{bbm, bm}
\usepackage{url}
\usepackage[colorlinks,linkcolor=black, citecolor=blue,urlcolor=blue]{hyperref}
\usepackage{varwidth}
\usepackage{multirow}
\usepackage{zref-xr}
\zxrsetup{toltxlabel=true, tozreflabel=false}
\usepackage{cancel}
\usepackage[normalem]{ulem}
\usepackage{subfig, float}
\usepackage{rotating}
\usepackage{array}
\allowdisplaybreaks
\usepackage{fancyvrb}

\def\cX{{\mathcal {X}}}
\def\cY{{\mathcal {Y}}}

\newcommand{\PreserveBackslash}[1]{\let\temp=\\#1\let\\=\temp}
\newcolumntype{C}[1]{>{\PreserveBackslash\centering}p{#1}}

\renewcommand{\thetable}{E\arabic{table}}

\numberwithin{equation}{section}
\def\E{\mathbb{E}}

\def\cov{{\rm cov}}
\def\L{{\mathcal L}}
\newcommand{\lm}[1]{\textcolor{blue}{(RN: #1)}}

\usepackage{array,bm,bbm,url,float}
\usepackage{multirow}
\usepackage{url}
\usepackage[ruled, linesnumbered, lined, commentsnumbered]{algorithm2e}
\usepackage{zref-xr}
\usepackage{physics}
\zxrsetup{toltxlabel=true, tozreflabel=false}

\newcommand{\beq}{\begin{equation}}
\newcommand{\eeq}{\end{equation}}
\newcommand{\beas}{\begin{eqnarray*}}
\newcommand{\eeas}{\end{eqnarray*}}
\newcommand{\bea}{\begin{eqnarray}}
\newcommand{\eea}{\end{eqnarray}}
\newcommand{\bei}{\begin{itemize}}
\newcommand{\eei}{\end{itemize}}
\newcommand{\ben}{\begin{enumerate}}
\newcommand{\een}{\end{enumerate}}

\newcommand{\argmax}{\mathop{\rm arg\max}}

\newcommand{\y}{{\mathbf{y}}}

\renewcommand{\P}{{\mathbf{P}}}

\newcommand{\W}{{\mathbf{W}}}

\newtheorem{Corollary}{Corollary}

\newtheorem{Theorem}{Theorem}

\newtheorem{condition}{Condition}

\def\bbeta{{\boldsymbol{\beta}}}

\newcommand{\R}{\mathbb{R}}
\newcommand{\Pro}{{\mathbb{P}}}

\newcommand{\Var}{{\rm Var}}

\title{Differentially Private Federated Learning:\\{Servers Trustworthiness, Estimation, and Statistical Inference}}

\author{Zhe Zhang\footnote{Rutgers University. Email: \href{mailto:zzres0131@gmail.com}{zzres0131@gmail.com}.} \and Ryumei Nakada\footnote{Rutgers University. Email: \href{mailto:rn375@rutgers.edu}{rn375@stat.rutgers.edu}.} \and Linjun Zhang\footnote{Rutgers University. Email: \href{mailto:lz412@stat.rutgers.edu}{lz412@stat.rutgers.edu}.}}

\begin{document}
\maketitle

\begin{abstract}
Differentially private federated learning is crucial for maintaining privacy in distributed environments. This paper investigates the challenges of high-dimensional estimation and inference under the constraints of differential privacy. First, we study scenarios involving an untrusted central server, demonstrating the inherent difficulties of accurate estimation in high-dimensional problems. Our findings indicate that the tight minimax rates depends on the high-dimensionality of the data even with sparsity assumptions. Second, we consider a scenario with a trusted central server and introduce a novel federated estimation algorithm tailored for linear regression models. This algorithm effectively handles the slight variations among models distributed across different machines. We also propose methods for statistical inference, including coordinate-wise confidence intervals for individual parameters and strategies for simultaneous inference. Extensive simulation experiments support our theoretical advances, underscoring the efficacy and reliability of our approaches.

\end{abstract}

 
\section{Introduction}

\subsection{Overview}
Federated learning is an efficient approach for training machine learning models on distributed networks, such as smartphones and wearable devices, without moving data to a central server \citep{konevcny2016federated, kairouz2021advances, li2020federated}. Since its proposal in \citep{mcmahan2017communication}, federated learning has gained significant attention in both practical and theoretical machine learning communities. One of the key attractions of federated learning is its ability to provide a certain level of data privacy by keeping raw data on local machines. However, without specific design choices, there are no formal privacy guarantees. To fully exploit the benefits of federated learning, researchers have introduced the concept of differential privacy \citep{abadi2016deep, dwork2018privacy, dwork2006calibrating, dwork2014algorithmic, dwork2017exposed} 
to quantify the exact privacy level in federated learning. A series of research papers have focused on federated learning with differential privacy, applying various algorithms and methods \citep{hu2020personalized, truex2020ldp, wei2020federated}. Despite these efforts, there remains a significant gap between practical usage and statistical guarantees, particularly in the high-dimensional setting with sparsity assumptions, where theoretical results for the optimal rate of convergence and statistical inference results are largely missing.

In this paper, we focus on studying the estimation and inference problems in the federated learning setting under differential privacy, particularly in the high-dimensional regime. In federated learning, there are several local machines containing data sets from different sources, and a central server to coordinate all local machines to train learning models collaboratively. We present our key results in two major settings for privacy and federated learning. In the first setting, we consider an untrusted central server \citep{lowy2021private, wei2020federated, hu2020personalized} 
where each machine sends only privatized information to the central server. For example, when using smartphones, where users may not fully trust the server and do not want their personal information to be directly updated on the remote central server. In the second setting, we consider a trusted central server where each machine sends raw information without making it private. \citep{mcmahan2022federated, geyer2017differentially, mcmahan2017learning}  
For example, in different hospitals, patient data may not be shared among hospitals to protect patient privacy, but they can all report their data to a central server, such as a non-profit organization or an institute, to gain more information and publish statistics on certain diseases.

In the first part of our paper, we demonstrate that under the assumption that the central server is untrusted, the optimal rate of convergence for mean estimation is O($sd/(mn\epsilon^2)$), where $m$ is the number of local machines and each containing $n$ data points, $d$ is the parameter of interest, $s$ is the sparsity level, and $\epsilon$ is the privacy parameter. As commonly assumed in high-dimensional settings where the dimension is comparable or even larger than the number of data, such an optimality result shows the incompatibility of untrusted central server setting and high-dimensional statistics. As a result, we can only hope to get a good estimation under the trusted central server setting in the high-dimensional regime. 

In the second part of the paper, we consider the case of a trusted central server and design algorithms that allow for accurate estimations and obtain a near-optimal rate of convergence up to logarithm factors. We also present statistical inference results, including the construction of coordinate-wise confidence intervals with privacy guarantees, and the solution to conduct simultaneous inference privately. This will assist in hypothesis testing problems and construction of confidence intervals for a given subset of indices of a vector simultaneously in high-dimensional settings.
We emphasize that our algorithms for estimation and inference are suited for practical purposes, considering its capacity to (1) leverage data from multiple devices to improve machine learning models and (2) draw accurate conclusions about a population from a sample while preserving individual privacy. 
For instance, in healthcare, we could combine patient data from multiple hospitals to develop more accurate models for disease diagnosis and treatment, while ensuring that patient privacy is protected. 
We summarize our major contributions as follows:
\begin{itemize}
    \item For the untrusted central server setting, we provably show that federated learning is not suited for high-dimensional mean estimation problems by providing the optimal rate of convergence under the untrusted central server constraints. 
    This suggests us to consider a trusted central server setting to utilize federated learning for such problems.
    \item For the trusted central server setting, we design novel algorithms to achieve private estimation with federated learning.  
    We first consider the estimation in homogeneous federated learning setting and then we extend it to a more complicated heterogeneous federated learning setting. 
    We also provide a sharp rate of convergence for our algorithm in both settings.
    \item In addition, we consider statistical inference problems in both homogeneous and heterogeneous federated learning settings. We provide algorithms for coordinate-wise and simultaneous confidence intervals, which are two common inference problems in high-dimensional statistics. It is worth mentioning that our proposed methods for high-dimensional differentially private inference problems are novel and unique, which has not been developed even for the single-source and non-federated learning setting. Theoretical results show that our proposed confidence intervals are asymptotically valid, supported by simulations.
\end{itemize}
\subsection{Related Work}

In the literature, several works focused on designing private algorithms in federated learning/distributed learning based on variants of stochastic gradient decent algorithms. \citep{agarwal2018cpsgd} proposed a communication efficient algorithm, CP-SGD algorithm for learning models with local differential privacy (LDP). 
\citep{erlingsson2020encode} proposed a distributed LDP gradient descent algorithm by applying LDP on gradients with ESA framework \citep{bittau2017prochlo}. \citep{girgis2021shuffled} extended works on LDP approach for federated learning and proposed a distributed communication-efficient LDP stochastic gradient descent algorithm through shuffled model and analyzed the upper bound of the convergence rate. However, the trade-off between statistical accuracy and the privacy cost has not been considered in these works.

In the distributed settings, the trade-off between statistical accuracy and information constraints has been discussed in various papers. Two common types of information constraints are communication constraints and privacy constraints. We refer to \citep{zhang2013information, braverman2016communication, han2018geometric, barnes2020lower, 
garg2014communication} for more discussions on communication constraints, considering the situation where the bits of the information during communication have constraints. 

A series of work discusses the trade-off between accuracy and privacy in high-dimensional and non-federated learning problems, including top-$k$ selection \citep{steinke2017tight}, sparse mean estimation \citep{cai2019cost}, linear regression \citep{cai2019cost, talwar2015nearly},  generalized linear models \citep{cai2020cost}, latent variable models \citep{zhang2021high}. However, the discussion on privacy constraints in the distributed settings are still largely lacking. Among the existing works, most of them focus on the local differential privacy (LDP) constraint. In \citep{barnes2020fisher}, the mean estimation under $\ell_2$ loss for Gaussian and sparse Bernoulli distributions are discussed. \citep{duchi2019lower} discussed the lower bounds under LDP constraints in the blackboard communication model for mean estimation of product Bernoulli distributions and sparse Gaussian distributions.
\cite{acharya2020general} proposed a more general approach to combine both communication constraints and privacy constraints. 
Compared with previous works, we focus on the problem where there are $n$ data points on each machine. Our interest lies in the $(\epsilon,\delta)$-DP instead of LDP, which is a weaker constraint containing broader settings. 
We further note that, compared with the blackboard communication model \citep{braverman2016communication, garg2014communication}, in the federated learning setting, we assume that the existence of a central server and that each server is only allowed to communicate with the central server. This setting enables us to enhance more privacy.

When we are finalizing this paper\footnote{An initial draft of this paper was published as a Ph.D. dissertation in 2023~\cite{zhethesis}.}, we realized an independent and concurrent work \cite{li2024federated}. 
\cite{li2024federated} also considers differentially private federated transfer learning under high-dimensional sparse linear regression model. Namely, they proposed a notion of federated differential privacy that allows multiple rounds of $(\epsilon,\delta)$-differentially private transmissions between local machines and the central server, and provides algorithms to filter out irrelevant sources, and exploit information from relevant sources to improve the performance of estimation of target parameters.
We differntiate our research with their paper as follows: 
(1) While they consider differentially private federated learning under untrusted server setting, we deal with both trusted and untrusted server settings. We also highlight a fundamental difficulty of pure $\epsilon$-differentially private estimation under untrusted central server settings in federated learning by establishing a tight minimax lower bound, and resort to trusted server settings for estimation and inference problems.
(2) While their investigation centers on differentially private estimation within a federated transfer learning framework---specifically focusing on parameter estimation for a target distribution using similar source data---our work focuses on private estimation and \textit{inference} for parameters that are either common across all participating machines, or vary across different machines.

We also cite papers that provided us inspirations for the design of our proposed algorithms and methods. \citep{javanmard2014confidence} introduces a de-biasing produce for the statistical inference problems. \citep{li2020transfer} considers the transfer learning problem in high-dimensional settings, which enables us to combine information from other sources to benefit the estimation problems. Such idea could be adopted in the hetergeneous federated learning problems. 
For the simultaneous inference problems, we refer to \citep{zhang2017simultaneous, yu2022distributed}, which discussed how to conduct simultaneous inference for high-dimensional problems.

\textbf{Notation.} We introduce several notations used throughout the paper. Let $\bm v = (v_1, v_2,\dots,v_d)^\top \in \mathbb{R}^d$ represent a vector. Given a set of indices $\mathcal{S}$, $\bm v_{\mathcal{S}}$ refers to the components of $\bm v$ corresponding to the indices in $\mathcal{S}$. The $\ell_q$ norm of $\bm v$, for $1 \leq q \leq \infty$, is given by $\|\bm v\|_q$, whereas $\|\bm v\|_0$ represents the number of non-zero elements in $\bm v$, also called as its sparsity level.

We use $m$ to indicate the number of machines, $n$ for the number of samples per machine, $d$ for the dimensionality of vectors, and $s$ for their sparsity level. The total number of samples across all machines is denoted by $n_0 = m \cdot n$. Additionally, we define the truncation function $\Pi_T : \mathbb{R}^d \rightarrow \mathbb{R}^d$, which projects a vector onto the $\ell_{\infty}$ ball of radius $T$ centered at the origin.

For a matrix $\Sigma$, $\max_{\|\bm v\|_2 = 1, \|\bm v\|_0\le s} \bm v^\top \Sigma \bm v$ and $\min_{\|\bm v\|_2 = 1, \|\bm v\|_0\le s} \bm v^\top \Sigma \bm v$ denote the largest and smallest $s$-restricted eigenvalues of $\Sigma$, denoted as $\mu_s(\Sigma)$ and $\nu_s(\Sigma)$, respectively.

For sequences $a_n$ and $b_n$, $a_n = o(b_n)$ implies $a_n/b_n \rightarrow 0$ as $n$ grows, $a_n = O(b_n)$ signifies that $a_n$ is upper bounded by a constant multiple of $b_n$, and $a_n = \Omega(b_n)$ indicates that $a_n$ is lower bounded by a constant multiple of $b_n$, where constants are independent of $n$. The notation $a_n \asymp b_n$ denotes that $a_n$ is both upper and lower bounded by constant multiples of $b_n$.

In this work, we often use symbols $c_0, c_1, m_0, m_1, C, C', K, K'$ to represent universal constants. Their specific values may vary depending on the context, but they are independent from other tunable parameters.

\section{Preliminaries}

\subsection{Differential Privacy}
We start form the basic concepts and properties of differential privacy \citep{dwork2006calibrating}. 
The intuition behind differential privacy is that a randomized algorithm produces similar outputs even when an individual's information in the dataset is changed or removed, thereby preserving the privacy of individual data.
The formal definition of differential privacy is given below.
\begin{definition}[Differential Privacy \citep{dwork2006calibrating}]
Let $\mathcal{X}$ be the sample space for an individual data, a randomized algorithm $M:\mathcal{X}^n\rightarrow\mathbb{R}$ is $(\epsilon, \delta)$-differentially private if and only if for every pair of adjacent data sets $ \bm X,  \bm X' \in \mathcal X^n$ and for any $S \subseteq  \R$, the inequality below holds:
\begin{align*}
	\mathbb{P}\left(M(\bm X) \in S\right) \leq e^\varepsilon \cdot \mathbb{P}\left(M( \bm X') \in S\right) + \delta,
\end{align*}
where we say that two data sets $\bm X = \{\bm x_i\}_{i=1}^n$ and $\bm X' = \{{{\bm x}_{i}'}\}_{i=1}^n$ are adjacent if and only if they differ by one individual datum.
\end{definition}
In the above definition, the two parameters $\epsilon,\delta$ control the privacy level. From the definition, with smaller $\epsilon$ and $\delta$, the outcomes given adjacent $\bm X$ and $\bm X'$ become closer, making it harder for an adversary to distinguish if the original dataset is $\bm X$ or $\bm X'$, indicating the privacy constraint becomes more stringent. Furthermore, when $\delta = 0$, we could use $\epsilon$-differentially private as the abbreviation of $(\epsilon,0)$-differentially private.  


In the rest of this section, we introduce several useful properties of differential privacy and how to create a differential private algorithm from non-private counterparts. One common strategy is through noise injection. The scale of noise is characterized by the sensitivity of the algorithm: 
\begin{definition}
 For any algorithm $f : \mathcal{X}^n \rightarrow {\R}^d$ and two adjacent data sets $\bm X$ and $\bm X'$, the $\ell_p$-sensitivity of $f$ is defined as:
	$$\Delta_{p}(f) = \sup_{\bm X,  \bm X'\in\mathcal{X}^n \text{ adjacent}}\|f(\bm X) - f(\bm X')\|_p.$$
\end{definition}
We then introduce two mechanisms. For algorithms with finite $\ell_1$-sensitivity, we add Laplace noises to achieve differential privacy, while for $\ell_2$-sensitivity, we inject Gaussian noises.

\begin{proposition}[The Laplace Mechanism \citep{dwork2006calibrating,dwork2014algorithmic}]
    Let $f: \mathcal X^n \to \R^d$ be a deterministic algorithm with $\Delta_1(f)< \infty$. For $\bm w \in \R^d$ with coordinates $w_1, w_2, \cdots, w_d$ be i.i.d samples drawn from Laplace$(\Delta_1(f)/\epsilon)$, $f(\bm X) +\bm  w$ is $(\epsilon, 0)$-differentially private. 
\end{proposition} 
\begin{proposition}[The Gaussian Mechanism \citep{dwork2006calibrating,dwork2014algorithmic}]
    Let $f: \mathcal X^n \to \R^d$ be a deterministic algorithm with $\Delta_2(f)< \infty$. For $ \bm w \in \R^d$ with coordinates $w_1, w_2, \cdots, w_d$ be i.i.d samples drawn from $ N(0, 2(\Delta_2(f)/\epsilon)^2\log(1.25/\delta))$, $f(\bm X) + \bm w$ is $(\epsilon, \delta)$-differentially private. 
\end{proposition}
The post-processing and composition properties are two key properties in differential privacy, which enable us to design complicated differentially private algorithms by combining simpler ones. Such properties are pivotal in the design of algorithms in later chapters. 
\begin{proposition}[Post-processing Property  \citep{dwork2006calibrating}]
    Let $M$ be an $(\epsilon, \delta)$-differentially private algorithm and $g$ be an arbitrary function which takes $M(\bm X)$ as input, then $g(M(\bm X))$ is also $(\epsilon, \delta)$-differentially private.
\end{proposition}
\begin{proposition}[Composition property \citep{dwork2006calibrating}]\label{composition} 
    For $i = 1, 2$, let $M_i$ be $(\varepsilon_i, \delta_i)$-differentially private algorithm, then $(M_1, M_2)$ is $(\epsilon_1 + \epsilon_2, \delta_1 + \delta_2)$-differentially private algorithm. 
\end{proposition}
We also mention NoisyHT algorithm (Algorithm~\ref{algo:peeling}) introduced by \cite{dwork2018differentially}, which stands for the noisy hard-thresholding algorithm. The algorithm aims to pursue both sparsity of the output and privacy at the same time.

\begin{algorithm}[H]\label{algo:peeling}
\caption{Noisy Hard Thresholding Algorithm (NoisyHT($\bm v, s, \lambda, \epsilon,\delta$)) \citep{dwork2018differentially}}

\textbf{Input: } vector-valued function $ \bm v =  \bm v(\bm X) \in \R^d$ with data $\bm X$, sparsity $s$, privacy parameters $\varepsilon, \delta$, sensitivity $\lambda$. \\
\textbf{Initialization:} $S = \emptyset$.\\
\textbf{For} $i$ in $1$ \KwTo $s$:\\
\quad Generate $\bm w_i \in \R^d$ with $w_{i1}, w_{i2}, \cdots, w_{id} \stackrel{\text{i.i.d.}}{\sim} \text{Laplace}\left(\lambda \cdot \frac{2\sqrt{3s\log(1/\delta)}}{\varepsilon}\right)$.\\
\quad Append $j^* = \text{argmax}_{j \in [d] \setminus S} (|v_j| + w_{ij})$ to $S$.\\
\textbf{End For}\\
\quad Generate $\tilde {\bm w}$ with $\tilde w_{1}, \cdots, \tilde w_{d} \stackrel{\text{i.i.d.}}{\sim} \text{Laplace}\left(\lambda \cdot \frac{2\sqrt{3s\log(1/\delta)}}{\varepsilon}\right)$.\\
\textbf{Output: } $P_S(\bm v+\bm \tilde{w})$. 

\end{algorithm}

In the last step, $P_S(\bm u)$ denotes the operator that makes $\bm u_{S^c}=0$ while preserving $\bm u_S$. This algorithm could be seen as a private top-k selection algorithm, which helps build our proposed algorithm in later section. 

Specifically, when the sparsity $s$ is chosen to be $1$, the algorithm outputs the maximum element chosen after a single iteration in the private manner. We refer this special case as the Private Max algorithm,
which is implemented in Algorithm~\ref{algo: bootstrap_simu} used for simultaneous inference.

\subsection{Federated Learning} 
\begin{figure}[ht]
\centering
\includegraphics[width= 11 cm]{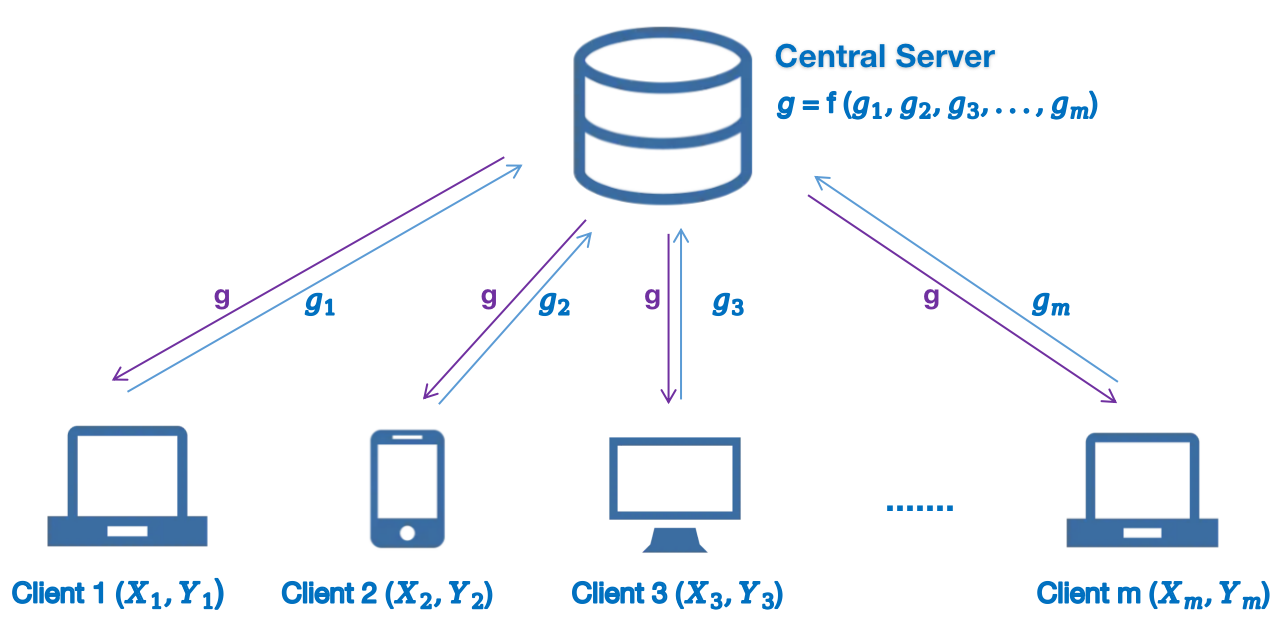}
\caption{Federated Learning 
}\label{fig:federated}
\end{figure}

Federated learning introduced in \citep{mcmahan2017communication} is a technique designed to train a machine learning algorithm across multiple devices, without exchanging data samples. A central server coordinates the process, with each local machine sending model updates to be aggregated centrally. Figure~\ref{fig:federated} illustrates the basic concept of federated learning

One characteristic of federated learning is that the training of machine learning models occurs locally, and only parameters and updates are transferred to the central server and shared by each node. Specifically, communication between local machines and the server is bidirectional: machines send updates to the central server, and in return, they receive aggregated information after processing. Communication among local machines is prohibited to prevent privacy leakage. Intuitively, federated learning inherently provides a certain level of privacy.

Although without rigorous definitions, there are two main branches of central server settings in federated learning: the untrusted central server setting and the trusted central server setting \citep{lowy2021private, wei2020federated, mcmahan2022federated, geyer2017differentially}. In the first setting, where the central server is untrusted, each piece of information sent from the machine to the central server should be differentially private. In the second setting, we assume a trusted central server exists. In this scenario, it is safe to send raw information from the machine to the central server without additional privacy measures. However, to prevent information leakage among local machines, the information sent back from the server should also be differentially private.

Another key aspect of federated learning is that the datasets on each local machine are commonly not independent and identically distributed (i.i.d.). This allows federated learning to train on heterogeneous datasets, aligning with practical scenarios where the datasets on different machines are typically diverse and their sizes may also vary. We will demonstrate that federated learning can efficiently estimate the local model when models on different local machines differ but share some similarities, a concept we refer to as heterogeneous federated learning in Section~\ref{section5}.

\subsection{Problem Formulation}\label{sec: formulation}

In this paper, we assume that there exists a central server and $m$ local machines. We denote the data on these machines by $\bm X_1, \bm X_2, \bm X_3,\dots, \bm X_m$, respectively, with $\bm X_i \in \mathbb{R}^{n_i \times d}$. On any machine $i =1,2,\dots,m$, there are $n_i$ data points $\bm X_i = [\bm X_{i1}, \bm X_{i2},\dots, \bm X_{in_i}]$. For simplicity, we assume that there are equal data points $n = n_1 = n_2 = \dots = n_i$ for each machine. We note that the result could be easily generalized to cases where the sample sizes on each machine differ. 

We consider both untrusted and trusted central server settings. 
For the untrusted setting, we require that the information sent from local machines to the server is private. 
In this scenario, we show that in the high-dimensional setting, even with sparsity assumptions, it is impossible to achieve small estimation error when the central server is untrusted. 
In the trusted setting, we consider the high-dimensional linear regression problem $\bm Y = \bm X \bbeta + \bm W$ with $s$-sparse $\bbeta$. We will first study the case where all machines share the same $\bbeta$, (referred to homogeneous federation learning,) and then study a more general case where models on different machines are not equal, but share certain similarities (referred to heterogeneous federation learning.) We show that our algorithm can adapt to such similarity---with larger similarity, the algorithm achieves a faster rate of convergence. 

\section{An Impossibility Result in the Untrusted Central Server setting}
In this section, we study the untrusted server setting where the local machines need to send privatized information to the central server to ensure privacy. We show an impossibility result that in high-dimensional settings where the data dimension is comparable to or greater than the sample size, accurate estimation is not feasible even if we consider a simple sparse mean estimation problem. 

As mentioned in Section \ref{sec: formulation}, we consider a federated learning setting with $m$ machines, where each machine $i\in[m]$ handles $n$ data points $\bm X_i:=[\bm X_{i1}, \bm X_{i2},\dots, \bm X_{in}] \in \mathbb{R}^{n \times d}$. Let $D_{\textnormal{all}}=\{\bm X_i\}_{i=1}^m$. We assume that each data point $\bm X_{ij} \in \mathbb{R}^d$ follows a Gaussian distribution $N(\bm \mu,\bm I_d)$, where $\bm \mu$ is a sparse $d$-dimensional vector with sparsity $s$. The goal is to estimate $\bm\mu$ in the federated learning setting when the central server is untrusted. 
In this section, we provide an optimal rate of convergence for this problem and show that the untrusted central server setting is not suited for high-dimensional problems. 

We begin by deriving the minimax lower bound, which characterizes the fundamental difficulty of this estimation problem. In untrusted server setting, we additionally assume that each piece of information sent from the local machine to the central server follows $\epsilon$-differential privacy.  
To achieve this, we introduce the privacy channel $\mathcal{W}^{\epsilon\textnormal{-priv}}: \mathcal{X}^n \to \mathcal{Z} $, a function that is responsible for privatizing the information transmitted from the local machines. Given the input $X \in \mathcal{X}^n$ and the privacy channel $\mathcal{W}^{\epsilon\textnormal{-priv}}$, $Z \in \mathcal{Z}$ representing all the information (from multiple rounds) transmitted to the central server. More precisely, we require privacy guarantees such that for any two adjacent datasets $X$ and $X' \in \mathcal{X}^n$, differing by only one data point on any local machine, and for an output $Z \in \mathcal{Z}$ representing the information sent from the local machine to the central server, differential privacy guarantee $\Pro(\mathcal{W}^{\epsilon\textnormal{-priv}}(X) = Z) \le e^{\epsilon} \cdot \Pro(\mathcal{W}^{\epsilon\textnormal{-priv}}(X') = Z)$ holds.

We consider any mechanism $M$ in the federated learning setting with $m$ local machines and one central server, operated on the dataset $D_{\textnormal{all}}$. ${M}$ serves as a procedure to estimate $\bm \mu$, where each local machine collaborates exclusively with the central server without direct interaction among themselves. On each machine $i$, the mechanism ${M}$ uses the privacy channel $\mathcal{W}_i^{\epsilon\textnormal{-priv}}$ and data sample $\bm X_i$ to generate $\bm Z_i$, which is then transmitted to the central server. The central server receives the information from all machines. After multi-rounds of collaboration between local machines and the central server, we obtain the sparse and private estimator $\hat{\bm \mu} \in \mathbb{R}^d$. We denote the class of all mechanisms that satisfy the above constraints as $\mathcal M^{\textnormal{untrust}}_{m,\epsilon}(D_{\textnormal{all}})$. 
Under this setting we establish a lower bound for the estimation error of the mean in Theorem~\ref{thm1}. 

\begin{Theorem}\label{thm1} 
Suppose $D_{\textnormal{all}}$ is generated as above. Let $\mu$ be a $s$-sparse $d$-dimensional mean of Gaussian distribution satisfying $\|\bm \mu\|_{\infty} \le 1$. We consider the estimation of the mean vector $\hat{\bm \mu}$ under the untrusted central server federated learning setting with $m$ local machines and  $n$ data points in each machine. 
Then, there exists a constant $c > 0$ such that
\begin{equation*}
    \inf_{M\in\mathcal M^{\textnormal{untrust}}_{m,\epsilon}} \sup_{\bm \mu \in \mathbb{R}^d, \|\bm \mu\|_{\infty} \le 1}\| \bm \mu-M(D_{\textnormal{all}})\|_2^2 \ge c \cdot \min\qty(\frac{s}{n}, \frac{sd}{mn\epsilon^2}).
\end{equation*}
\end{Theorem} 
The lower bound contains two terms. The first term, of order $s/n$, represents the minimax risk of mean estimation using only the samples from a local machine. The second term, of order $sd/(mn\epsilon^2)$, accounts for the error from federated learning across multiple machines under privacy constraints. Theorem~\ref{thm1} suggests that we cannot perform better than either choosing to estimate the mean using only the local machine or adopting the federated learning approach and combining information from different machines. However, in the latter approach, we must at least incur a rate of $O(d/(mn\epsilon^2))$, which is linearly proportional to the dimension $d$. This result suggests that privacy constraints significantly impact the efficiency of federated learning in high-dimensional settings. Furthermore, as the number of machines $m$ increases, we can possibly attain better performance, highlighting the merit of federated learning.


We also show the tightness of the lower bound in Theorem~\ref{thm1} by providing the upper bound. 
\begin{Theorem}\label{thm2}
    Suppose that conditions in Theorem 1 hold.
    Then, there exists an $\epsilon$-differentially private algorithm for the estimation of $\bm \mu$ as $\hat{\bm \mu}$ 
    \begin{equation*}
        \|\bm \mu-\hat{\bm \mu}\|_2^2 \le c \cdot \frac{s \cdot d}{mn\epsilon^2},
    \end{equation*}
    where $c > 0$ is some constant. 
\end{Theorem}

The proof follows by constructing an algorithm that transforms Gaussian mean to Bernoulli mean according to the sign of the Gaussian mean, motivated by Algorithm~2 discussed in \citep{acharya2020general}, where the authors discuss $l$-bit protocol for estimating the product of Bernoulli family. More details of the algorithm are deferred to Section~\ref{Proof:2}. 
Based on the results from Theorems~\ref{thm1} and~\ref{thm2}, we obtain the optimal rate of convergence for sparse mean estimation under differentially private federated learning setting. As a result, when the central server is untrusted, it is impossible to find an approach to achieve accurate estimation under the untrusted server assumption. This highlights the necessity of the trusted server setting for statistical estimation and inference in high-dimensional federated learning scenarios. In the following sections, we develop estimation and inference procedures under the trusted server settings.


\section{Homogeneous Federated Learning Setting}\label{sec:homo}
\subsection{Algorithms for Estimation Problems}
In this section, we consider the setting of a trusted central server, where local machines fully trust the central server and send unprivatized information to it without implementing privacy measures. However, when the central server sends information back to the local machines, it must ensure that this information is privatized to avoid any privacy leakage across local machines.

In this subsection, we first focus on the statistical estimation problems in this setting and then develop inference results in the next subsection. More specifically, our primary focus is on the linear regression problem in a high-dimensional setting, where the ground truth, denoted as $\bbeta$, is a sparse $d$-dimensional vector. We initially study the simpler case in this section, where the underlying generative models for each local machine are identical, which we refer to as the homogeneous federated learning setting. A more complicated heterogeneous setting will be discussed in the following section. Specifically, we consider the following high-dimensional linear regression model:
$$\bm Y = \bm X \bm \beta + \bm W,$$
where we assume $\bm W$ is the error term whose coordinates are independent and following sub-Gaussian distribution with variance proxy $\sigma^2$, denoted by $W_i \sim \text{subG}(\sigma^2)$. 
$\bm X$ is a random matrix whose rows are following sub-Gaussian distribution with a covariance matrix $\bm \Sigma$. 

We first introduce the parameter estimation algorithm under differentially private federated settings with a trusted central server. 

\begin{algorithm}[H]\label{algo: privateregression}
	\SetAlgoLined
	\SetKwInOut{Input}{Input}
	\SetKwInOut{Output}{Output}
	\SetKwFunction{Peeling}{Peeling}
	\Input{Dataset $D_{\textnormal{all}}=\{(\bm X_i, \bm Y_i)\}_{i \in[m]}$, number of machines $m$, number of samples on each machine $n$, step size $\eta^0$, privacy parameters $\varepsilon, \delta$, noise scale $B_0$, number of iterations $T$, truncation level $R$, feasibility parameter $C_0$, sparsity $s$, initial value $\bbeta^0$.}
	\For{$t$ \rm{\textbf{from}} $0$ \KwTo $T-1$}{
	\textbf{Step 1:} 
	    
	    \quad On each local machine $i = 1,2,\dots,m$, calculate the local gradient $$\bm g_i =  \frac{1}{n} \sum_{j=1}^n (\bm X_{ij}^\top \bbeta^t-\Pi_{R}(y_{ij}))\bm X_{ij}.$$ Send the gradient $ \bm g_i$ to the central server.\\ 
	\textbf{Step 2:}

		\quad Compute $\bbeta^{t + 0.5} = \bbeta^t - (\eta^0/m)\sum_{i=1}^{m} \bm g_i$ at the central server\;
		\quad Compute $\bbeta^{t+1} = \Pi_{C_0}\left(\text{NoisyHT}(\bbeta^{t+0.5}, s, \frac{\varepsilon}{T}, \frac{\delta}{T}, \frac{\eta^0 B_0}{mn})\right)$ at the central server. 
	
	\textbf{Step 3: }Send the output $\bbeta^{t+1}$ back to each local machine from the server. 
	}
	
	\textbf{Output: } Return $\bbeta^T$.
	
	\caption{Differentially Private Sparse Linear Regression under Federated Learning}
\end{algorithm}

In Step 1 of Algorithm~\ref{algo: privateregression}, the information computed on each local machine is transmitted to the central server. The second step involves calculations performed at the central server. Prior to sending the information back to the local machines, it undergoes privacy preservation through the application of the NoisyHT algorithm, as introduced in Algorithm~\ref{algo:peeling}. Subsequently, the local machine updates its estimation based on the information received from the central server.

We compare Algorithm~\ref{algo: privateregression} with Algorithm 4.2 
in \citep{cai2019cost}, which addresses the private estimation in linear regression under non-federated learning settings. Unlike the latter, 
our algorithm does not transmit all data points to the central server. Instead, we calculate the gradient updates locally on each machine and send only these local gradients to the server. This design enhances privacy protection, as the original data remains visible only on the local machine and is not exposed externally. Furthermore, this approach of gradient updates also reduces communication costs by transmitting only a $d$-dimensional vector from each local machine for the gradient update. 
Previous research has also considered non-private distributed methods for linear regression problems, such as \citep{lee2017communication, zhang2012communication}. Our algorithm, however, ensures differential privacy.
In practice, the sparsity level $s$ can be determined using a private version of cross-validation, while other parameters may be pre-chosen based on our theoretical analysis.

\subsection{Algorithms for Inference Problems}

In this subsection, we focus on statistical inference problems in the homogeneous federated learning setting, such as constructing coordinate-wise confidence intervals for parameters and performing simultaneous inference. To begin, we develop a method for constructing coordinate-wise confidence intervals, for example, for the $k$-th index of $\bbeta$, $\beta_k$. However, it is important to note that the output of Algorithm~\ref{algo: privateregression} is biased due to hard thresholding. To overcome this bias, we employ a de-biasing procedure, a common technique in high-dimensional statistics, as demonstrated in previous studies \citep{javanmard2014confidence}. This procedure involves approximating the $k$-th column of the precision matrix $ \bm \Theta = \bm \Sigma^{-1}$ to construct confidence intervals for each $\beta_k$. 
Subsequently, we focus on obtaining an estimate of the precision matrix in a private manner.

\begin{algorithm}[H]\label{algo:privateprecision}
	\SetAlgoLined
	\SetKwInOut{Input}{Input}
	\SetKwInOut{Output}{Output}
	\SetKwFunction{Peeling}{Peeling}
	\Input {Number of machines $m$, number of data points in each machine $n$, dataset $\bm X_i = (\bm x_{i1}, \dots, \bm x_{in})$ for $i = 1,\dots,m $, step size $\eta^1$, privacy parameters $\varepsilon, \delta$, noise scale $B_1$, number of iterations $T$, feasibility parameter $C_1$, sparsity $s$, initial value $\bm \Theta_k^0$.}
	\For{$t$ \rm{\textbf{from}} $0$ \KwTo $T-1$}{
	\textbf{Step 1: }  On each local machine $i = 1,2,\dots,m$, calculate local gradient $\bm g_i =  \frac{1}{n} \sum_{j=1}^n \bm X_{ij} \bm X_{ij}^\top \bm \Theta_k^t -\bm e_j$. Send the gradients $(\bm g_{1}, \bm g_2,\dots,\bm g_m)$ to the central server. \\
	\textbf{Step 2:}

		\quad Compute $\bm \Theta_k^{t + 0.5} = \bm \Theta_k^t - (\eta^0/m)\sum_{i=1}^{m} \bm g_i$ at the central server\;
		\quad Compute $\bm \Theta_k^{t+1} = \Pi_{C_1}\left(\text{NoisyHT}(\bm w_k^{t+0.5}, s, \frac{\varepsilon}{T}, \frac{\delta}{T}, \frac{\eta^1 B_1}{mn})\right)$ at the central server.
	
	\textbf{Step 3: }Send $\bm \Theta_k^{t+1}$ back to each local machine from the server. 
	}
	
	\textbf{Output: } Return $\bm \Theta_k^T$.
	
	\caption{Differentially Private Precision Matrix Estimation in Federated Learning}
\end{algorithm}

The structure of Algorithm~\ref{algo:privateprecision} is similar to Algorithm~\ref{algo: privateregression}, as both adopt an iterative communication between the central server and the local machines; the information is initially transmitted from the local machines to the server, then, the central server performs calculations and use the NoisyHT algorithm (Algorithm~\ref{algo:peeling}) to ensure the privacy of the information. Subsequently, each local machine updates the gradient and progresses to the next iteration. The primary distinction between two algorithms lies in the computation of the gradient on each machine.


Denote the output of Algorithm~\ref{algo: privateregression} as $\hat{\bm \beta}$ and the output of Algorithm~\ref{algo:privateprecision} as $\hat{\bm \Theta}_k$. Then the de-biased differentially private estimator of $\beta_k$ is given by 
\begin{align}\label{eq: debiase-estimator}
    \hat{\beta}^{\text{u}}_k = \hat{ \beta}_k + \frac{1}{m} \sum_{i=1}^m  \hat{ \bm \Theta}_k^\top \bm g_i + E_k, 
\end{align}
where     $\bm g_i = (1/n) \sum_{j=1}^n (\bm X_{ij}^\top \hat{\bm \beta}-\Pi_{R}(y_{ij}))\bm X_{ij}$, and
$E_k$ is the injected random noise to ensure privacy, following a Gaussian distribution $N(0,8 \Delta_1^2 \log(1.25/\delta)/(n^2 m^2 \epsilon^2) )$, where $\Delta_1 =  \sqrt{s} c_1 c_x R + s c_0 c_1 c_x^2$ with some constants $c_0, c_1, c_x$ defined later.

The debiased estimator in \eqref{eq: debiase-estimator} enables us to construct a differentially private confidence intervals. Although the variance $\sigma$ of the error term $\bm W$ in the linear regression model is usually unknown, we can estimate $\sigma$ from the data in a private manner. The estimation is based on the residual term between the response $\bm Y$ and the fitted value $\bm X \hat{\bbeta}$. 
We summarize the method to estimate $\sigma$ in the private federated learning setting in Algorithm~\ref{algo:privatevar}.

\begin{algorithm}[H]\label{algo:privatevar}
	\SetAlgoLined
	\SetKwInOut{Input}{Input}
	\SetKwInOut{Output}{Output}
	\SetKwFunction{Peeling}{Peeling}
	\Input {Dataset $(\bm X_i, \bm Y_i)_{[i=1,2,\dots,m]}$, privacy parameters $\varepsilon$, noise scale $B_{2}$, truncation level $R$, estimated parameter $\hat{\bbeta}$ from Algorithm~\ref{algo: privateregression}.}
        \textbf{Step 1: } On each machine $i = 1,2,\dots,m$, compute $\hat{W}_i =  \|\Pi_R(\bm Y_i) -\bm X_i \hat{\bbeta}\|_2^2/n$ and send $\hat W_i$ to the central server. \\ 
        \textbf{Step 2: } Generate a random variable $E^{\textnormal{var}}$, where $E^{\textnormal{var}} \sim$ $N (0, 2B_{2}^2 \log(1.25/\delta)/\epsilon^2)$\\
	    \textbf{Step 3: } Compute $\hat{\sigma}^2$ such that $\hat{\sigma}^2 = \sum_{i = 1}^m \hat{W}_i/m + E^{\textnormal{var}} $ at the central server\\
	\Output{ Estimated variance $\hat{\sigma}^2$.}
	\caption{Differentially Private Variance Estimation in Federated Learning}
\end{algorithm} 

When examining the convergence rates of $\hat{\bbeta}$ and $\hat{\bm \Theta}_k$ in Theorem~\ref{thm: federated_estimation}, we observe the crucial roles of the largest and smallest restricted eigenvalues of $\boldsymbol{\Sigma}$. Since these eigenvalues directly influence the construction of confidence intervals and cannot be directly obtained from the data, their private estimation becomes essential. Below, we outline an algorithm to estimate the largest restricted eigenvalue, $\mu_s(\Sigma)$. To estimate the smallest restricted eigenvalue, $\nu_s(\Sigma)$, the same algorithm can be used by modifying Step 4 from ``argmax'' to ``argmin''.


\begin{algorithm}[H]\label{algo:privateeigen}
	\SetAlgoLined
	\SetKwInOut{Input}{Input}
	\SetKwInOut{Output}{Output}
	\SetKwFunction{Peeling}{Peeling}
	\Input {Number of machines $m$, dataset $(\bm X_i)_{[i=1,\dots,m]}$, number of data points in each machine $n$, privacy parameters $\varepsilon$, noise scale $B_3$, number of vectors $n_1$.}
        \textbf{Step 1: } Sample $n_1$ $d$-dimensional, $s$-sparse unit vectors $\bm v_1, \bm v_2, \dots, \bm v_{n_1}$. \\
        \textbf{Step 2: } On each machine, compute $t_{i,k} = (\bm v_k^T \bm X_i^T \bm X_i \bm v_k)/n$ where $k = 1,2\dots,n_1$ and send them on to the central server. 
    
	   \textbf{Step 3: } Sample $\xi_1,\dots,\xi_{n_1} \sim$ Laplace $(2 B_3/ \epsilon)$. \\
	   \textbf{Step 4: } Compute $k_{\max}$ such that $k_{\max} = \argmax_{k} \sum_{i=1}^m t_{i,k}/m  + \xi_{k} $.
	
	\Output{ $\mu_{s}(\hat{\bm \Sigma}) = \sum_{i=1}^m t_{i,k_{\max}}/m  + \xi $,  where $\xi \sim \operatorname{Laplace}(2B_3/\epsilon)$ independently.}
	\caption{Differentially Private Restricted Eigenvalue Estimation in Federated Learning}
\end{algorithm} 

Based on Algorithms \ref{algo:privatevar} and \ref{algo:privateeigen}, we provide a constuction for coordinate-wise confidence intervals in Algorithm~\ref{algo:coordinate-wise inference}. 

\begin{algorithm}[H]\label{algo:coordinate-wise inference}
	\SetAlgoLined
	\SetKwInOut{Input}{Input}
	\SetKwInOut{Output}{Output}
	\SetKwFunction{Peeling}{Peeling}
	\Input {Number of machines $m$, dataset $(\bm X_i, \bm Y_i)_{[i=1,\dots,m]}$, number of data points in each machine $n$, privacy parameters $\varepsilon, \delta$, truncation level $R$, sparsity $s$, estimators of parameters $\hat{\bbeta}$, $\hat{\bm \Theta}_k$ from Algorithms~\ref{algo: privateregression} and \ref{algo:privateprecision}, constants $\Delta_1, \gamma$. }
 {
	\textbf{Step 1:} 
	    \quad On each local machine $i = 1,2,\dots,m$, calculate local gradient $ \bm g_i =  \frac{1}{n} \sum_{j=1}^n (\bm X_{ij}^\top \hat{\bm \beta}-\Pi_{R}(y_{ij}))\bm X_{ij}$. Send the gradient $(\bm g_{1}, \bm g_2,\dots,\bm g_m)$ to the central server. \\
	\textbf{Step 2:}
            \quad Generate a random variable $E$ from the Gaussian distribution $N(0,8 \Delta_1^2\log(1.25/\delta)/n^2 m^2 \epsilon^2 )$. \\ 
        \textbf{Step 3:}
            \quad Calculate de-biased estimation, $ \hat{\beta}^{\text{u}}_k = \hat{ \beta}_k + \frac{1}{m} \sum_{i=1}^m  \hat{ \bm \Theta}_k^\top \bm g_i + E$\\
            
        \textbf{Step 4:}  Estimate $\hat{\sigma}$ from Algorithm~\ref{algo:privatevar} and $\hat{\mu}_s, \hat{\nu}_s$ from Algorithm~\ref{algo:privateeigen}. \\
	
	\textbf{Step 5:} \quad Calculate the confidence interval $J_k(\alpha)$. \begin{align*}
	    J_k(\alpha) = \biggl[\hat{\beta}_k^{\text{u}} -  \frac{\gamma \hat{\mu}_s^2}{\hat{\nu}_s^2} \frac{s^2 \log^2 d \log (1/\delta)\log^3 mn}{m^2 n^{2} \epsilon^2}- \Phi^{-1}(1-\alpha/2) \frac{\sigma}{\sqrt{mn}}\sqrt{\hat{\bm \Theta}_k^\top \hat{\bm \Sigma} \hat{\bm \Theta}_k+ \frac{8\Delta_1^2 \log(1/\delta)}{mn\epsilon^2}} , \notag\\\hat{\beta}_k^{\text{u}} +   \frac{\gamma \hat{\mu}_s^2}{\hat{\nu}_s^2} \frac{s^2 \log^2 d \log (1/\delta)\log^3 mn}{m^2n^{2} \epsilon^2}+\Phi^{-1}(1-\alpha/2) \frac{\sigma}{\sqrt{mn}}\sqrt{\hat{\bm \Theta}_k^\top \hat{\bm \Sigma} \hat{\bm \Theta}_k+ \frac{8\Delta_1^2 \log(1/\delta)}{mn\epsilon^2}} \biggr] 
	\end{align*}
	}
	\textbf{Output: } Return the final result $J_k(\alpha)$.
	
	\caption{Differentially Private Coordinate-wise Confidence interval for $\beta_k$ in Federated Learning}
\end{algorithm}
So far we focused on constructing confidence intervals for individual coordinates of the parameter vector $\boldsymbol{\beta}$. However, in high-dimensional settings, we are often interested in group inference problem, where we test hypotheses involving multiple coordinates simultaneously. Specifically, we consider the problem of testing the null hypothesis given by
\[H_0: \hat{\beta}_k = \beta_k \text{, for all } k \in G \] 
against the alternative hypothesis,
\[H_1: \hat{\beta}_k \neq \beta_k, \] 
for at least one $k \in G$, where $G$ is a subset of all coordinates $\{1, 2, \dots, d\}$ and we allow $|G|$ to be the same order as $d$. Additionally, we also construct simultaneous confidence intervals for all coordinates in $G$.
Note that the problem discussed above are common in high-dimensional data analysis, with applications such as multi-factor analysis of variance \citep{hothorn2008simultaneous}, additive modeling \citep{wiesenfarth2012direct}.
Previous research works have discussed similar problems in the non-private setting, including \citep{chernozhukov2013gaussian, zhang2017simultaneous, yu2022distributed}. 

To address the problem, simultaneous inference can be conducted using a test statistic
\[\max_{k \in G} |\hat{\beta}_k^{\text{u}}-\beta_k|.\]

Major challenges of simultaneous inference in a private federated learning setting include: (1) minimizing the communication cost from local machines to the server while retaining all data on the local machines, and (2) ensuring the privacy of the procedure, which necessitates a tailored privacy-preserving mechanism at each step of the algorithm.

In our framework, we propose an algorithm based on the bootstrap method. As previously mentioned, to build confidence intervals, our interest lies in the statistic computed by the maximum coordinate of $\hat{\beta}_k^{\text{u}} - \beta_k$ over $G$. By decomposing this statistic, we obtain a term $\frac{\sigma}{\sqrt{mn}}  \sum_{i=1}^m \sum_{j=1}^n \hat{\bm \Theta} \bm X_{ij} (\y_{ij} - \bm X_{ij}^T \bbeta )$. To determine the distribution of this term, we bootstrap the residuals $\y_{ij} - \bm X_{ij}^T \bbeta$. 

We outline the algorithm as follows: we first estimate $\hat{\bbeta}$ and $\hat{\bm \Theta}_k$ using Algorithm~\ref{algo: privateregression} and \ref{algo:privateprecision}, respectively. Accordingly, by stacking $\hat{\bm \Theta}_k$ for all $k$, we get an estimator of the precision matrix $\bm \hat{\bm \Theta}$. 
The details are provided in Algorithm~\ref{algo: bootstrap_simu}.


\begin{algorithm}[H]\label{algo: bootstrap_simu}
	\SetAlgoLined
	\SetKwInOut{Input}{Input}
	\SetKwInOut{Output}{Output}
	\SetKwFunction{Peeling}{Peeling}
	\Input{number of machines $m$, dataset $( \bm X_i, \bm Y_i)_{[i = 1,2,\dots m]}$, number of data on each machine $n$, privacy parameters $\varepsilon, \delta$, estimators of parameters $\hat{\bbeta}$, $\hat{\bm \Theta}$ from Algorithms~\ref{algo: privateregression} and \ref{algo:privateprecision}, number of iterations for bootstrap $q$, quantile $\alpha$, noise level $B_4$, subset of coordinates $G$.}
 \For{$t$ \rm{\textbf{from}} $0$ \KwTo $q$}{
 \textbf{Step 1: } For each local machine $i = 1,\dots, m$, generate $n$ independent standard Gaussian random variables $e_{i1}, \dots, e_{in}$. Calculate $\bm u_i = \frac{1}{\sqrt{n}} \sum_{j=1}^n \hat{\bm \Theta} \bm X_{ij} e_{ij}$.\\
 \textbf{Step 2: } Send $(\bm u_i)_{[i=1,2,\dots,m]}$ from local machines to the central server. \\
 \textbf{Step 3: } Calculate  $U_t = \text{Privatemax}([(1/\sqrt{m}) \sum_{i=1}^m \bm u_i]_G,\epsilon, \delta, B_4)$ at the central server.}
	\textbf{Output: } Compute the $\alpha$-quantile $C_U(\alpha)$ of $(|U_1|, |U_2|, \dots|U_q|)$ for $\alpha \in (0,1)$.
	
	\caption{Private Bootstrap Method for Simultaneous Inference in Federated Learning }
\end{algorithm}

On line 4 of Algorithm~\ref{algo: bootstrap_simu}, we employ the Private Max algorithm, which we mentioned earlier as a variation of NoisyHT algorithm (Algorithm~\ref{algo:peeling}) by directly picking $s = 1$,  
to obtain the maximum element in a vector in a private manner. 
It is also important to note that the Private Max algorithm is applied to a subset of $G$. After presenting the algorithm, we denote $M$ 
as:
\begin{equation*}
    M(\hat{\bbeta}) = \max_{k \in G} |\sqrt{mn} (\hat{\beta}_k^{\text{u}} - \beta_k) |.\label{eq: M}
\end{equation*}
$M$ is used as the statistic for inference problems later. 

As previously mentioned, we can easily construct a simultaneous confidence interval for each $k \in G$ by:
\begin{equation*}
    \qty[\hat{\beta}_k^{\text{u}}-\frac{\hat{\sigma}}{\sqrt{mn}}C_U(\alpha), \ \ \hat{\beta}_k^{\text{u}}+\frac{\hat{\sigma}}{\sqrt{mn}}C_U(\alpha)],
\end{equation*}
where $C_U(\alpha)$ is obtained from our algorithm with prespecified $\alpha$. We can similarly perform hypothesis testing; first calculate the test statistic and obtain $C_U(\alpha)$ from our algorithm with prespecified $\alpha$, then reject if the statistic lies in the rejection region. 

\subsection{Theoretical Results}
In this subsection, we provide theoretical guarantee for the algorithms and methods discussed in the previous subsections. Before proceeding, we outline key assumptions concerning the design matrix $\bm X$, precision matrix $\bm \Theta$, and the true parameter $\boldsymbol{\beta}$ of the linear regression model, which are essential for our subsequent analyses.

\begin{itemize}
	\item [(P1)] Parameter Sparsity: The true parameter vector $\bbeta$ satisfies $\|\bbeta\|_2 < c_0$ for some constant $0 < c_0 < \infty$ and $\|\bbeta\|_0 \leq s_0^* = o(n)$.
	\item [(P2)] Precision matrix sparsity: For each column of the precision matrix $\bm \Theta_k$, $k = 1,2,\dots,d$, it satisfies that $\|\bm \Theta_k\|_2 < c_1$ for some constant $0 < c_1 < \infty$ and $\|\bm \Theta_k\|_0 \leq s_1^* = o(n)$.
	\item[(D1)] Design Matrix: for each row of the design matrix $\bm X$, denote by $\bm x$, $\bm x \bm \Sigma^{-1/2}$ is sub-Gaussian with sub-Gaussian norm $\kappa := \|\bm \Sigma^{-1/2}\bm x\|_{\psi_2}$.  
	\item [(D2)] Bounded Eigenvalues of the covariance matrix: For the covariance matrix $\bm \Sigma = \E \bm x \bm x^\top$, there exists a constant $0 < L < \infty$ such that $0 < 1/L <  \lambda_{\min} (\bm \Sigma) \leq \lambda_{\max} (\bm \Sigma) < L$. 
\end{itemize}
The above assumptions (P1) and (P2) bounds the $\ell_2$ norm and $\ell_0$ norm of the parameters $\bbeta$ and $\bm \Theta_k$, and assumption (D1) guarantees that each row of $X$ follows a sub-Gaussian distribution, and assumption (D2) requires the covariance matrix has bounded eigenvalues. These assumptions are commonly used for theoretical analysis of differentially private algorithms and debiased estimators \citep{cai2020cost, cai2019cost, javanmard2014confidence}.

With assumptions (P1)-(D2), we analyze the algorithms we presented. We begin with the estimation problem and provide a rate of convergence of $\hat{\boldsymbol{\beta}}$ and $\hat{\bm \Theta}_k$. 
\begin{Theorem}\label{thm: federated_estimation}
    Let $\{(y_{ij}, \bm X_{ij})\}_{i \in [m], j\in [n]}$ be an i.i.d. samples from the high-dimensional linear model. Suppose that assumptions (P1), (P2), (D1), (D2) are satisfied. Additionally,
    \begin{itemize}
        \item we choose parameters as follows: let $s^* = max(s_0^*, s_1^*)$, $R = \sigma\sqrt{2\log mn}$, $C_0 = c_0$, $C_1 = c_1$, $c_x = 3\sqrt{2L\kappa^2\log d}$, $B_0 = 2(R + \sqrt{s}c_0c_{x})c_{x}$, $B_1 = 2\sqrt{s}c_1 c_x^2$, $\Delta_1 =  \sqrt{s} c_1 c_x R + s c_0 c_1 c_x^2$ and $\gamma = \max(\mu_s(9\mu_s +1/4), 17/16 \mu_s +1/96)$, where $\mu_s, \nu_s$ are the largest and smallest s-restricted eigenvalues of $\hat{\Sigma}$. 
        \item  we set $\bbeta^0 = \bm 0$ and $\bm \Theta_k^ 0 = \bm 0$ as the initialization used in Algorithm~\ref{algo: privateregression} and Algorithm~\ref{algo:privateprecision}.
    \end{itemize}
    Then there exists some absolute constant $\rho > 0$ such that, if $s = \rho L^4 s^*$, $\eta^0 = \eta^1 = s/6L$, $T = \rho L^2 \log(8c_0^2Ln)$ and $n \geq K R(s^*)^{3/2}\log d \sqrt{\log(1/\delta)}\log n/\varepsilon$ for a sufficiently large constant $K > 0$, then, for the output from Algorithm~\ref{algo: privateregression} and Algorithm~\ref{algo:privateprecision}, 
    \begin{align}
        \|\hat{\bbeta} - \bbeta\|_2^2 \le \sigma^2 \qty(k_0\cdot \frac{s\log d}{mn} + \frac{6\gamma \mu_s}{\nu_s^2} \cdot \frac{{s}^2 \log^2 d \log (1/\delta)\log^3 mn} { m^2 n^2 \epsilon^2}),\label{eq: algo regression}
    \end{align}
    and
    \begin{align}
        \|\hat{\bm \Theta}_k - \bm \Theta_k\|_2^2 \le \sigma^2 \qty(k_1\cdot \frac{s\log d}{mn} + \frac{6\gamma \mu_s}{\nu_s^2} \frac{{s}^2 \log^2 d \log (1/\delta)\log^3 mn} { m^2 n^2 \epsilon^2}),\label{eq: algo precision} 
    \end{align}
    hold with probability $1-\exp(- \Omega(\log (d/s\log d) + \log n))$.
\end{Theorem}


The upper bound of Algorithm~\ref{thm: federated_estimation} in \eqref{eq: algo precision} can be interpreted as follows. The first term represents the statistical error, while the second term accounts for the privacy cost. Furthermore, the result is comparable to that of Theorem 4.4 in \citep{cai2019cost}, which addresses private linear regression in a non-federated setting. This comparison suggests that the federated learning approach does not affect the convergence rate adversely; instead, it allows us to leverage the benefits of federated learning. We also note that the advantages of federated learning will be further explored in the heterogeneous federated learning setting, which will be discussed in the next chapter.

The remainder of this subsection presents the theoretical results for the inference problem. We begin with the construction of coordinate-wise confidence intervals. As mentioned before, $\sigma$ is usually unknown and we estimate $\sigma$ in a private manner, presented in Algorithm~\ref{algo:privatevar}. Lemma \ref{lemma:2} states the statistical guarantee of our algorithm.

\begin{lemma}\label{lemma:2}
    Let $\hat \sigma^2$ be the output from Algorithm~\ref{algo:privatevar} by choosing $R =O( \sqrt{2\log mn})$,  $B_2 = \frac{4}{mn} (R^2 + s^2 c_0^2 c_x^2)$ and $\hat{\bbeta}$ as the output from Algorithm~\ref{algo: privateregression}. Then, Algorithm~\ref{algo:privatevar} is $(\epsilon, \delta)$-differentially private, and it follows that
    \begin{align*}
        |\sigma^2 - \hat{\sigma}^2| \le c \cdot \qty(\frac{1}{\sqrt{mn}} + \frac{s\log d}{mn} + \frac{s^2\log^2 d \log(1/\delta) \log^3 mn }{m^2n^2\epsilon^2}),
    \end{align*}
    where $c > 0$ is a universal constant. 
\end{lemma}

Next, we consider a simplified version of the confidence interval, where the privacy cost is dominated by the statistical error. In this scenario, we assume that the privacy level is relatively low and the privacy constraints are loose, meaning that the privacy parameters $\epsilon$ and $\delta$ are relatively large, allowing for nearly cost-free estimation. We present our result in the following theorem.

\begin{Theorem}\label{thm: confidence interval}
	Suppose that the conditions in Theorem~\ref{thm: federated_estimation} hold. Assume that $\frac{s^* \log d}{\sqrt{mn}} = o(1)$ and $\frac{{s^*}^2 \log^2 d \log(1.25/\delta)\log^3 mn}{m n \epsilon^2} = o(1)$. Also assume that the privacy cost is dominated by statistical error, i.e., there exists a constant $k_0$ such that $\frac{{s^*}^2 \log^2 d \log(1/\delta) \log^3 mn}{m^2 n^2 \epsilon^2} \le k_0 \cdot \frac{s^* \log d}{m n} $. Then, given the de-biased estimator $\hat{\beta}_k^{\text{u}}$ defined in \eqref{eq: debiase-estimator}, the confidence interval is asymptotically valid:
	\[\lim_{mn \to \infty } \mathbb{P} (\beta_k \in J_k(\alpha)) = 1-\alpha, \]
	where
	\[J_k(\alpha) = \qty[\hat{\beta}_k^{\text{u}} - \Phi^{-1}(1-\alpha/2) \frac{\hat{\sigma}}{\sqrt{mn}}\sqrt{\hat{\bm \Theta}_k^\top \hat{\bm \Sigma} \hat{\bm \Theta}_k} , \ \ \hat{\beta_k} + \Phi^{-1}(1-\alpha/2) \frac{\hat{\sigma}}{\sqrt{mn}}\sqrt{\hat{\bm \Theta}_k^\top \hat{\bm \Sigma} \hat{\bm \Theta}_k}  ] \]
	Also, the confidence interval $J_k(\alpha)$ is $(\epsilon,\delta)$-differentially private. 
\end{Theorem}

Theorem~\ref{thm: confidence interval} assumes that the privacy cost is dominated by the statistical error. However, when the privacy constraint is more stringent
with small privacy parameters $\epsilon$ and $\delta$, the privacy cost may be larger than the statistical error. In this scenario, we generalize Theorem~\ref{thm: confidence interval} to analyze Algorithm~\ref{algo:coordinate-wise inference}. We note that the largest and smallest restricted eigenvalues of $\hat{\bm \Sigma}$ also need to be estimated by Algorithm~\ref{algo:privateeigen}. Lemma~\ref{lemma:1} quantifies the estimation error of the largest restricted eigenvalue of $\hat{\bm \Sigma}$.

\begin{lemma}~\label{lemma:1}
If $n_1 = c d^s$ and $B_3 = 2 s c_x^2 /n$ for some constant $c > 0$, then the output from Algorithm~\ref{algo:privateeigen} is $(\epsilon, 0 )$-differentially private. Moreover, $(1/9) \lambda_s \le \hat{\lambda}_s \le \lambda_s$ holds where $\lambda_s$ is the largest restricted eigenvalue of $\hat{\bm \Sigma}$.
\end{lemma}

We then present a theoretical result for the confidence interval in a more general case in Theorem~\ref{thm:ci2}.
\begin{Theorem}\label{thm:ci2}
	Assume the conditions in Theorem~\ref{thm: federated_estimation} hold. Suppose that $\frac{s^* \log d}{\sqrt{mn}} = o(1)$ and $\frac{s^{* 2}\log^2 d \log(1.25/\delta) \log^3 mn}{m^2 n^2 \epsilon^2} = o(1)$, then, given the de-biased estimator $\hat{\beta}_k^{\text{u}}$ defined in \eqref{eq: debiase-estimator} and the estimated restricted eigenvalues $\hat{\mu}_s$ and $\hat{\nu}_s$ from Algorithm~\ref{algo:privateeigen}, the confidence interval constructed by Algorithm~\ref{algo:coordinate-wise inference} is asympotically valid:
	\[\lim_{mn \to \infty } \mathbb{P} (\beta_k \in J_k(\alpha)) = 1-\alpha, \]
	where
	\begin{align*}
	    J_k(\alpha) = \biggl[\hat{\beta}_k^{\text{u}} -  \frac{\gamma \hat{\mu}_s^2}{\hat{\nu}_s^2} \frac{s^2 \log^2 d \log (1/\delta)\log^3 mn}{m^2 n^{2} \epsilon^2}- \Phi^{-1}(1-\alpha/2) \frac{\hat{\sigma}}{\sqrt{mn}}\sqrt{\hat{\bm \Theta}_k^\top \hat{\bm \Sigma} \hat{\bm \Theta}_k+ \frac{8\Delta_1^2 \log(1/\delta)}{mn\epsilon^2}} , \notag\\\hat{\beta_j} +   \frac{\gamma \hat{\mu}_s^2}{\hat{\nu}_s^2} \frac{s^2 \log^2 d \log (1/\delta)\log^3 mn}{m^2n^{2} \epsilon^2}+\Phi^{-1}(1-\alpha/2) \frac{\hat{\sigma}}{\sqrt{mn}}\sqrt{\hat{\bm \Theta}_k^\top \hat{\bm \Sigma} \hat{\bm \Theta}_k+ \frac{8\Delta_1^2 \log(1/\delta)}{mn\epsilon^2}}  \biggr] .
	\end{align*}
	Also, $J_k(\alpha)$ is $(\epsilon,\delta)$-differentially private. 
\end{Theorem}

Compared to the non-private counterpart in \citep{javanmard2014confidence}, we claim that our confidence interval has a similar form but with additional noise injected to ensure privacy. When the noise level is low, the confidence interval closely approximates the non-private counterpart, allowing us to nearly achieve privacy without incurring additional costs. Furthermore, when the privacy level is high, the confidence interval has a larger length to attain the same confidence level.

Finally, for the simultaneous inference problems, we demonstrate that $\alpha$-quantile of statistic $M$ in \eqref{eq: M} is close to the $\alpha$-quantile of $U$ calculated in Algorithm~\ref{algo: bootstrap_simu} for each $\alpha\in (0,1)$ using the bootstrap method. The next theorem states the statistical properties of Algorithm~\ref{algo: bootstrap_simu}.
\begin{Theorem}\label{thm: bootstrap_simu}
    Assume the conditions in Theorem~\ref{thm: confidence interval} hold. Additionally, we assume that $\frac{s^* \log d}{\sqrt{mn}}= o(1)$ and the privacy cost is dominated by the statistical error, i.e., there exists a constant $c > 0$ such that $\frac{s^{* 2} \log^2 d \log(1/\delta)\log^3 mn}{m^2 n^2 \epsilon^2} \le c \cdot \frac{s^* \log d}{mn}$. We also assume that there exists a constant $k_0$ such that $\log^7(dmn)/mn \le \frac{1}{(mn)^{k_0}}$, and that  $q = o(mn)$, where $q$ is the number of iterations for bootstrap $q$. The noise level is chosen as $B_4 ={4L\sqrt{\log m }}c_x(R+c_0c_x\sqrt{s^*})/{\sqrt{mn}}$. Then, $C_U(\alpha)$ computed in Algorithm~\ref{algo: bootstrap_simu} satisfies
    \begin{equation*}
        \sup_{\alpha \in (0,1)} |\Pro(M \le C_U(\alpha)) - \alpha | = o(1). 
    \end{equation*}
\end{Theorem}

Theorem~\ref{thm: bootstrap_simu} has useful applications: we can obtain a good estimator of the $\alpha$-quantile of $U$ using the bootstrap method and then use it to construct confidence intervals or perform hypothesis testing. Numerical results will be presented in later chapters to further support our claims.  

\section{Heterogeneous Federated Learning Setting} \label{section5}
\subsection{Methods and Algorithms}

In this section, we consider a more general setting where the parameters of interest on each machine are not identical, but they share some similarities. Specifically, we consider the scenario where, on each machine $i =1,2,\dots,m$, we assume a linear regression model:

\[\bm Y = \bm X \bbeta^{(i)} + \bm W_i,\]
where $\bbeta^{(i)}$ represents the true parameter on machine $i$. We assume that each $\bm W_i$ is a vector whose coordinates follow a sub-Gaussian distribution: $\bm W_{ik} \sim$ subG$(\sigma^2)$, $k =1,2\dots,d$ i.i.d. 
We also assume that each row of $\bm X$ follows a sub-Gaussian distribution i.i.d. with mean zero and covariance matrix $\bm \Sigma$. We further quantify the similarity of each $\bbeta^{(i)}$ by assuming that that there exists a subset $\mathcal{S} \in \{1,2,\dots,d\}$ with $|\mathcal{S}| = s_0$ satisfying $\bbeta^{(i_1)}_{\mathcal{S}} = \bbeta^{(i_2)}_{\mathcal{S}} $ for any $i_1, i_2 \in \{1,2,\dots,m\}$. 

A naive approach would be estimating each $\bbeta^{(i)}$ locally, as in the non-private setting. However, in the context of federated learning, we can improve the estimation with a sharper rate of convergence by exploiting similarities of the model across machines. To achieve this, we decompose $\bbeta^{(i)}$ into the sum of two vectors, $\bbeta^{(i)}= \bm u + \bm v_i$, where $\bm u$ captures the signals common to all $\bbeta^{(i)}$, and $\bm v_i$ captures the signals unique to each machine.

We employ a two-stage procedure to estimate each $\bbeta^{(i)}$: in the first stage, we estimate $\bm u$ using Algorithm~\ref{algo: privateregression} with a sparsity level of $\|\bm u\|_0 = s_0$ indicating the number of shared signals. In the second stage, we estimate $\bm v_i$ on the individual machine. Our final estimation of $\bbeta^{(i)}$ is given by $\hat{\bbeta}^{(i)} = \bm \hat{\bm v}_i + \hat{\bm u}$. The procedure is summarized in Algorithm~\ref{algo:pri_gen}. 

\begin{algorithm}[!htb]
	\SetAlgoLined
	\SetKwInOut{Input}{Input}
	\SetKwInOut{Output}{Output}
	\SetKwFunction{Peeling}{Peeling}
	\Input{Number of machines $m$, dataset $(\bm y_i, \bm X_i)_{[i = 1,\dots,m]}$, number of data on each machine $n$, step size $\eta^0$, privacy parameters $\varepsilon, \delta$, noise scale $ B_5$, number of iterations $T$, truncation level $R$, feasibility parameter $C_0$, initial value $\bm v_i^0$, sparsity level of similar vector $s_0$, sparsity level $s$.}

 \textbf{Step 1: } Estimate a $s_0$ sparse vector $\bm u$ using Algorithm~\ref{algo: privateregression}.\\
 \textbf{Step 2: } Estimate a $s_1 := s-s_0$ sparse vector $\bm v_i$ with samples $(\bm y_i, \bm X_i)$ on machine $i$ with the following iterations from line 3-6.\\
 \For{$t$ \rm{\textbf{from}} $0$ \KwTo $T-1$}{
		\quad\quad Compute $\bm v_i^{t + 0.5} = \bm v_i^t - (\eta^0/n)\sum_{i=1}^{n}  (\bm x_i^\top \bm v_i^t-\Pi_{R}(y_i-\bm x_i^\top \bm u))\bm x_i$\;
		\quad\quad$\bm v_i^{t+1} = \Pi_{C_0}\left(\text{NoisyHT}(\bm v_i^{t+0.5}, (\bm y_i, \bm X_i), s, \varepsilon/T, \delta/T, \eta^0 B_5/n)\right)$.
	\quad\quad}
 \textbf{Step 3: } Estimate $\bbeta^{(i)}$ by $\hat{\bbeta}^{(i)} := \hat{\bm v}_i + \hat{\bm u}$.\\
 \Output{$\hat{\bbeta}^{(i)}$.}
	\caption{Differentially Private Sparse Linear Regression in Heterogeneous Federated Learning Setting}
   \label{algo:pri_gen}
\end{algorithm}
Similar to the previous section, we next address inference problems. Our algorithms consist of two parts: the construction of coordinate-wise confidence intervals and simultaneous inference. We begin by describing the algorithm for coordinate-wise confidence intervals in Algorithm~\ref{algo: coordinate-wise inference2}.

\begin{algorithm}[!htb]
	\SetAlgoLined
	\SetKwInOut{Input}{Input}
	\SetKwInOut{Output}{Output}
	\SetKwFunction{Peeling}{Peeling}
	\Input {Number of machines $m$, dataset $(\bm X_i, \bm Y_i)_{[i=1,\dots,m]}$, number of data points in each machine $n$, privacy parameters $\varepsilon, \delta$, truncation level $R$, sparsity $s$, estimated parameters $\hat{\bbeta}^{(i)}$, $\hat{\bm \Theta}_k$ from Algorithms~\ref{algo:pri_gen} and \ref{algo:privateprecision}, and estimated eigenvalues $\hat{\mu}_s$, $\hat{\nu}_s$ from Algorithm~\ref{algo:privateeigen}, constants $\Delta_1, \gamma$.}
 {
	\textbf{Step 1:}
            \quad Generate a random variable $E_3$ from a Gaussian distribution $N(0,8 \Delta_1^2 \log(1.25/\delta)/(n^2 \epsilon^2) )$. \\
        \textbf{Step 2:}
            \quad Calculate de-biased estimation, $ \hat{\beta}^{(i,\text{u})}_k= \hat{\beta}^{(i)}_k + \frac{1}{n} \sum_{j=1}^n (\hat{\bm \Theta}_k^\top  \bm X_{ij} \Pi_R(\bm y_{ij}) - \hat{\bm \Theta}_k^\top  \bm X_{ij}  \bm X_{ij}^\top  \hat{\beta}^{(i)}_k)+ E_3$.
	
	\textbf{Step 3:} \quad Calculate the confidence interval $J_k(\alpha)$. \begin{align*}
	    J_k(\alpha) = &\biggl[\hat{\beta}^{(i,\text{u})}_k - a - \frac{\sigma\Phi^{-1}(1-\alpha/2) }{\sqrt{n}}\sqrt{\hat{\bm \Theta}_k^\top \hat{\bm \Sigma} \hat{\bm \Theta}_k+ \frac{8\Delta_1^2 \log(1/\delta)}{n\epsilon^2}} , \notag\\&\hat{\beta}^{(i,\text{u})}_k + a + \frac{\sigma\Phi^{-1}(1-\alpha/2) }{\sqrt{n}}\sqrt{\hat{\bm \Theta}_k^\top \hat{\bm \Sigma} \hat{\bm \Theta}_k+ \frac{8\Delta_1^2 \log(1/\delta)}{n\epsilon^2}}  \biggr],
	\end{align*}
    where $a$ is defined in \eqref{eq: a}.
	}
 
	\textbf{Output: } Return the final result $J_k(\alpha)$.
	
	\caption{Differentially Private Coordinate-wise Confidence interval for $\beta_k$ in Heterogeneous Federated Learning}
    \label{algo: coordinate-wise inference2}
\end{algorithm}
In Algorithm~\ref{algo: coordinate-wise inference2}, $\hat{\bm \Theta}_j$ is the $(\epsilon,\delta)$-differentially private estimator of the $j$-th row of the precision matrix of covariance matrix $\hat{\Sigma} =1/(mn) \sum_{i=1}^m \sum_{j=1}^n \bm X_{ij} \bm X_{ij}^\top$. We define the variable $a$ in step 3 by
\begin{align}
    a := \frac{2\gamma \hat{\mu}_s^2}{\hat{\nu}_s^2} \frac{s_1^2 \log^2 d \log (1/\delta)\log^3 mn}{m^2 n^2 \epsilon^2} + \frac{2\gamma \hat{\mu}_s^2}{\hat{\nu}_s^2} \frac{s_0^2 \log^2 d \log (1/\delta) \log^3 n }{n^2 \epsilon^2}.\label{eq: a}
\end{align}

We then provide Algorithm \ref{algo: bootstrap_simu_single} for the simultaneous inference problem. Similar to the previous chapter, we can perform simultaneous inference for each $\bbeta^{(i)}$ to build simultaneous confidence interval and hypothesis testing.

\begin{algorithm}[H]\label{algo: bootstrap_simu_single}
	\SetAlgoLined
	\SetKwInOut{Input}{Input}
	\SetKwInOut{Output}{Output}
	\SetKwFunction{Peeling}{Peeling}
	\Input{Dataset $(\bm y_i, \bm X_i)$, number of data $n$, privacy parameters $\varepsilon, \delta$, estimators of parameters $\hat{\bbeta}^{(i)}$, $\hat{\bm \Theta}$ from Algorithms~\ref{algo:pri_gen} and \ref{algo:privateprecision}, number of iterations for Bootstrap $q$, quantile $\alpha$, noise level $B_6$. \lm{$\sigma$?}}
	\For{$t$ \rm{\textbf{from}} $0$ \KwTo $q$}{
	Generate $n$ independent standard Gaussian random variables $e_1, \dots, e_n$.\\ Calculate $U_t = \|\text{Privatemax}([\frac{\sigma}{\sqrt{n}}\sum_{j=1}^n \hat{\Theta}\bm X_{ij} e_j]_G,\epsilon, \delta, B_6)\|_{\infty}$}
	
	\textbf{Output: } Compute the $\alpha$-quantile $C_U(\alpha)$ of $(|U_1|, |U_2|, \dots,|U_q|)$ for $\alpha \in (0,1)$.
	
	\caption{Private Bootstrap Method for Simultaneous Inference in Heterogeneous Federated Learning for Machine $i \in \{1, \dots, m\}$}
\end{algorithm}
Compared with Algorithm \ref{algo: bootstrap_simu} introduced for simultaneous inference in homogeneous federated learning, bootstrap algorithm in Algorithm \ref{algo: bootstrap_simu_single} runs within the local machine of interest.
Using the output from Algorithm \ref{algo: bootstrap_simu_single}, we build a simultaneous confidence interval for each $\beta_k^{(i)}$ ($k \in G$) using $C_U(\alpha)$ by
$$\qty[\hat{\beta}_k^{(i,\text{u})}-\frac{1}{\sqrt{n}} C_U(\alpha), \ \ \hat{\beta}_k^{(i,\text{u})}+\frac{1}{\sqrt{n}} C_U(\alpha)].$$

\subsection{Theoretical Results}
In this subsection, we provide theoretical analysis for the algorithms in heterogeneous federated learning settings.
We begin our theoretical analysis with the estimation problem, which resembles Theorem~\ref{thm: federated_estimation}.

Intuitively, when $\bbeta^{(i)}$ are similar but not identical, federated learning can be used to estimate their common elements and the remaining parameters can be estimated individually on each machine. 
This results in a sharper rate of convergence as the estimation of the common component $\bm u$ can exploit the information from more data points. We summarize the result in Theorem \ref{thm:estimate_diffbeta}.
\begin{Theorem}\label{thm:estimate_diffbeta}
	Assume that the conditions in Theorem~\ref{thm:ci2} hold.
    Further assume that for Algorithm~\ref{algo:pri_gen}, $\|\bm v_i\|_0 = s_1 = s-s_0$ for all $i=1,\dots,m$, $\|\bm u\|_0 = s_0$, $\|\bm u\|_2 \le c_0/2$, and $\|\bm v_i\|_2 \le c_0/2$. 
    Let $B_5 = c_x(2R + \sqrt{s_1}c_0c_x)$. Then, for the output $\hat{\bbeta}^{(i)}$ from Algorithm~\ref{algo:pri_gen}, we have
    \begin{align}
        \|\hat{\bbeta}^{(i)} - \bbeta^{(i)}\|_2^2 \le c_0 \frac{s_0 \log d}{mn} + c_1 \frac{{s_0}^2 \log d^2 \log(1/\delta)\log^3 mn}{m^2 n^2 \epsilon^2} + c_2 \frac{s_1\log d}{n} + c_3 \frac{{s_1}^2 \log d^2 \log(1/\delta)\log^3 n}{n^2 \epsilon^2},\label{eq: beta i m n}
    \end{align}
	where $c_0,c_1,c_2,c_3 > 0$ are some constants.
\end{Theorem}
In the case where $s_0 \ll s_1$, i.e., the models are largely different across machines, the third and fourth term on the right hand side of \eqref{eq: beta i m n} dominates the estimation error, and the estimation accuracy of $\bbeta^{(i)}$ via federated learning becomes closer to that with a single machine ($m=1$). 
In high level, this is because the information from other machines is not helpful in the estimation when there exists a large dissimilarity of models across machines.
However, with a large $s_0 \gg s_1$, federated learning can leverage the similarity of models to improve estimation accuracy. As a result, the rate in \eqref{eq: beta i m n} becomes closer to the rate in \ref{eq: algo regression} for homogeneous federated learning setting when $s_0/s_1 \to 0$.

We next present our results for the inference problems.
To start we verify that the output from Algorithm~\ref{algo: coordinate-wise inference2} is a asymptotic $1-\alpha$ confidence interval for $\beta^{(i)}_k$.
\begin{Theorem}\label{thm:ci3}
    Assume the conditions in Theorem~\ref{thm: federated_estimation} hold and assume that $\frac{s^* \log d}{\sqrt{n}} = o(1)$ and $\max(\frac{2\gamma \hat{\mu}_s^2}{\hat{\nu}_s^2} \frac{s_1^2 \log^2 d \log (1/\delta)\log^3 mn}{m^2 n^2 \epsilon^2} , \frac{2\gamma \hat{\mu}_s^2}{\hat{\nu}_s^2} \frac{s_0^2 \log^2 d \log (1/\delta) \log^3 n }{n^2 \epsilon^2} )= o(1)$. 
    Let $a$ be the variable defined in \eqref{eq: a}.
    Then, for the de-biased estimator $\hat{\beta}_k^{(i,\text{u})}$ defined in \eqref{eq: debiase-estimator}, the constructed confidence interval is asympotically valid:
	\[\lim_{n \to \infty } \mathbb{P} (\beta_k^{(i)} \in J_k(\alpha)) = 1-\alpha, \]
	where
	   \begin{align*}
   	J_k(\alpha) = &\biggl[\hat{\beta}_k^{(i,\text{u})} - a - \frac{\hat{\sigma}\Phi^{-1}(1-\alpha/2) }{\sqrt{n}}\sqrt{\hat{\bm \Theta}_k^\top \hat{\bm \Sigma} \hat{\bm \Theta}_k+ \frac{8\Delta_1^2 \log(1/\delta)}{n\epsilon^2}} , \notag\\&\hat{\beta}_k^{(i,\text{u})} + a + \frac{\hat{\sigma}\Phi^{-1}(1-\alpha/2) }{\sqrt{n}}\sqrt{\hat{\bm \Theta}_k^\top \hat{\bm \Sigma} \hat{\bm \Theta}_k+ \frac{8\Delta_1^2 \log(1/\delta)}{n\epsilon^2}} \biggr] 
\end{align*}
	Also, $J_k(\alpha)$ is $(\epsilon,\delta)$-differentially private. 
\end{Theorem}

Finally, we provide a statistical guarantee for Algorithm~\ref{algo: bootstrap_simu_single}. Similar to the previous section, we define $M$ as:
\begin{equation*}
    M = M(\hat{\bbeta}^{(i,\text{u})}) = \max_{k \in G}|\sqrt{n} (\hat{\beta}_k^{(i,\text{u})} - \beta_k^{(i)}) |.
\end{equation*}

\begin{Theorem}\label{thm: bootstrap_simu_single}
    Assume that the conditions in Theorem~\ref{thm: confidence interval} hold. We additionally assume that $\frac{s^* \log d}{\sqrt{n}}= o(1)$ and the privacy cost is dominated by the statistical error, i.e., there exists a constant $c$ such that $\frac{s^{* 2} \log^2 d \log(1/\delta)\log^3 mn}{m^2 n^2 \epsilon^2} \le c \cdot \frac{s^* \log d}{mn}$ and $\frac{s^{* 2} \log^2 d \log(1/\delta)\log^3 n}{n^2 \epsilon^2} \le c \cdot \frac{s^* \log d}{n}$. 
    We also assume that there exists a constant $k_0$ such that $\log^7(dn)/n \le \frac{1}{n^{k_0}}$. The noise level is chosen as $B_6 =2\sqrt{\frac{s \log n}{n}} c_x c_1$. Then,
    \begin{equation*}
        \sup_{\alpha \in (0,1)} |\Pro(M \le C_U(\alpha)) - \alpha | = o(1). 
    \end{equation*}
\end{Theorem}

Theorem \ref{thm: bootstrap_simu_single} states that $\alpha$-quantile of $M$ is asymptotically close to $C_U(\alpha)$, which validates the $1-\alpha$ simultaneous confidence intervals based on $C_U(\alpha)$ obtained by the bootstrap method.
This result allows us to perform simultaneous inference such as the confidence intervals and hypothesis testing based on $C_U(\alpha)$.

\section{Simulations}
In this section, we conduct simulations to investigate the performance of our proposed algorithm as discussed in the preceding sections. Specifically, we explore the more complex heterogeneous federated learning setting, where each machine operates on different models yet exhibits similarities. Our simulations are divided into three main parts.

In Section~\ref{sec: simu1}, we present the simulation results for the coordinate-wise estimation problem within a private federated setting, discussing the differences between the estimated $\hat{\bbeta}$ and the true $\bbeta^*$ across various scenarios. We also examine the coverage of our proposed confidence intervals. 
Section~\ref{sec:simu3} extends the settings to simultaneous inference. 

We generate simulateion simulation datasets as follows.
First, we sample the data $\bm X_i$, for $i = 1, 2, \ldots, m$, where each $\bm X_i$ follows a Gaussian distribution with mean zero and covariance matrix $\bm \Sigma$. We set $\bm \Sigma$ such that for each $j, j' \in \{1,2,\ldots,d\}$, $\Sigma_{j,j'} = 0.5^{|j-j'|}$. 
On each machine, we assume a $s^*$-sparse unit vector $\bbeta^{(i)}$ with $s^* = s_0 + s_1$, where $s_0$ is the number of non-zero shared signals. For each $\bbeta^{(i)}$, we set the first $s_0$ shared elements to $1/\sqrt{s^*}$ and additionally select machine-specific $s_1$ entries from the remaining $d-s_0$ indices to be $1/\sqrt{s^*}$. We then compute $\bm Y_i = \bm X_i \bbeta^{(i)} + \bm W_i$, where each $\bm W_i$ follows a Gaussian distribution $N(\bm 0, \sigma^2 \bm I)$ with $\sigma = 0.5$.

\subsection{Estimation and Confidence Interval}\label{sec: simu1}
In this subsection, we investigate the estimation accuracy and confidence interval coverage of our algorithm for coordinate-wise inference. Namely, we consider the following scenarios:
\begin{itemize}
    \item Fix number of machines $m = 15$, $\epsilon = 0.8$, $\delta = 1/(2mn)$, $d=800$, $s^* = 15$ and $s_0 = 6$. Set the number of samples on each machine to be $4000, 5000, 6000$, respectively. 
    \item Fix number of samples on each machine $n = 4000$, $\epsilon = 0.8$, $\delta = 1/(2mn)$, $d=800$, $s^* = 15$ and $s_0 = 6$. Set the number of machines $m$ to be $5, 10, 15$, 
    \item Fix number of machines $m = 15$, number of samples on each machine $n = 4000$, $\epsilon = 0.8$, $\delta = 1/(2mn)$, $d=800$. Set, $s^* = 15, s_0 = 6$, $s^* = 10, s_0 = 4$, $s^* = 20, s_0 = 8$, respectively.  
    \item Fix number of machines $m = 15$, number of samples on each machine $n = 4000$, $\epsilon = 0.8$, $\delta = 1/(2mn)$, $d=800$, $s^* = 15$. Set $s_0 = 6, 8, 10$, respectively. 
    \item Fix number of machines $m = 15$, number of samples on each machine $n = 4000$, $\delta = 1/(2mn)$, $d=800$, $s^* = 15$ and $s_0 = 6$. Set $\epsilon = 0.3, 0.5, 0.8$ respectively. 
    \item Fix number of machines $m = 15$, number of samples on each machine $n = 4000$, $\epsilon = 0.8$, $\delta = 1/(2mn)$, $s^* = 15$ and $s_0 = 6$. Set $d = 600,800,1000$, respectively. 
\end{itemize}
For each setting, we report the average estimation error $\|\hat{\bbeta} - \bbeta^*\|_2^2$ among 50 replications. Also, in each setting, we calculate the confidence interval with $\alpha = 0.95$ for each index of $\bbeta^*$ using our proposed algorithm. To evaluate the quality of confidence interval, we define $\cov$ as the coverage of the confidence interval:
\[ \cov := d^{-1}\sum_{i=1}^d \Pro[\beta^*_i \in J_i(\alpha)].\]
We also define the coverage for non-zero and zero entries of $\bm \beta^*$ by $\cov_{\mathcal{S}}$ and $\cov_{\mathcal{S}^c}$, respectively, where $\mathcal{S}$ is the set of non-zero indices in $\bbeta^*$.
\[\cov_{\mathcal{S}} = |\mathcal{S}|^{-1} \sum_{i\in \mathcal{S}} \Pro[\beta^*_i \in J_i(\alpha)] \quad , \quad \cov_{\mathcal{S}^c} = |\mathcal{S}^c|^{-1}\sum_{i\in \mathcal{S}^c} \Pro[\beta^*_i \in J_i(\alpha)].\]

We report the estimation error, coverage of true parameter and length of confidence interval for each configuration listed above in Table~\ref{fig:table}:\\

\renewcommand{\figurename}{Table}
\begin{figure}[h]
\centering
	\begin{tabular}{ |p{4cm}||p{3.5cm}|p{1cm}|p{1cm}|p{1cm}|p{1cm}| }
 \hline
 \multicolumn{6}{|c|}{Simulation Results} \\
 \hline
 $(n, m, d, s^*, s_0, \epsilon)$ & Estimation Error (Sd) & $\cov$ &$\cov_{\mathcal{S}}$&$\cov_{\mathcal{S}^c}$&length\\
 \hline
 (3000,15,800,15,8,0.8)  & 0.0213 (0.0028) &0.940&0.929&0.940&0.0532\\
 (4000,15,800,15,8,0.8) &  0.0170 (0.0032)  & 0.945   &0.960&0.944&0.0437\\
 (5000,15,800,15,8,0.8) & 0.0141 (0.0021) & 0.940&0.945&0.940&0.0378\\
 (4000,10,800,15,8,0.8) & 0.0218 (0.0047) & 0.945&0.945&0.945&0.0437\\
 (4000,20,800,15,8,0.8) & 0.0126 (0.0025)  & 0.944&0.941&0.944&0.0437\\
 (4000,15,600,15,8,0.8) & 0.0162 (0.0031) &0.946&0.933&0.946&0.0436\\
 (4000,15,1000,15,8,0.8) & 0.0191 (0.0027)  & 0.940&0.933&0.940&0.0439\\
 (4000,15,800,15,4,0.8) & 0.0188 (0.0032)  & 0.952&0.945&0.953&0.0420\\
 (4000,15,800,15,12,0.8) & 0.0137 (0.0016)  & 0.944&0.937&0.944&0.0462\\
 (4000,15,800,10,8,0.8) & 0.0105 (0.0017)  & 0.946&0.947&0.946&0.0389\\
 (4000,15,800,20,8,0.8) & 0.0243 (0.0036)  & 0.941&0.932&0.941&0.0497\\
 (4000,15,800,15,8,0.5) & 0.0240 (0.0038)  & 0.940&0.949&0.940&0.0550\\
 (4000,15,800,15,8,0.3) & 0.0943 (0.0281)  & 0.928&0.941&0.928&0.0792\\
 \hline
\end{tabular}
\caption{Table for Simulation Results of the private federated linear regression}\label{fig:table}
\end{figure}
\renewcommand{\figurename}{Fig}

\begin{figure}[H]
\centering
\includegraphics[width=10cm]{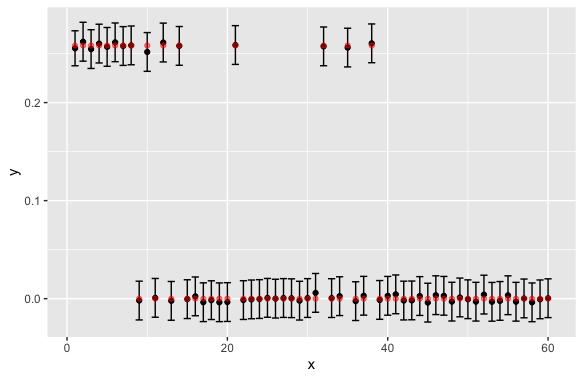}
\caption{Confidence intervals for $\beta_k$ for each coordinate $k$ randomly selected from $800$ coordinates. vertical axis stands for the value of $\beta_k$. Red points stand for the true $\beta_k$ while black points stand for the estimated $\beta_k$. We mention that the result averaged over 50 iterations.}\label{fig:ci}
\end{figure}

From Table \ref{fig:table}, we observe a consistent result with our theory.
Namely, for the estimation error, the error becomes small as $\epsilon$ gets larger as we require less level of privacy. Also, more data points on each machine, more number of machines, smaller sparsity level lead to better estimation accuracy.
For confidence intervals, we observe that the coverage is close to $0.95$ for $\cov$, $\cov_\mathcal{S}$, and $\cov_{\mathcal{S}^c}$, is and stable in different settings. 
To further illustrate our claim, we pick the setting of $(n,m,d,s,s_0,\epsilon)=(4000,15,800,15,8,0.8)$ and plot the confidence intervals versus the true value among 50 replications in Figure~\ref{fig:ci}. We randomly select 60 out of $800$ coordinates.





We also summarize our results in Figure \ref{fig:esti}, where we plot the estimation error against the change in the number of samples, sparsity, and number of machines.
For the figure, we fixed $m=15$, $d=800$, $s^*=16$, $s_0=8$, for the middle figure, we fixed $n=4000$, $m=15$, $d=800$, $\epsilon=0.5$, and for the right figure, we fixed $d=800$, $s^*=16$, $s_0=8$, $\epsilon=0.5$. The error is averaged over $200$ replications.
\begin{figure}[H]
\centering
\includegraphics[width=14cm]{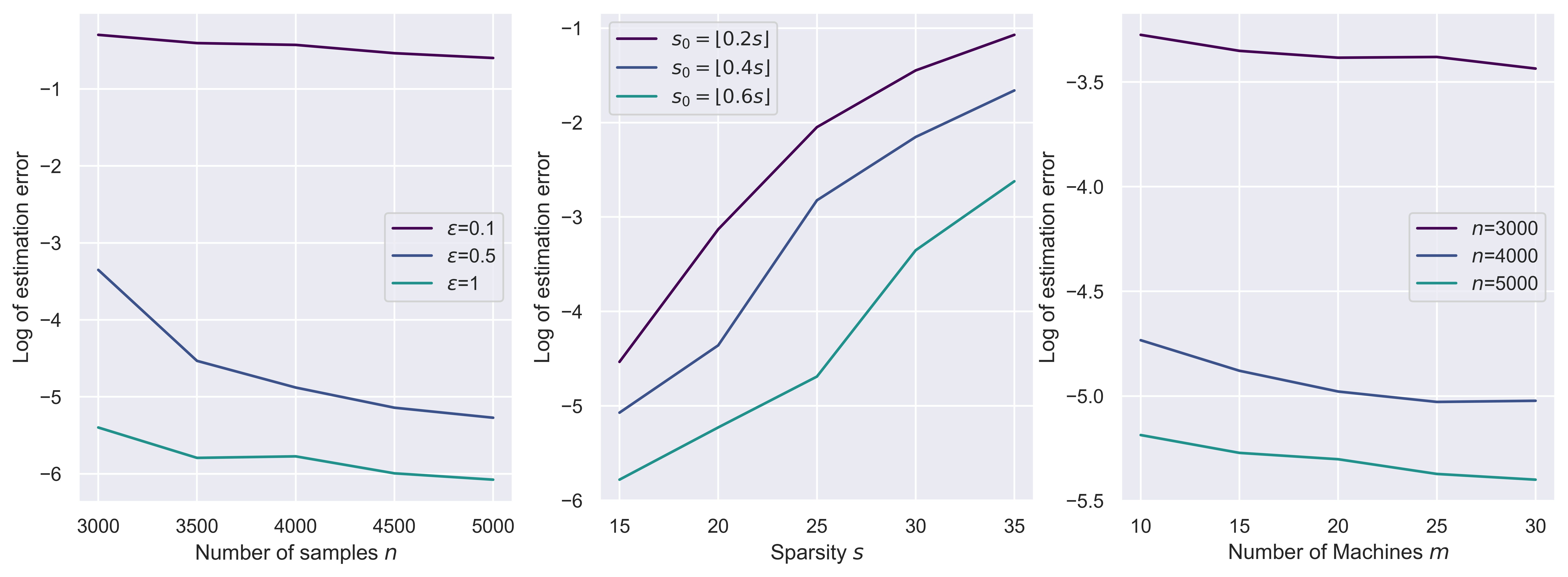}
\caption{Plot for the estimation results. \textbf{Left:} Log estimation error with different number of samples $n$, \textbf{Middle:} Log estimation error with different sparsity $s^*$, \textbf{Right:} Log estimation error with different number of machines $m$.}\label{fig:estimate}
\end{figure}\label{fig:esti}
From the left figure in Figure \ref{fig:esti}, we observe the decreasing error when we increase $n$. When the privacy parameter $\epsilon$ is large, we have better estimation error.
From the middle figure, we observe that as the sparsity level $s$ grows, the estimation error also increases. Also, when the sparsity for the shared signal $s_0$ becomes large, the estimation error also becomes large. In the right figure, we observe a consistent decrease of error when we increase the number of machines. All these figures support Theorem \ref{thm:estimate_diffbeta}.

\subsection{Simultaneous Inference}\label{sec:simu3}
In this subsection, we investigate our proposed algorithms for simultaneous inference problems. We aim to build a simultaneous confidence interval when $\alpha = 0.05$ under three settings: $G=\{1, 2, \dots, d\}$, $G = S$, and $G=S^c$. 
For each setting, we repeat 50 simulations and report the coverage and length of the confidence intervals. The results are shown in Table~\ref{table:simu}.

\renewcommand{\figurename}{Table}
\begin{figure}
\centering
	\begin{tabular}{ |p{3.6cm}||p{1.3cm}|p{1.3cm}|p{1.3cm}|p{1.3cm}|p{1.3cm}|p{1.4cm}| }
 \hline
 \multicolumn{7}{|c|}{Simulation Results for Simultaneous Inference} \\
 \hline
$(n, m, d, s^*, s_0, \epsilon)$ & $\cov$&$\cov_{\mathcal{S}}$&$\cov_{\mathcal{S}^c}$&len($\cov$)&len($\cov_{\mathcal{S}}$)&len($\cov_{\mathcal{S}^c}$)\\
 \hline
 (3000,15,800,15,8,0.8)   &0.981 &0.883 & 0.983 & 0.091&0.066&0.091\\
 (4000,15,800,15,8,0.8) & 0.985    &0.910 & 0.987 & 0.079&0.057&0.079\\
 (5000,15,800,15,8,0.8)  & 0.987 &0.875 & 0.990 & 0.071&0.051&0.071\\
 (4000,10,800,15,8,0.8)     & 0.989 &0.894 & 0.991 & 0.079&0.057&0.079\\
 (4000,20,800,15,8,0.8)  & 0.983 &0.898 & 0.986 & 0.079&0.057&0.079\\
 (4000,15,600,15,8,0.8) & 0.993 &0.878 & 0.995 & 0.077&0.057&0.077\\
 (4000,15,1000,15,8,0.8) & 0.994 &0.878 & 0.997 & 0.080&0.057&0.080\\
 (4000,15,800,15,4,0.8)  & 0.983 &0.772 & 0.993 & 0.079&0.057&0.079\\
 (4000,15,800,15,12,0.8) & 0.975 &0.957 & 0.974 & 0.079&0.058&0.079\\
 (4000,15,800,10,8,0.8) & 0.986 &0.976 & 0.985 & 0.079&0.055&0.079\\
 (4000,15,800,20,8,0.8)  & 0.974 &0.850 & 0.982 & 0.078&0.059&0.078\\
 (4000,15,800,15,8,0.5) & 0.940 &0.882 & 0.940 & 0.103&0.083&0.102\\
 (4000,15,800,15,8,0.3) & 0.953 &0.789 & 0.975 & 0.127&0.097&0.126\\
 \hline
\end{tabular}
\caption{Simulation results of the private simultaneous inference in different settings. }\label{table:simu}
\end{figure}
From simulation results, we can observe that our proposed simultaneous confidence interval mostly exhibit over-coverage for $G = \mathcal{S}^c$, and under-coverage for $G = \mathcal{S}$. 
This pattern has also been observed in previous works addressing simultaneous inference \citep{van2014asymptotically, zhang2017simultaneous}. Therefore, this could be attributed to the inherent nature of simultaneous inference rather than to algorithmic reasons.
\renewcommand{\figurename}{Fig}

\section{Discussions and Future Work}

In this paper, we study the high-dimensional estimation and inference problems within the context of federated learning. In scenarios involving an untrusted central server, our findings reveal that accurate estimation is infeasible, as the rate of convergence is adversely proportional to the dimension $d$. Conversely, in the trusted central server setting, we developed algorithms that achieve an optimal rate of convergence. We also explored inference challenges, detailing methodologies for both point-wise confidence intervals and simultaneous inference.

There are several extensions for further research. Currently, our models presume that each machine operates under a linear regression framework. We can possibly expand our algorithm to accomodate more complex models, such as generalized linear models, classification models, or broader machine learning models. 
Moreover, an interesting extension would be to refine our understanding of model similarity across machines. Although Section~\ref{section5} currently bases model similarity on $L_0$ norms, reflecting non-sparse patterns, future studies could explore $L_p$ norm-based similarities, particularly focusing on $L_1$ and $L_2$ norms, to enhance our approach to heterogeneous federated learning settings.

\bibliographystyle{plain}
\bibliography{reference}

\newpage
\appendix

\section{Proof of main results}
\subsection{Proof of Theorem~\ref{thm1}}
We show the proof of the lower bound of the estimation. The main idea of the proof is as follows, we will first assume that in the general case where each data point on each machine follows a general distribution $p_{\theta}$, then we will further assume some conditions of this distribution, and prove that the lower bound of the mean estimation could be attained under these conditions. Finally, we will show that under the assumptions that the data points follow the normal distribution, the specific conditions hold, thus we could finish the proof.

To start this proof, we first introduce the perturbation space $\mathcal{A} = \{-1, 1\}^k$, where $k$ is a pre-chosen constant and associate each parameter $\theta$ with $a \in \mathcal{A}$ and refer the distribution $p_{\theta}$ as $p_a$. We characterize the distance between two parameters $\theta$ and $\theta'$ by the hamming distance of $z$ and $z'$, such approach will be compatible with the Assouad's method, as will be shown later in the proof. We note that when the hamming distance of $a$ and $a'$ get smaller, it indicts that the distance between $\theta$ and $\theta'$ becomes closer. Also, for each $a\in \mathcal{A}$, we further denote
$a^{\oplus i}\in\mathcal{A}$ as the vector which flips the sign of the
$i$-th coordinate of $a$. Then, we state below conditions: 
\begin{condition}~\label{Cond:1}
For every $a\in\mathcal{A}$ and $i\in [k]$, it holds that $\bm p_{a^{\oplus i}} \ll \bm p_{a}$. Further,
there exist $q_{a,i}$ and measurable functions $\phi_{a,i}\colon\cX\to\R$
such that $|q_{a,i}| \leq \alpha$, which $q$ is a constant and: 
\[
 \frac{d\bm p_{a^{\oplus i}}}{d\bm p_a}=1+q_{a,i}\phi_{a,i}.
 \]
\end{condition}
\begin{condition}~\label{Cond:2}
For all $a\in\mathcal{A}$ and $i,j\in[k]$,
$\E_{\bm p_{a}}[\phi_{a,i}\phi_{a,j}]=\mathbf{1}_{i=j}$.
\end{condition}
\begin{condition}~\label{Cond:3}
There exists some $\sigma \ge 0$ such
that, for all $a\in\mathcal{A}$, the random vector $\phi_a(X) =
(\phi_{a,i}(X))_{i\in[k]}\in\mathbb{R}^k$ is $\sigma^2$-sub-Gaussian
for $X\sim \bm p_a$ with independent coordinates.
\end{condition}
The above conditions characterize the distribution $\bm p_a$, we will later verify that the Gaussian distribution could satisfy the above conditions in the later proof. Then, we state our first claim. 
\begin{Corollary}~\label{corollary:1}
For each coordinate of the $A$, for any $i = 1,2,\dots,k$, fix $\tau = \mathbb{P}(A_i = 1) \in(0,1/2]$. Let $X_1, \ldots, X_m$ be the inputs on the local servers,  i.i.d.\ with common distribution $\bm p_A^{\otimes n }$. Let $Z^{m}$ be the information sent from all the local servers to the central machine generated through the channel $\mathcal{W}$. Then, if the condition~\ref{Cond:1} satisfies, there exists a constant $c$, we have:
\begin{align*}
&\qty(\frac{1}{k}\sum_{i=1}^k d_{\text{TV}}({\bm p_{+i}^{Z^{m}}},{\bm p_{-i}^{Z^{m}}}))^2 \le    
\frac{c}{k}q^2 m n^2  \max_{a\in\mathcal{A}}\sum_{i=1}^k \int_{\cY} \frac{\E_{\bm p_{a}^{\otimes n}}[{\phi_{a,i}(X)\mathcal{W}(z | X)}]^2}{\E_{\bm p_{a}^{\otimes n}}[{\mathcal{W}(z| X)}]} d {\mu},
\end{align*}
where $\bm p_{+i}^{Z^{m}} =  \E [\bm p_A^{Z^{m}} | A_i= + 1]$, 
$\bm p_{-i}^{Z^{m}} =  \E [\bm p_A^{Z^{m}}|A_i= - 1]$.
\end{Corollary}
The proof of the above corollary is in appendix~\ref{proof:col1}. The above corollary characterizes the difference between the distribution of $\bm p_{+i}^{Z^{m}} $ and $\bm p_{-i}^{Z^{m}}$, which is the difference between the distribution of the information about the each coordinate of $A$, which could be seen as the information between $Y$ and $A$, namely, the information between the information and the parameters. 

In the precious corollary, we just assumed a general channel $\mathcal{W}$, in the following corollary, we could further specifies the above corollary when the channel $\mathcal{W}$, be a $\epsilon$-differentially private constraint channel $\mathcal{W}^{priv}$ and we could further simplify the upper bound in Corollary~\ref{corollary:1}.

\begin{Corollary}~\label{corollary:2}
If $W^{priv}$ be a privacy constraint channel and for any family of distributions $\{\bm p_a, a\in \{-1,1\}^k\}$
satisfying condition 1 and condition 2. With the same notations as Corollary~\ref{corollary:1} we have:
\begin{align*}
\qty(\frac{1}{k}\sum_{i=1}^k d_{TV}({\bm p_{+i}^{Z^{m}}},{\bm p_{-i}^{Z^{m}}}))^2 
 \le \frac{7}{k} m n^2 q^2 (e^{n\epsilon^2}-1)
\end{align*}
\end{Corollary}

The proof of the above corollary could be found in appendix~\ref{proof:col2}. The above corollary focus on the upper bound of $\frac{1}{k}\sum_{i=1}^k d_{TV}({\bm p_{+i}^{Z^{m}}},{\bm p_{-i}^{Z^{m}}})$, in the next corollary, we will focus on the lower bound, which is an Assouad-type bound. We first introduce another condition:

\begin{condition}~\label{Cond:4}
Fix $p \in [1,\infty)$. Let $\rho$ be the $\ell_p$ loss between the true parameter and the estimation. Then, for every $a,a^\prime\in\mathcal{A} \in \{-1,+1\}^k$, the below inequalities hold:
\begin{align*}
l_p(\theta_a,\theta_{a^\prime})\ge 4 \rho \qty(\frac{d_{Ham}(a,a')}{\tau k})^{1/p},
\end{align*}
where $d_{Ham}(a,a')$ denotes the Hamming distance with definition $d_{Ham}(a,a') = \sum_{i=1}^k \mathbf{1}{(a_i \neq a_i')}$, and $\tau = \P(a_i =1) \in (0,1/2]$ for each coordinate $a_i$.
\end{condition}

The above condition characterizes the connection between $\theta$ with the perturbation space. With the above assumption, we could further obtain the lower bound of $\frac{1}{k}\sum_{i=1}^k d_{TV}({\bm p_{+i}^{Z^{m}}},{\bm p_{-i}^{Z^{m}}})$: 
\begin{Corollary}~\label{corollary:3}
Let $p \ge 1$ and assume that $\{\bm p_a, a\in \mathcal{A}\}$, $\tau \in[0,1/2]$ satisfy Condition~\ref{Cond:4}. Let $A$ be a random variable on $\{-1,1\}^k$ with distribution $Rad(\tau)^{\otimes k }$. Suppose that $\hat{\theta}$ constitutes an $(n, \rho)$-estimator of the true parameter $\theta^*$ under $l_p$ loss and $\mathbb{P} [\bm p_A \in \mathcal{P}_\Theta] \ge 1 -  \tau/4$. Then the below inequality holds:
\begin{align*}
	\frac{1}{k}\sum_{i=1}^k d_{TV}(\bm p_{+i}^{Z^m}, \bm p_{-i}^{Z^m}) \geq \frac{n}{4},
\end{align*}
where $\bm p_{+i}^{Z^m} =  \E[\bm p_A^{Z^m}|{A_i= + 1}]$, $\bm p_{-i}^{Z^m} =  \E [{\bm p_A^{Z^m}}|{A_i= - 1}]$.
\end{Corollary}

The proof of the above corollary could be found in appendix~\ref{proof:col3}. In the following proof, we are going to verify that the Gaussian distribution satisfies all the above conditions, thus the result in Corollary~\ref{corollary:2} and Corollary~\ref{corollary:3} holds. Then, according to these two corollaries, we will present the lower bound for the mean estimation in the high-dimensional federated learning setting. \\

For the parameters, we could fix $p =2$, $k=d$, $\mathcal{A}= \{-1, +1\}^d$. For the probability where $\tau = \Pro(a_i) = 1$, we fix $\tau=\frac{s}{2d}$. Let $\varphi$ denote the probability density function of the standard Gaussian distribution $N(\mathbf{0}, \bm I)$. We first suppose that, for some $\rho \in (0,1/8]$, there exists an $(n,\rho)$-estimator for the true parameter $\mu$ under $\ell_p$ loss. Then, if we have $\rho^2 \ge s/n$, then we could finish the proof. Otherwise, we fix a parameter $\gamma  =  \frac{4\rho}{\sqrt{s/2}}\in(0,1/2]$, this is possible with a choice of $s$, the sparsity level. We could design the parameter, the mean of the Gaussian distribution $\mu$ and $A$ by the formula: $\mu_a  =  \gamma (a+ \bm 1_{d})$, where $a\in\mathcal{A}$. Then, we could verify that $\Pro[\|\mu_a\|_0 \le 2 \tau d] \ge 1-\tau/4$, where $\norm{\mu_a}_0 = \sum_{i=1}^d \bm 1_{a_i=1} = \|a\|_{+}$. From the definition of Gaussian density, for $a\in\mathcal{A}$, we have:  

\begin{equation*}
\nonumber
  \bm p_a(x) = e^{-\gamma^2\|\mu_a\|_2^2/2}\cdot e^{\gamma \langle {x},{a+\bm 1_{d}}\rangle}\cdot \varphi(x).
\end{equation*}

Therefore, for $a\in\mathcal{A}$ and $i\in[d]$, we have
 \[
    \bm p_{a^{\oplus i}}(x) 
      =  e^{-2\gamma x_ia_i}e^{2\gamma^2a_i}\cdot\bm p_{a}(x)
      = ({1+q\cdot\phi_{a,i}(x)})\cdot \bm p_{a}(x),
 \]
 where $q =  \sqrt{e^{4\gamma^2}-1}$ and
 $\phi_{a,i}(x) =  \frac{1-e^{-2\gamma x_ia_i}e^{2\gamma^2a_i}}{\sqrt{e^{4\gamma^2}-1}}$. By using the Gaussian moment-generating function, we could verify that, for $i\neq j$, 
 \[
    \E_{\bm p_a}[{\phi_{a,i}(X)}] = 0,\quad \E_{\bm p_a}[{\phi_{a,i}(X)^2}] = 1, \text{ and } \E_{\bm p_a}[{\phi_{a,i}(X)\phi_{a,j}(X)} ]= 0,
 \]
 so that the condition~\ref{Cond:1} and condition~\ref{Cond:2} are satisfied. Here, notice that in the proof of Corollary~\ref{corollary:1}, we require that $|q\cdot\phi_{a,i}(x)| = C /n$ where $C$ is a constant, we could verify that since $\rho \le c \cdot \sqrt{s/n}$, then $\gamma \le c \sqrt{n}$ from the definition of $\gamma$, Then we have $|q\cdot\phi_{z,i}(x)|  \le c_0 |\gamma^2 -\gamma x_i| \le c_0/n$, which could verify the condition for corollary~\ref{corollary:1}. Also, by the choice of $\gamma$ and $\rho$, it is easy to verify that condition~\ref{Cond:4} also holds with:  
\[
\ell_2 (\mu(\bm p_a),\mu(\bm p_{a'})) 
    = {4\rho}\cdot\sqrt{ \frac{d_{ham} (a, a')}{\tau d}}.
    \]

Thus, all the conditions mentioned above have been verified. Then, we could finish the proof of our lower bound. Combining the result of corollary~\ref{corollary:2} and corollary~\ref{corollary:3}, we get the result below: 
\[
n^2  d\le c m n^2 q^2 (e^{n \epsilon^2}-1),
\]
where $c$ is a constant. Also, notice that $q^2=e^{4\gamma^2}-1 \leq 8\gamma^2$ holds since $\gamma \leq
1/2$, we could find a constant $c_0$, it follows that
\[
\rho^2 \ge c_0 \cdot \frac{s \cdot d }{m n \epsilon^2},
\]
From the choice of $\rho$, we could claim that $\rho \ge \Omega(\sqrt{\frac{sd}{mn\epsilon^2} }\wedge 1)$, then we could obtain our lower bound, which finished the proof. \hfill $\square$

\subsection{Proof of Theorem~\ref{thm2}}\label{Proof:2}
In the proof of Theorem~\ref{thm2}, we will design a mechanism to get an estimation of the parameter and then we obtain the upper bound of $\| \bm \mu-\hat{\bm \mu}\|_2^2 $. The overall mechanism is designed as follows: we first calculate the mean for $n$ data points on each machine. Then, we transform the Gaussian mean to Bernoulli mean according to the sign of the Gaussian mean motivated by the Algorithm 2 discussed in \citep{acharya2020general}, l-bit protocol for estimating product of Bernoulli family. 

Then, we could use the $\epsilon$-local differentially private mechanism to achieve mean estimation for the product of Bernoulli family in the federated learning setting. After obtaining the estimation, we could convert the estimated Bernoulli mean back to Gaussian mean estimation.

First, for each data point on the machine, it follows the distribution of $N(\bm \mu,\bm I_d)$ . Then for the mean on $i$-th machine, the mean $\bar{\bm  X_i}$ follows a distribution of $N(\bm \mu,1/n \bm I)$. Then, we could convert it to a Bernoulli variable $\bf Z$, where $Z_i = 1$ when $\bar{X_{ij}} > 0$ and $Z_i = -1$ when $\bar{X_{ij}} \le 0$. Then the mean of $\bm Z$, which denote as $\bm v$ is:
\begin{equation*}
    v_i = 2 \Pro(X_i > 0) -1 = \operatorname{Erf}\qty(\frac{\sqrt{n}\mu_i}{\sqrt{2}}),
\end{equation*}
for each coordinate of $\bm v$.
Suppose the estimation of $\bm v$ is denoted by $\hat{\bm v}$, then suppose the estimation $\hat{\bm \mu}$ is given by $\hat{\mu_i} = \frac{\sqrt{2}}{\sqrt{n}}\text{erf}^{-1}(\hat{v_i})$, we could find such relationship:
\begin{align}\label{eq:1}
     \| \bm \mu-\hat{\bm \mu}\|_2^2
        &= \sum_{i=1}^d |\mu_i-\hat{\mu}_i|^2
        = \frac{2}{n}\cdot\sum_{i=1}^d |\text{Erf}^{-1}(v_i)-\text{Erf}^{-1}(\hat{v}_i)|^2
        \leq c \cdot \frac{1}{n} \cdot \sum_{i=1}^d |v_i-\hat{v}_i|^2,
\end{align}
where $c$ is a constant. The last inequality comes from the Lipschitz condition of a Erf function. Then, we could get the upper bound for the Bonoulli mean estimation directly from Theorem 3 in \citep{acharya2020general}, where
\begin{equation*}\label{eq:2}
    \| \bm v-\hat{\bm v}\|_2^2 \le c \cdot \frac{d\cdot s}{m \epsilon^2},
\end{equation*}
where $\epsilon$ is the privacy parameter, $m$ is the number of machines. Combining the last two inequalities (\ref{eq:1}) and (\ref{eq:2}), we could get the upper bound for the mean Gaussian estimation:
\begin{equation*}
    \| \bm \mu-\hat{\bm \mu}\|_2^2 \le c \cdot \frac{ s\cdot d}{mn\epsilon^2},
\end{equation*}
which finished the proof.\hfill$\square$
\subsection{Proof of Theorem~\ref{thm: federated_estimation}}
It is not difficult to observe that the convergence rate would be the same as in the non-federated learning setting. We denote $L_n$ as the sample loss function and $L$ be the population level. In the estimation of $\bbeta$, $L(\bbeta) = \|\bm Y- \bm X \bbeta\|_2^2$ and $L_n$ is the sample version. In the estimation of $\bm \Theta_k$, $L(\bm \Theta_k)= \frac{1}{2} \bm \Theta_k^\top \Sigma \bm \Theta_k - \langle \bm e_j, \bm \Theta_k \rangle$. We start from the estimation of $\bbeta$ and the estimation of $\Theta$ is the same. In this proof, we use $n_0$ to refer the total number of samples $n_0 = m \cdot n$. Then, it holds that:

\begin{lemma}\label{lm:rscrsm}
		Under assumptions of Theorem~\ref{thm:ci2}, it holds that:
		\begin{align}\label{eq: rsc and rsm}
			{8\nu_s}\|\bbeta^t - \hat\bbeta\|_2^2 \leq \langle \nabla \L_n(\bbeta^t) - \nabla \L_n(\hat\bbeta), \bbeta^t - \hat\bbeta \rangle \leq {8\mu_s} \|\bbeta^t - \hat\bbeta\|_2^2.
		\end{align}
	\end{lemma}
\textbf{Proof:} From direct calculation, we could obtain that:
\begin{equation*}
    \langle \nabla \L_n(\bbeta^t) - \nabla \L_n(\hat\bbeta), \bbeta^t - \hat\bbeta \rangle = 2 (\bbeta^t - \hat\bbeta)^T \hat{\bm \Sigma} (\bbeta^t - \hat\bbeta) \le 2 \mu_{s + s^*}  \|\bbeta^t - \hat\bbeta\|_2^2 \le 2 \mu_{2s} \|\bbeta^t - \hat\bbeta\|_2^2
\end{equation*}
The last inequality is according to the choice of $s$ such that $s^* \le s$. Then, we also have $\mu_{2s} \le 4\mu_{s}$. Thus we have obtained the right hand side of the inequality. By a similar approach, we could also obtain the left hand side. 

\begin{lemma}\label{lm:contraction}
		Under assumptions of Theorem~\ref{thm:ci2}, it holds that there exists an absolute constant $\rho$ such that
		\begin{align}\label{eq: high-dim regression contraction}
			\L_n(\bbeta^{t+1}) - \L_n(\hat\bbeta) \leq \left(1 - \frac{\nu_s}{24 \mu_s}\right)\left(\L_n(\bbeta^{t}) - \L_n(\hat\bbeta)\right) + c_3 \left(\sum_{i \in [s]} \|\bm w^t_i\|_\infty^2 + \|\tilde{\bm w}^t_{S^{t+1}} \|_2^2\right),
		\end{align} 
		where $c_3$ is a constant number such that $c_3 = \max (\mu_s (72\cdot 8\mu_s +13), 68\mu_s + 2/3)$
	\end{lemma}
Notice that $w_i,\title{w}$ are injected from the NoisyHT algorithm. The proof of the above lemma follows from the result in Lemma 8.3 from \citep{cai2019cost}. Then, we could start the proof by iterating \eqref{eq: high-dim regression contraction} over $t$. Denote $\bm W_t = c_3\left(\sum_{i \in [s]} \|\bm w^t_i\|^2_\infty + \|\tilde {\bm w}^t_{S^{t+1}}\|_2^2\right)$ to obtain
	\begin{align}
		\L_{n_0}(\bbeta^T) - \L_{n_0}(\hat\bbeta) &\leq \left(1-\frac{\nu_s}{24 \mu_s}\right)^T\left(\L_{n_0}(\bbeta^0) - \L_{n_0}(\hat\bbeta)\right)  + \sum_{k=0}^{T-1}\left(1-\frac{1}{\rho L^2}\right)^{T-k-1}\bm W_k \notag\\
		&\leq  
		\left(1-\frac{\nu_s}{24 \mu_s}\right)^T 4\mu c_0^2 + \sum_{k=0}^{T-1}\left(1-\frac{\nu_s}{24 \mu_s}\right)^{T-k-1}\bm W_k.\label{eq: high-dim regression suboptimality upper bound}
	\end{align}
	The second inequality is a consequence of the upper inequality in \eqref{eq: rsc and rsm} and the $\ell_2$ bounds of $\bbeta^0$ and $\hat\bbeta$. We can also bound $\L_{n_0}(\bbeta^T) - \L_{n_0}(\hat\bbeta)$ from below by the lower inequality in \eqref{eq: rsc and rsm}:
	\begin{align}\label{eq: high-dim regression suboptimality lower bound}
		\L_{n_0}(\bbeta^T) - \L_{n_0}(\hat\bbeta) \geq \L_{n_0}(\bbeta^T) - \L_{n_0}(\bbeta^*) \geq 4 \nu_s \|\bbeta^T - \bbeta^*\|_2^2 - \langle \nabla \L_{n_0}(\bbeta^*), \bbeta^* - \bbeta^T \rangle.
	\end{align}
	Now \eqref{eq: high-dim regression suboptimality upper bound} and \eqref{eq: high-dim regression suboptimality lower bound} imply that, with $T = (\rho L^2)\log(8c_0^2Ln_0)$, 
	\begin{align}
		4 \nu_s \|\bbeta^T - \bbeta^*\|_2^2 &\leq  
		\|\nabla \L_{n_0}(\bbeta^*)\|_\infty \sqrt{s+s^*}\|\bbeta^* - \bbeta^T\|_2 + \frac{1}{n_0} + \sum_{k=0}^{T-1}\left(1-\frac{\nu_s}{24 \mu_s}\right)^{T-k-1}\bm W_k. \label{eq: high-dim regression fundamental inequality}\\
		&\le \|\nabla \L_{n_0}(\bbeta^*)\|_\infty \sqrt{s+s^*}\|\bbeta^* - \bbeta^T\|_2 + \frac{1}{n_0} + \frac{24\mu_s}{\nu_s} \max_k \bm W_k.
	\end{align}
Thus,
\begin{equation*}
    \|\bbeta^T - \bbeta^*\|_2^2 \le k \cdot \frac{s^* \log d}{n_0} + \frac{\mu_s}{\nu_s^2} \max_k \bm W_k.
\end{equation*}
In the above inequality, $k$ is a constant. Then, we could calculate the upper bound of $\W_k$. From the result of tail bound of Laplace random variables, we could find that with high probability that $\W_k \le c_4 s^2 \log^2 d \log (1/\delta) \log n^3 / n^2 \epsilon^2)$, where $c_4 = \max(\mu_s(9\mu_s +1/4), 17/16 \mu_s +1/96)$. Then, we have with high probability:
\begin{equation*}
      \|\bbeta^T - \bbeta^*\|_2^2 \le k \cdot \frac{s \log d}{n_0} + \frac{6c_4 \mu_s}{\nu_s^2} s^2 \log^2 d \log (1/\delta)  \log n_0^3 / n_0^2 \epsilon^2.
\end{equation*}
Similarly, we could obtain the same result for the estimation of $\hat{\bm \Theta}_k$, which finishes the proof. 
\subsection{Proof of Theorem~\ref{thm: confidence interval}}
The structure of the proof consist of three part, the first part is to show that our algorithm provides an $(\epsilon, \delta)$-differentially private confidence interval. In the second part, we will show that $\hat{\beta_k}$ is a consistent estimator of true $\beta_k$, which is unbiased. In the last part, we will show that the $(1-\alpha)$ confidence interval is asymptotically valid. Before we start the first part, let us first analyze $c_x$: \\
According to the assumptions of the theorem, we have learnt that for each row of $\bm X$, $\bm x\bm \Sigma^{-1/2}$ is sub-Gaussian with $\kappa = \|\bm \Sigma^{-1/2}\bm x\|_{\psi_2}$. Then according to the properties of sub-Gaussian random variables, we have: $\|\bm x\bm \Sigma^{-1/2}\|_{\infty} \le 3\sqrt{2\kappa^2 \log d}$ with probability $1-d^{-2}$. Then for each element of $x_i$, $i =1,2,\dots,d$, we have:
\[x_i =  \bm e_j^\top \bm x = \bm e_j^\top \bm \Sigma^{1/2} \bm \Sigma^{-1/2} \bm x\]
Thus,
\begin{align*}
    x_i \le \| \bm e_j^\top \bm \Sigma^{1/2} \|_1 \|\bm \Sigma^{-1/2} \bm x\|_{\infty} \le \|\bm \Sigma^{1/2}\|_2 \|\bm \Sigma^{-1/2} \bm x\|_{\infty} 
\end{align*}
Then, with probability $1-d^{-2}$, we have $x_i \le 3  \sqrt{2L\kappa^2 \log d}$. By a union bound, we could have with probability $1-d^{-1}$, $\|\bm x\|_{\infty} \le 3  \sqrt{2L\kappa^2 \log d}$. By the choice of $c_x$ in the theorem, we have $\|\bm x\|_{\infty} \le c_x$ with a high probability.\\
Then, we could verify that the confidence interval is $(\epsilon, \delta)$-differentially private. From \citep{cai2019cost}, we could obtain that the output $\hat{\bbeta^{\text{u}}}$ is $(\epsilon,\delta)$-DP. In a similar manner, we could also verify that the output $\hat{\bm \Theta}_k$ is also $(\epsilon,\delta)$-DP. Thus, for two adjacent data sets $(\bm X,\bm Y)$ and $(\bm X', \bm Y')$ which differ by one data $(\bm x_{ij}, y_{ij})$ and $(\bm x_{ij}',  y_{ij}')$, we have:
\begin{align*}
    | \frac{1}{n_0} (\hat{\bm \Theta}_k^\top  \bm x_{ij} \Pi_R(y_{ij}) - \hat{\bm \Theta}_k^\top  \bm x_{ij}  \bm x_{ij}^\top  \hat{\bbeta^{\text{u}}})|
    &\le \frac{1}{n_0} | \hat{\bm \Theta}_k^\top  \bm x_{ij} \Pi_R(y_{ij})| +\frac{1}{n_0}  | \hat{\bm \Theta}_k^\top  \bm x_{ij}  \bm x_{ij}^\top  \hat{\bbeta^{\text{u}}}| \notag\\
    &\le  \frac{1}{n_0} | \hat{\bm \Theta}_k^\top  \bm x_{ij}| |\Pi_R(y_{ij})| +\frac{1}{n_0}  |\hat{\bm \Theta}_k^\top  \bm x_{ij}  | |\bm x_{ij}^\top  \hat{\bbeta^{\text{u}}}| \notag \\
    &\le \frac{1}{n_0}\sqrt{s} c_1 c_x R + \frac{1}{n_0} s c_0 c_1 c_x^2
\end{align*}
Thus,
\begin{align*}
     &| \frac{1}{n_0} (\hat{\bm \Theta}_k^\top  \bm x_{ij} \Pi_R( y_{ij}) - \hat{\bm \Theta}_k^\top  \bm x_{ij}  \bm x_{ij}^\top  \hat{\bbeta^{\text{u}}})- \frac{1}{n_0} (\hat{\bm \Theta}_k^\top  \bm x_{ij}' \Pi_R(y_{ij}') - \hat{\bm \Theta}_k^\top  \bm x_{ij} ' \bm x_{ij}'^\top  \hat{\bbeta^{\text{u}}})|\notag\\
    &=| \frac{1}{n_0} (\hat{\bm \Theta}_k^\top  \bm x_{ij} \Pi_R(y_{ij}) - \hat{\bm \Theta}_k^\top  \bm x_{ij}  \bm x_{ij}^\top  \hat{\bbeta^{\text{u}}})|+| \frac{1}{n_0} (\hat{\bm \Theta}_k^\top  \bm x_{ij}' \Pi_R( y_{ij}') - \hat{\bm \Theta}_k^\top  \bm x_{ij}'  \bm x_{ij}'^\top  \hat{\bbeta^{\text{u}}})|\notag\\
     &\le \frac{2}{n_0}\sqrt{s} c_1 c_x R + \frac{2}{n} s c_0 c_1 c_x^2
\end{align*}
Denote $\Delta_1 =  \sqrt{s} c_1 c_x R + s c_0 c_1 c_x^2$. Thus, if $E_k$ follows $N(0,8 \Delta_1^2/n_0^2 \epsilon^2 \log(1.25/\delta))$, $\hat{\beta_j}$ is $(\epsilon,\delta)$-DP.
For the term $\hat{\bm \Theta}_k^\top \hat{\bm \Sigma} \hat{\bm \Theta}_k$, we could obtain that:
\begin{align*}
    \hat{\bm \Theta}_k^\top \hat{\bm \Sigma} \hat{\bm \Theta}_k = \frac{1}{n_0}\sum_{i=1}^n \hat{\bm \Theta}_k^\top \bm x_{ij} \bm x_{ij}^\top \hat{\bm \Theta}_k= \frac{1}{n_0}\sum_{i=1}^n (\hat{\bm \Theta}_k^\top \bm x_{ij})^2
\end{align*}
Thus, for two adjacent data sets $\bm X$ and $\bm X'$ differ by one data $\bm x_{ij}$ and $\bm x_{ij}'$, we have:
\begin{align*}
    |\hat{\bm \Theta}_k^\top \hat{\bm \Sigma} \hat{\bm \Theta}_k  - \hat{\bm \Theta}_k^\top \hat{\bm \Sigma'} \hat{\bm \Theta}_k | &\le \frac{1}{n_0} (\hat{\bm \Theta}_k^\top \bm x_{ij})^2 + \frac{1}{n_0} (\hat{\bm \Theta}_k^\top \bm x_{ij}')^2
\end{align*}
By Holder inequality and Cauchy inequality, we have $|\hat{\bm \Theta}_k^\top \bm x_{ij}| \le \sqrt{s} c_1 c_x$, thus we have:
\begin{align*}
     |\hat{\bm \Theta}_k^\top \hat{\bm \Sigma} \hat{\bm \Theta}_k  - \hat{\bm \Theta}_k^\top \hat{\bm \Sigma'} \hat{\bm \Theta}_k | &\le \frac{2}{n_0} (\sqrt{s} c_1 c_x)^2 = \frac{2}{n_0} s c_1^2 c_x^2 
\end{align*}
Denote $\Delta_2 = s c_1^2 c_x^2 $. Then, let $E'$ follows a Gaussian distribution of $N(0,8 \Delta_2^2/n^2 m^2 \epsilon^2 \log(1.25/\delta))$. We could claim that $\hat{\bm \Theta}_k^\top \hat{\bm \Sigma} \hat{\bm \Theta}_k+E'$ is $(\epsilon,\delta)$-differentially private.\\
We start the second part of the proof. First, with probability $1-k_0 \exp(-k_1 n_0)$, we have $\Pi_{R}(y_i) = y_i$ for each $i = 1,2,\dots,d$, so we could decompose $\hat{\beta_k}$ by the following approach:
\begin{align*}
    \hat{\beta_k} &= \hat{\beta_k^{\text{u}}} +  \frac{1}{n_0} \hat{\bm \Theta}_k^\top \bm X^\top (\bm X \bbeta + \bm W -\bm X \hat{\bbeta^{\text{u}}}) + E_k \notag\\
    &= \hat{\beta_k^{\text{u}}} +  \frac{1}{n_0} \hat{\bm \Theta}_k^\top \bm X^\top \bm X ( \bbeta - \hat{\bbeta^{\text{u}}}) +  \frac{1}{n_0} \hat{\bm \Theta}_k^\top \bm X^\top \bm W  + E_k \notag\\
    &= {\beta_k} +   (\hat{\bm \Theta}_k^\top \hat{\bm\Sigma}-\bm e_k) ( \bbeta - \hat{\bbeta^{\text{u}}}) +  \frac{1}{n_0} \hat{\bm \Theta}_k^\top \bm X^\top  \bm W  + E_k
\end{align*}
Thus, we have:
\begin{equation}\label{eq:eq1}
    \sqrt{n_0}(\hat{\beta_j} - \beta_j) =\underbrace{\sqrt{n_0} (\hat{\bm \Theta}_k^\top \hat{\bm\Sigma}-\bm e_k^\top) ( \bbeta - \hat{\bbeta^{\text{u}}})}_{\ref{eq:eq1}.1} +  \underbrace{\frac{1}{\sqrt{n_0}} \hat{\bm \Theta}_k^\top \bm X^\top \bm W }_{\ref{eq:eq1}.2} + \underbrace{\sqrt{n_0}  E_k}_{\ref{eq:eq1}.3}
\end{equation}
We will analyze the three terms in (\ref{eq:eq1}) one by one. For the first term, we could further decompose this term as:
\begin{align}\label{eq:eq2}
    \sqrt{n_0} (\hat{\bm \Theta}_k^\top \hat{\bm\Sigma}-\bm e_k^\top) ( \bbeta - \hat{\bbeta^{\text{u}}}) &= \sqrt{n_0} (\hat{\bm \Theta}_k^\top \hat{\bm\Sigma} -{\bm \Theta}_k^\top \hat{\bm\Sigma}+{\bm \Theta}_k^\top \hat{\bm\Sigma}-\bm e_k^\top) ( \bbeta - \hat{\bbeta^{\text{u}}}) \notag\\
    &= \sqrt{n_0} (\hat{\bm \Theta}_k^\top \hat{\bm\Sigma} -{\bm \Theta}_k^\top \hat{\bm\Sigma}) ( \bbeta - \hat{\bbeta^{\text{u}}}) +\sqrt{n_0} ({\bm \Theta}_k^\top \hat{\bm\Sigma}-\bm e_k^\top) ( \bbeta - \hat{\bbeta^{\text{u}}})
\end{align}
For the first term in (\ref{eq:eq2}), we could further decompose this term from $\hat{\bm \Sigma} =\frac{1}{mn}\sum_{i=1}^{m} \sum_{j=1}^{n} \bm x_{ij} \bm x_{ij}^\top$:
\begin{align} \label{eq:eq3}
     \sqrt{n_0} (\hat{\bm \Theta}_k^\top \hat{\bm\Sigma} -{\bm \Theta}_k^\top \hat{\bm\Sigma}) ( \bbeta - \hat{\bbeta^{\text{u}}}) 
     &={\sqrt{n_0}} (\hat{\bm \Theta}_k^\top-{\bm \Theta}_k^\top) \hat{\Sigma} (\bbeta - \hat{\bbeta^{\text{u}}}) \notag\\
     &\le {\sqrt{n_0}} \lambda_{s}(\hat{\Sigma}) \|\hat{\bm \Theta}_k-{\bm \Theta}_k\|_2 |\bbeta - \hat{\bbeta^{\text{u}}}\|_2 
\end{align}
In the last inequality, we use $\lambda_{s}$ to denote the largest s-restricted eigenvalue of the covariance matrix $\hat{\Sigma}$. 
From Theorem~\ref{thm: federated_estimation}, we could obtain that there exists a constant $c$ such that:
\begin{equation*}\label{eq:eq4}
    \|\bbeta - \hat{\bbeta^{\text{u}}}\|_2^2 \le c \cdot \sigma^2\left(\frac{s^*\log d}{n_0} + \frac{(s^*\log d)^2 \log(1/\delta)\log^3 n_0}{n_0^2\varepsilon^2}\right) 
\end{equation*}
Also, for the output $\hat{\bm \Theta}_k$, we could have the similar result:
\begin{equation*}\label{eq:eq5}
    \|\hat{\bm \Theta}_k- \bm \Theta_k\|_2^2 \le c \cdot \sigma^2\left(\frac{s^*\log d}{n_0} + \frac{(s^*\log d)^2 \log(1/\delta)\log^3 n_0}{n_0^2\varepsilon^2}\right) 
\end{equation*}
Combining (\ref{eq:eq4}) and (\ref{eq:eq5}), we could obtain that 
\begin{equation*}\label{eq:eq6}
    \sqrt{n_0} (\hat{\bm \Theta}_k^\top \hat{\bm\Sigma} -{\bm \Theta}_k^\top \hat{\bm\Sigma}) ( \bbeta - \hat{\bbeta^{\text{u}}}) = o \qty(\frac{s^* \log d}{\sqrt{n_0}} ) = o(1)
\end{equation*}
Then, we could focus on the second term of (\ref{eq:eq2}). We first introduce the following lemma:
\begin{lemma}{(Lemma 6.2 in \citep{javanmard2014confidence})}\label{lemma1}
For the vector ${\bm \Theta}_k^\top \hat{\bm\Sigma}-\bm e_k$. Denote $\kappa = \|\Sigma^{-1/2}\bm X_1\|_{\phi_2}$, then with probability $1-2d^{1-a^2/24e^2\kappa^4 L^2}$, we have:
\[\|{\bm \Theta}_k^\top \hat{\bm\Sigma}-\bm e_k\|_{\infty} \le a \sqrt{\frac{\log d}{n_0}}\]
\end{lemma}
Thus, for the second term of (\ref{eq:eq2}), we have:
\begin{align}\label{eq:eq7}
  \sqrt{n_0} ({\bm \Theta}_k^\top \hat{\bm\Sigma}-\bm e_j^\top) ( \bbeta - \hat{\bbeta^{\text{u}}}) &\le \sqrt{n_0} \|{\bm \Theta}_k^\top \hat{\bm\Sigma}-\bm e_k^\top\|_{\infty} \|\bbeta - \hat{\bbeta^{\text{u}}}\|_1 \notag \\
  &\le  k \sqrt{n_0} \sqrt{\frac{\log d}{n_0}}\sqrt{s^*} \|\bbeta - \hat{\bbeta^{\text{u}}}\|_2 \notag \\
  &\le k \cdot\sqrt{ {s^* \log d} } \cdot \sqrt{\frac{s^* \log d}{n_0}} = o(1)
\end{align}
Combine the result from (\ref{eq:eq6}), (\ref{eq:eq7}) to (\ref{eq:eq2}), we could obtain that the first term of (\ref{eq:eq1}) is $o(1)$. We could also analyze the third term of (\ref{eq:eq1}),$\sqrt{n_0}\bm E_k \sim N(0,8 \Delta_1^2 \log(1.25/\delta) /n_0 \epsilon^2 )$. Then, by the definition of $\Delta_1$, we have $8 \Delta_1^2 \log(1.25/\delta) /n_0 \epsilon^2 \sim \frac{{s^*}^2 \log^2 d \log(1.25/\delta)}{n_0 \epsilon^2} = o(1)$ from the assumption. Also, we notice that $\bm E' =N(0, c \cdot \frac{{s^*}^2 \log^2 d \log(1.25/\delta)}{n_0^2 \epsilon^2} )$. By the concentration of Gaussian distribution, we also have that $\bm E' = o(1)$. 

Finally, we analyze the term $\frac{1}{\sqrt{n_0}} \hat{\bm \Theta}_j^\top \bm X^\top \bm W$. From our definition, $W$ is sub-Gaussian random noise. Then, from the central limit theorem, we could conclude that:
\begin{equation*}
    \frac{1}{\sqrt{n_0}} \hat{\bm \Theta}_j^\top \bm X^\top \bm W \to N(0, \sigma^2 \hat{\bm \Theta}_j^\top \bm \hat{\Sigma} \hat{\bm \Theta}_j )
\end{equation*}

Thus, $ \sqrt{n_0}(\hat{\beta_j} - \beta_j) =  \frac{1}{\sqrt{n_0}} \hat{\bm \Theta}_j^\top \bm X^\top \bm W + \sqrt{n_0}\bm E_k \sim N(0, \sigma^2 \hat{\bm \Theta}_j^\top \bm \hat{\Sigma} \hat{\bm \Theta}_j + o(1))$. Also, from lemma~\ref{lemma:2}, we could claim that under our assumptions, $\hat{\sigma}^2 = \sigma^2 + o(1)$. We could get the result where with high probability, $\frac{ \sqrt{n_0}(\hat{\beta_j} - \beta_j)}{\hat{\sigma}\sqrt{\hat{\bm w_j}^\top \hat{\bm \Sigma} \hat{\bm w_j}}} \to N(0,1)$. 

Therefore, we could claim that $[\hat{\beta_j} - \Phi^{-1}(1-\alpha/2) \frac{\hat{\sigma}}{\sqrt{n_0}}\sqrt{\hat{\bm \Theta}_j^\top \hat{\bm \Sigma} \hat{\bm \Theta}_j} , \hat{\beta_j} + \Phi^{-1}(1-\alpha/2) \frac{\hat{\sigma}}{\sqrt{n_0}}\sqrt{\hat{\bm \Theta}_j^\top \hat{\bm \Sigma} \hat{\bm \Theta}_j}  ]$ is asymptotically $1-\alpha$ confidence interval for $\beta_j$. Therefore, we have finished the proof of theorem. \hfill$\square$ 

\subsection{Proof of Theorem~\ref{thm:ci2}}
The proof is similar to the proof of Theorem~\ref{thm: confidence interval}, the difference is that we need to consider the case where the privacy cost is not dominated by the statistical error. Then, for the proof of Theorem~\ref{thm:ci2}, we follow the proof of Theorem~\ref{thm: confidence interval} until (\ref{eq:eq1}). The analysis for the second term and the third term for (\ref{eq:eq1}) stays the same. On the other hand, for the first term of (\ref{eq:eq1}), we have:
We will analyze the three terms in (\ref{eq:eq1}) one by one. For the first term, in the same manner, we could decompose this term as:
\begin{align}\label{eq:eq10}
    \sqrt{n_0} (\hat{\bm \Theta}_k^\top \hat{\bm\Sigma}-\bm e_k^\top) ( \bbeta - \hat{\bbeta^{\text{u}}}) &= \sqrt{n_0} (\hat{\bm \Theta}_k^\top \hat{\bm\Sigma} -{\bm\Theta}_k^\top \hat{\bm\Sigma}+{\bm \Theta}_k^\top \hat{\bm\Sigma}-\bm e_k^\top) ( \bbeta - \hat{\bbeta^{\text{u}}}) \notag\\
    &= \sqrt{n_0} (\hat{\bm \Theta}_k^\top \hat{\bm\Sigma} -{\bm \Theta}_k^\top \hat{\bm\Sigma}) ( \bbeta - \hat{\bbeta^{\text{u}}}) +\sqrt{n} ({\bm \Theta}_k^\top \hat{\bm\Sigma}-\bm e_k^\top) ( \bbeta - \hat{\bbeta^{\text{u}}})
\end{align}
For the first term in (\ref{eq:eq10}), we could further decompose this term from $\hat{\bm \Sigma} =\frac{1}{n}\sum_{i=1}^n \bm x_i \bm x_i^\top$:
\begin{align} \label{eq:eq11}
     \sqrt{n_0} (\hat{\bm \Theta}_k^\top \hat{\bm\Sigma} -{\bm \Theta}_k^\top \hat{\bm\Sigma}) ( \bbeta - \hat{\bbeta^{\text{u}}}) 
     &={\sqrt{n_0}} (\hat{\bm \Theta}_k^\top-{\bm \Theta}_k^\top) \hat{\Sigma} (\bbeta - \hat{\bbeta^{\text{u}}}) \notag\\
     &\le {\sqrt{n_0}} \lambda_{s}(\hat{\Sigma}) \|\hat{\bm \Theta}_k-{\bm \Theta}_k\|_2 |\bbeta - \hat{\bbeta^{\text{u}}}\|_2 \notag\\
     &\le \frac{k \mu_s^2}{\nu_s^2} s^2 \log^2 d \log (1/\delta)\log^3 n_0/ n_0^{3/2} \epsilon^2
\end{align}
Thus, for the second term of (\ref{eq:eq10}), by Lemma~\ref{lemma1}, we have:
\begin{align}\label{eq:eq12}
  \sqrt{n_0} ({\bm \Theta}_k^\top \hat{\bm\Sigma}-\bm e_k^\top) ( \bbeta - \hat{\bbeta^{\text{u}}}) &\le \sqrt{n_0} \|{\bm \Theta}_k^\top \hat{\bm\Sigma}-\bm e_k^\top\|_{\infty} \|\bbeta - \hat{\bbeta^{\text{u}}}\|_1 \notag \\
  &\le  k \sqrt{n_0} \sqrt{\frac{\log d}{n_0}}\sqrt{s^*} \|\bbeta - \hat{\bbeta^{\text{u}}}\|_2 \notag \\
  &\le k \cdot\sqrt{ {s^* \log d} } \cdot \|\bbeta - \hat{\bbeta^{\text{u}}}\|_2
\end{align}
When the privacy cost is not dominated by the statistical error and also $s^* \log d/ \sqrt{n} = o(1)$, we can observe that the equation (\ref{eq:eq12}) has smaller convergence rate that (\ref{eq:eq11}). Then, combining (\ref{eq:eq11}) and (\ref{eq:eq12}), there exists a constant $k_1$, such that:
\begin{align*}
    \sqrt{n} (\hat{\bm \Theta}_k^\top \hat{\bm\Sigma}-\bm e_j^\top) ( \bbeta - \hat{\bbeta^{\text{u}}})  \le \frac{\gamma \mu_s^2}{\nu_s^2} \frac{s^2 \log^2 d \log (1/\delta)\log^3 n_0}{n_0^{3/2} \epsilon^2}
\end{align*}
Then, insert the result into (\ref{eq:eq1}), we have:
\begin{align}\label{eq:eq13}
    \sqrt{n_0}(\hat{\beta_j} - \beta_j ) =  O\qty(-  \frac{ \mu_s^2}{\nu_s^2} \frac{s^2 \log^2 d \log (1/\delta) \log^3 n_0}{n_0^{3/2} \epsilon^2}) + \frac{1}{\sqrt{n_0}} \hat{\bm \Theta}_k^\top \bm X^\top \bm W +\sqrt{n_0}  E_k
\end{align}
Notice that for the first term on the right hand, the constant could be set to 1 because it comes from the tail bound of Laplace random variable. From the result in (\ref{eq:eq13}), we could also apply the central limit theorem to show that the second term is asymptotically Gaussian, notice that in the right hand side, the second term and the third term asymptotically follows a distribution of $N(0, \sigma^2 \hat{\bm \Theta}_k^\top \hat{\bm \Sigma} \hat{\bm \Theta}_k+ \frac{8\Delta_1^2 \log(1/\delta)}{n_0\epsilon^2})$. Also, by the concentration of Gaussian distribution, we have with high probability, $E' \le \frac{8\Delta_1^2 \log(1/\delta)}{n_0\epsilon^2}$.Thus, the privacy conditions are satisfied. Therefore, we have:
\begin{equation*}
     \sqrt{n_0}[\hat{\beta_j} - \beta_j -  \sqrt{n_0} ({\bm \Theta}_k^\top \hat{\bm\Sigma}-\bm e_k^\top) ( \bbeta - \hat{\bbeta^{\text{u}}})] / \sqrt{\sigma^2 \hat{\bm \Theta}_k^\top \hat{\bm \Sigma} \hat{\bm \Theta}_k+ \frac{8\Delta_1^2 \log(1/\delta)}{n_0\epsilon^2}}\sim N(0,1)
\end{equation*}
Then, also by our assumptions and the result in Lemma~\ref{lemma:2}, we could claim that $\sigma^2 = \hat{sigma}^2 + o(1)$. Thus, finally, the confidence interval is given by:	
\begin{align*}
   	J_j(\alpha) = \biggl[\hat{\beta_j} -  \frac{\gamma \hat{\mu_s}^2}{\hat{\nu_s}^2} \frac{s^2 \log^2 d \log (1/\delta)\log^3 n_0}{n_0^{2} \epsilon^2})- \Phi^{-1}(1-\alpha/2) \frac{\sigma}{\sqrt{n_0}}\sqrt{\hat{\bm \Theta}_k^\top \hat{\bm \Sigma} \hat{\bm \Theta}_k+ \frac{8\Delta_1^2 \log(1/\delta)}{n_0\epsilon^2}} , \notag\\\hat{\beta_j} +   \frac{\gamma \hat{\mu_s}^2}{\hat{\nu_s}^2} \frac{s^2 \log^2 d \log (1/\delta)\log^3 n_0}{n_0^{2} \epsilon^2})+\Phi^{-1}(1-\alpha/2) \frac{\sigma}{\sqrt{n_0}}\sqrt{\hat{\bm \Theta}_k^\top \hat{\bm \Sigma} \hat{\bm \Theta}_k+ \frac{8\Delta_1^2 \log(1/\delta)}{n_0\epsilon^2}} \biggr] 
\end{align*}
which finishes our proof. 
\subsection{Proof of Theorem~\ref{thm: bootstrap_simu}}
Let us first show that our algorithm is $\epsilon, \delta$ private. The major proof lies in the choice of noise level $B_3$. To decompose, we can find: 
\begin{align*}
    \hat{\bm \Theta}\frac{1}{\sqrt{m}} \sum_{i=1}^m e_i \sqrt{n}(\bm g_i - \bar{\bm g}) = \hat{\bm \Theta}\frac{1}{\sqrt{m}} \sum_{i=1}^m e_i \sqrt{n} \bm g_i - \hat{\bm \Theta}\frac{\sqrt{n}}{\sqrt{m}} \qty(\sum_{i=1}^m e_i) \bar{\bm g}
\end{align*}
Then, suppose in an adjacent data set, the different data in denoted as $(\bm x_{ij}, y_{ij})$ and $(\bm x'_{ij}, y'_{ij})$. Then, we calculate:
\begin{align*}
    &\norm{(\hat{\bm \Theta}\frac{1}{\sqrt{m}} \sum_{i=1}^m e_i \sqrt{n} \bm g_i - \hat{\bm \Theta}\frac{\sqrt{n}}{\sqrt{m}} \qty(\sum_{i=1}^m e_i) \bar{\bm g}) -(\hat{\bm \Theta}\frac{1}{\sqrt{m}} \sum_{i=1}^m e_i \sqrt{n} \bm g'_i - \hat{\bm \Theta}\frac{\sqrt{n}}{\sqrt{m}} \qty(\sum_{i=1}^m e_i) \bar{\bm g'}) }_{\infty}\\
    &\le \norm{\hat{\bm \Theta}\frac{\sqrt{n}}{\sqrt{m}} e_i (\bm g_i -\bm g'_i) - \hat{\bm \Theta}\frac{\sqrt{n}}{\sqrt{m}} \qty(\sum_{i=1}^m e_i) (\bar{\bm g}-\bar{\bm g'})}_{\infty}\\
    &\le \norm{\hat{\bm \Theta}}_{\max}\norm{\frac{\sqrt{n}}{\sqrt{m}} e_i (\bm g_i -\bm g'_i) - \frac{\sqrt{n}}{\sqrt{m}} \qty(\sum_{i=1}^m e_i) (\bar{\bm g}-\bar{\bm g'})}_{\infty}\\
    &\le  (\norm{\hat{\bm \Theta}- \bm \Theta }_{\max}+ \norm{\bm \Theta}_{\max})\norm{\frac{\sqrt{n}}{\sqrt{m}} e_i (\bm g_i -\bm g'_i) - \frac{\sqrt{n}}{\sqrt{m}} \qty(\sum_{i=1}^m e_i) (\bar{\bm g}-\bar{\bm g'})}_{\infty}\\
    &\le (\norm{\hat{\bm \Theta}- \bm \Theta }_{1}+ \norm{\bm \Theta}_{2})\norm{\frac{\sqrt{n}}{\sqrt{m}} e_i (\bm g_i -\bm g'_i)}_{\infty} +\norm{ \frac{\sqrt{n}}{\sqrt{m}} \qty(\sum_{i=1}^m e_i) (\bar{\bm g}-\bar{\bm g'})}_{\infty})\\
    &\le  (o(1)+ L)\frac{\sqrt{n}}{\sqrt{m}} \sqrt{\log m} \norm{(\bm g_i -\bm g'_i)}_{\infty}  +  \frac{\sqrt{n}}{\sqrt{m}} m\sqrt{\log m}\norm{\bar{\bm g}-\bar{\bm g'}}_{\infty})\\
    &\le L \frac{4\sqrt{\log m }}{\sqrt{mn}} \norm{\bm x_{ij}(\pi_R(y_{ij}) - \bm x_{ij} \hat{\bbeta})}_{\infty}\\
    &\le  L \frac{4\sqrt{\log m }}{\sqrt{mn}}c_x(R+c_0c_x\sqrt{s^*})
\end{align*}
Thus, the privacy could be guaranteed. Then, let us start the proof of consistency. Throughout the proof, we define $n_0 = m\cdot n$, $(\bm X , \bm Y)$ be the whole data set where $\bm X \in \mathbb{R}^{n_0 * d}$ and $\bm Y \in \mathbb{R}^{d}$.  
$U' =\max_{ k \in G} \hat{\bm \Theta}\frac{1}{\sqrt{m}} \sum_{i=1}^m e_i \sqrt{n}(\bm g_i - \bar{\bm g})$.
Let us define another multiplier bootstrap statistic:
\begin{equation*}
    U^* = \max_{k \in G} \frac{1}{mn} \sum_{i = 1}^m \sum_{j=1}^n \Sigma^{-1} \bm x_{ij} (y_{ij} - \bm x_{ij} \bbeta) e_{ij}, 
\end{equation*}
where $e_{ij}$ are all standard Gaussian variables. At the same time, we also define:
\begin{equation*}
    M_0 = \max_{k \in G} \frac{1}{mn} \sum_{i = 1}^m \sum_{j=1}^n \Sigma^{-1} \bm x_{ij} (y_{ij} - \bm x_{ij} \bbeta) 
\end{equation*}
The proof consists three major steps, we start from the first step and measure $\sup_{\alpha \in (0,1)} |\Pro(M_0 \le C_{U^*}(\alpha)) - \alpha | $. This measurement is quite straightforward, we could apply Theorem 3.1 from \citep{chernozhukov2013gaussian}. However, we need to verify Corollary 2.1 from \citep{chernozhukov2013gaussian}. Notice that for any $k$, $\E [\Theta_k^T \bm x_{ij} (y_{ij} - \bm x_{ij} \bbeta)]^2 = \sigma^2 \Theta_k^T \Sigma \Theta_k \ge \sigma^2/L$. Also, it is not difficult to verify that $\Theta_k^T \bm x_{ij} (y_{ij} - \bm x_{ij} \bbeta)$ is sub-exponential. Since from assumption D1, we have $\bm x_{ij}$ is sub-Gaussian and from the linear model, we know that $(y_{ij} - \bm x_{ij} \bbeta)$ is also, sub-Gaussian. Then, the condition could be verified. Thus, by applying Theorem 3.1 and also under the condition where there exists a constant $k, k_0, k_1$ such that $\log^7(dmn)/mn \le \frac{1}{(mn)^{k}}$we could have:
\begin{align}
    \sup_{\alpha \in (0,1)} |\Pro(T_0 \le C_{U^*}(\alpha)) - \alpha | &\le k_0 \cdot \frac{1}{(mn)^{k_1}} + k_2 v^{1/3} (\max(1,\log(d/v)))^{2/3} + P(\Delta > v) \notag\\
    &\le k_2 v^{1/3} (\max(1,\log(d/v)))^{2/3} + P(\square > v) + o(1) \label{eq:eq19},
\end{align}
where $\square$ represents the maximum element between the two matrix $\Omega_1$ and $\Omega_2$, denote as $\|\Omega_1 - \Omega_2\|_{\max}$, where $\Omega_1$ and $\Omega_2$ are defined as:
\begin{equation*}
    [\Omega_1]_{k,l} = \frac{1}{mn}\sum_{i=1}^m \sum_{j=1}^n \Theta_k^\top \bm x_{ij} (y_{ij} - \bm x_{ij}^T \bbeta) \Theta_l 
\end{equation*}
and 
\begin{equation*}
    \Omega_2 = \sigma^2 \Theta
\end{equation*}
Then, from Corollary 3.1 in \citep{chernozhukov2013gaussian} and Lemma E.2 in \citep{yu2022distributed}, we could verify that $\|\Omega_1 - \Omega_2\|_{\max} = O(\sqrt{\frac{\log d}{n_0}} + \frac{\log^2(d n_0) \log d}{n_0})$. With a proper choice of $v$, e.g, there exists a constant $\kappa$ and let $v = (\sqrt{\frac{\log d}{n_0}} + \frac{\log^2(d n_0) \log d}{n_0})^{1-\kappa} $, we have $ k_2 v^{1/3} (\max(1,\log(d/v)))^{2/3} + P(\square > v) = o(1)$. Next, we would like to associate $M$ with $M_0$. Similarly, from Theorem 3.2 in \citep{chernozhukov2013gaussian} and (\ref{eq:eq19}), we could have:
\begin{equation*}
    \sup_{\alpha \in (0,1)} |\Pro(M \le C_{U^*}(\alpha)) - \alpha | \le  o(1) + v_1 \sqrt{max(1,\log(d/v_1))} + \Pro(\|M-M_0\| > v_1),
\end{equation*}
From the definition of $M$ and $M_0$, we have: 
\begin{equation*}
    \sqrt{n_0}(M- M_0) = \max_{1 \le k \le d} \frac{1}{\sqrt{n_0}} |(\hat{\bm \Theta_k}^\top - \bm \Theta_k^\top) \bm X^\top \bm W|
\end{equation*}
Then, for any $k$ in $1,\dots,d$, by Holder inequality and Cauchy–Schwarz inequality, we have:
\begin{align*}
    \frac{1}{\sqrt{n_0}}|(\hat{\bm w_k}^\top - \bm w_k^\top) \bm X^\top \bm W|
    \le \|\hat{\bm w_k}^\top - \bm w_k^\top\|_1 \|\frac{1}{\sqrt{n_0}}\bm X^\top \bm W\|_{\infty}
    \le \sqrt{s^*} \|\hat{\bm w_k}^\top - \bm w_k^\top\|_2 \|\frac{1}{\sqrt{n_0}}\bm X^\top \bm W\|_{\infty}
\end{align*}
On one hand, from previous proof, we obtain that $\|\hat{\bm w_k}^\top - \bm w_k^\top\|_2 = c \cdot \sqrt{\frac{s^* \log d}{mn}}$ when the privacy cost is dominated by statistical error uniformly for $k$. On the other hand, by the fact that $\bm \Sigma$ have bounded maximum eigenvalue and traditional linear regression model, we could apply Bernstein inequality and also obtain that $\|\frac{1}{\sqrt{n_0}}\bm X^\top \bm W\|_{\infty}$ is $O(\sqrt{\frac{\log d}{n_0}})$. Combine these two results, we could claim that there exist constants $k_0$ such that:
\begin{equation*}
    \frac{1}{\sqrt{n_0}}|(\hat{\bm \Theta_k}^\top - \bm \Theta_k^\top) \bm X^\top \bm W| = k_0 \cdot (s^* \log d /\sqrt{n_0})
\end{equation*}
uniformly for all $k$, then we can choose $v_1$ properly such that $\sup_{\alpha \in (0,1)} |\Pro(M \le C_{U^*}(\alpha)) - \alpha | = o(1)$.
At last, we need to relate $U^*$ with $U$. Our major goal is to prove that $C_U(\alpha)$ and $C_{U^*}$ are close to each other for any $\alpha \in (0,1)$. We first associate $U$ with $U'$. From the design of private max algorithm, from Lemma 3.4 in \citep{cai2019cost}, suppose $l_1$ is the element chosen from $U'$ and $l_2$ is from $U$ without noise injection, we use $w$ to represent the noise injected when we pick the largest value privately, we find that, for any $c>0$:
\begin{equation*}
    l_2^2 \le l_1^2 \le (1+c) l_2^2 + 4(1+1/c) \|w\|_{\infty}^2
\end{equation*}
From Lemma A.1 in \citep{cai2019cost}, we can verify that there exists constant $k_0, k_1$ such that $\|w\|_{\infty}^2 \le k_0 \cdot \frac{s^* \log^4 d \log m }{n_0}$. When we choose $c = o(1)$, e.g, $c = k_1 \frac{s^* \log d}{n_0}$, then from the conditions, we could claim that $l_1 = l_2 + o(1)$, also notice that the scale of noise we injected is small, it is easy to verify that $U = U' + o(1)$. The following discussions will be between $U'$ and $U^*$. Denote $\ominus$ as the symmetric difference, then we have:
\begin{align}
    &\Pro({T \le C_U(\alpha)} \ominus {T \le C_{U^*}(\alpha)}) \notag\\
    &\le 2\Pro(C_{U^*}(\alpha - \pi(u)) < T \le C_{U^*}(\alpha + \pi(u))) + \Pro( C_{U^*}(\alpha - \pi(u)) > C_U(\alpha)) + \Pro( C_{U^*}(\alpha + \pi(u)) < C_U(\alpha))\label{eq:eq20}
\end{align}
For the first term in (\ref{eq:eq20}), define $\pi(u) = u^{1/3} \max(1,\log(d/u))^{2/3}$, then exist a constant $k_0$, such that:
\begin{equation*}
    \Pro(C_{U^*}(\alpha - \pi(u)) < M \le C_{U^*}(\alpha + \pi(u))) \le \Pro(M \le C_{U^*}(\alpha + \pi(u)))- \Pro(M \le C_{U^*}(\alpha - \pi(u))) \le k \cdot \pi(u) + o(1)
\end{equation*}
Then, for the second term and third term in (\ref{eq:eq20}), from Lemma 3.2 in \citep{chernozhukov2013gaussian}, we have:
\begin{equation*}
    \Pro( C_{U^*}(\alpha - \pi(u)) > C_U(\alpha)) + \Pro( C_{U^*}(\alpha + \pi(u)) < C_U(\alpha)) \le 2 \Pro(\|\Omega_1 - \Omega_3\|_{\max} > u),
\end{equation*}
where $\Omega_3$ is defined as:
\begin{equation*}
    [\Omega_3]_{k,l} = \frac{1}{m}\sum_{i=1}^m n \hat{\bm \Theta_k}(\bm g_i - \bar{\bm g})(\bm g_i - \bar{\bm g})^\top \hat{\bm \Theta_l},
\end{equation*}
and $\Omega_1$ is defined the same as we defined before. Then, our major focus is to analyze $\|\Omega_1 - \Omega_3\|_{\max}$, by triangle inequality, we have $\|\Omega_1 - \Omega_3\|_{\max} \le \|\Omega_1 - \Omega_2\|_{\max} + \|\Omega_3 - \Omega_2\|_{\max}$. Since we have analyzed $\|\Omega_1 - \Omega_2\|_{\max}$ before, we will focus on $\|\Omega_3 - \Omega_2\|_{\max}$. 
\begin{align}
    \|\Omega_3 - \Omega_2\|_{\max} \le \norm{\frac{1}{m}\sum_{i=1}^m n \hat{\bm \Theta}(\bm g_i - \bar{\bm g})(\bm g_i - \bar{\bm g})^\top \hat{\bm \Theta} - \sigma^2 \hat{\bm \Theta}\Sigma \hat{\bm \Theta}}_{\max} + \|\sigma^2 \hat{\bm \Theta}\Sigma \hat{\bm \Theta} - \sigma^2 \bm \Theta\|_{\max} \label{eq:eq21}
\end{align}
We will analyze the two terms separately. We start from the second term in (\ref{eq:eq21}), we have:
\begin{align*}
    &\|\hat{\bm \Theta}\Sigma \hat{\bm \Theta} -  \bm \Theta\|_{\max} \notag\\
    &\le \|(\hat{\bm \Theta}-\bm \Theta + \bm \Theta)\Sigma (\hat{\bm \Theta}-\bm \Theta + \bm \Theta)-  \bm \Theta\|_{\max} \notag\\
    &\le \|\hat{\bm \Theta}-\bm \Theta\|_1^2 \|\bm \Sigma\|_{\max} + 2\|\hat{\bm \Theta}-\bm \Theta\|_1 \notag\\
    &\le k_0 \frac{{s^*}^2 \log d}{n_0} + k_1 s^*\sqrt{\frac{ \log d}{n_0}}
\end{align*}
On the other hand, for the first term in (\ref{eq:eq21}),notice that:
\begin{equation}\label{eq:eq22}
    \frac{1}{m}\sum_{i=1}^m n \hat{\bm \Theta}(\bm g_i - \bar{\bm g})(\bm g_i - \bar{\bm g})^\top \hat{\bm \Theta} = \frac{1}{m}\sum_{i=1}^m n \hat{\bm \Theta}\bm g_i  \bm g_i^\top \hat{\bm \Theta} - n \hat{\bm \Theta} \bar{\bm g} \bar{\bm g}^\top \hat{\bm \Theta}^\top
\end{equation}
Denote the data set on the $i$-th local machine as $(\bm X_i, \bm Y_i)$ and in the linear model, the random noise as $\bm W_i$. Also, we can further decompose the first term by:
\begin{align}
    &\frac{1}{m}\sum_{i=1}^m n \bm g_i  \bm g_i^\top \notag\\
    &= \frac{1}{m}\sum_{i=1}^m n [\frac{\bm X_i^\top W_i + \bm X_i^\top(\bbeta-\hat{\bm \beta})}{n}]  [\frac{\bm X_i^\top W_i + \bm X_i^\top(\bbeta-\hat{\bm \beta})}{n}]^\top \notag\\
    &= \frac{1}{m}\sum_{i=1}^m n [\frac{\bm X_i^\top W_i }{n}]  [\frac{\bm X_i^\top W_i }{n}]^\top + \frac{1}{m}\sum_{i=1}^m n [\frac{\bm X_i^\top(\bbeta-\hat{\bm \beta})}{n}]  [\frac{ \bm X_i^\top(\bbeta-\hat{\bm \beta})}{n}]^\top + \frac{2}{m}\sum_{i=1}^m n [\frac{ \bm X_i^\top(\bbeta-\hat{\bm \beta})}{n}]  [\frac{\bm X_i^\top W_i }{n}]^\top \label{eq:eq23}
\end{align}
Then, for the equation (\ref{eq:eq22}), we have:
\begin{align}
    &\|\frac{1}{m}\sum_{i=1}^m n \hat{\bm \Theta}(\bm g_i - \bar{\bm g})(\bm g_i - \bar{\bm g})^\top \hat{\bm \Theta} - \sigma^2 \hat{\bm \Theta}\Sigma \hat{\bm \Theta}\|_{\max} \notag\\
    &\le \|\hat{\bm \Theta}\|_{\max}\|\frac{1}{m}\sum_{i=1}^m n (\bm g_i - \bar{\bm g})(\bm g_i - \bar{\bm g})^\top \hat{\bm \Theta} - \sigma^2 \Sigma\hat{\bm \Theta} \|_{\max} \notag \\
    &\le \|\hat{\bm \Theta}\|_{\max}^2 \|\frac{1}{m}\sum_{i=1}^m n (\bm g_i - \bar{\bm g})(\bm g_i - \bar{\bm g})^\top - \sigma^2 \Sigma \|_{\max} \label{eq:eq24}
\end{align}
And, we could insert (\ref{eq:eq23}) into (\ref{eq:eq24}), 
\begin{align}
   & \|\frac{1}{m}\sum_{i=1}^m n (\bm g_i - \bar{\bm g})(\bm g_i - \bar{\bm g})^\top - \sigma^2 \Sigma \|_{\max}\notag\\
   &\le \|\frac{1}{m}\sum_{i=1}^m n [\frac{\bm X_i^\top W_i }{n}]  [\frac{\bm X_i^\top W_i }{n}]^\top - \sigma^2 \Sigma \|_{\max} + n\|\bar{\bm g} \bar{\bm g}^\top\|_{\max} +\|\frac{1}{m}\sum_{i=1}^m n [\frac{\bm X_i^\top(\bbeta-\hat{\bm \beta})}{n}]  [\frac{ \bm X_i^\top(\bbeta-\hat{\bm \beta})}{n}]^\top\|_{\max}\notag\\&+ \|\frac{2}{m}\sum_{i=1}^m n [\frac{ \bm X_i^\top(\bbeta-\hat{\bm \beta})}{n}]  [\frac{\bm X_i^\top W_i }{n}]^\top\|_{\max}\label{eq:eq18}
\end{align}
We will analyze the four terms in (\ref{eq:eq18}) one by one. For the first term, it is quite simple, from the proof of Lemma F.2 in \citep{yu2022distributed}, we have the first term is $O_p (\sqrt{\frac{\log d}{m}}+ \frac{\log^2(dm)\log d}{m})$. For the second term, we have:
\begin{align*}
    &n\|\bar{\bm g} \bar{\bm g}^\top\|_{\max} 
    \le n\|\bar{\bm g}\|_{\infty}^2 = n\|\frac{1}{n_0}\bm X^\top (\bm Y-\bm X\hat{\bbeta})\|_{\infty}^2
\end{align*}
Also, we have:
\begin{align}
    &\|\frac{1}{n_0}\bm X^\top (\bm Y-\bm X\hat{\bbeta})\|_{\infty} \notag\\
    &\le \|\frac{1}{n_0}\bm X^\top (\bm Y-\bm X{\bbeta})\|_{\infty} +\|\frac{1}{n_0}\bm X^\top \bm X(\hat{\bbeta}- {\bbeta})\|_{\infty} \notag\\
    &\le \|\frac{1}{n_0} \bm X^\top \bm W\|_{\infty} + \|(\hat{\Sigma} -\Sigma) (\hat{\bbeta}- {\bbeta})\|_{\infty} + \|\Sigma\|_{\max} \|\hat{\bbeta}- {\bbeta}\|_1 \notag \\
    &\le k_0(\sqrt{\frac{\log d}{n_0}}) + k_1(\sqrt{\frac{\log d}{n_0}}) \|\hat{\bbeta}- {\bbeta}\|_1 + \|\hat{\bbeta}- {\bbeta}\|_1 \notag \\
    &\le k_0 \sqrt{\frac{\log d}{n_0}} + k_1 \frac{s^* \log d}{n_0} + k_2 s^*\sqrt{\frac{ \log d}{n_0 }} \label{eq:eq27}
\end{align}
Thus, for the second term, we can obtain that $n\|\bar{\bm g} \bar{\bm g}^\top\|_{\max} \le k_0 {s^*}^2 \log d/m + k_1 {s^*}^2 \log^2 d/m^2 n$. For the third term, we have:
\begin{align}
    &\|\frac{1}{m}\sum_{i=1}^m n [\frac{\bm X_i^\top \bm X_i (\bbeta-\hat{\bm \beta})}{n}]  [\frac{ \bm X_i^\top \bm X_i (\bbeta-\hat{\bm \beta})}{n}]^\top\|_{\max} \notag\\
    &\le\frac{1}{m}\sum_{i=1}^m n  \|[\frac{\bm X_i^\top \bm X_i (\bbeta-\hat{\bm \beta})}{n}]  [\frac{ \bm X_i^\top \bm X_i (\bbeta-\hat{\bm \beta})}{n}]^\top\|_{\max} \notag\\
    &\le \frac{1}{m}\sum_{i=1}^m n   \|[\frac{\bm X_i^\top \bm X_i (\bbeta-\hat{\bm \beta})}{n}] \|_{\infty}^2 \notag\\
    &\le \frac{1}{m}\sum_{i=1}^m n   (\|\hat{\Sigma_i} - \Sigma\|_{\max} + \|\Sigma\|_{\max})^2 \|\bbeta-\hat{\bm \beta}\|_1^2 \notag\\
    &\le \frac{1}{m}\sum_{i=1}^m 2n   (\|\hat{\Sigma_i} - \Sigma\|_{\max}^2 + \|\Sigma\|_{\max}^2) \|\bbeta-\hat{\bm \beta}\|_1^2 \notag\\
    &\le \frac{1}{m}\sum_{i=1}^m 2n  (O(\sqrt{\frac{\log d}{n}})+O(1)) \|\bbeta-\hat{\bm \beta}\|_1^2 \notag\\
    &\le k_0 {s^*}^2 \frac{\log d}{m}\label{eq:eq25}
\end{align}
For the fourth term, we could apply Cauchy-Schwarz inequality, which give us the result:
\begin{align}
    &\|\frac{2}{m}\sum_{i=1}^m n [\frac{ \bm X_i^\top\bm X_i(\bbeta-\hat{\bm \beta})}{n}]  [\frac{\bm X_i^\top W_i }{n}]^\top\|_{\max} \notag\\
    &\le \frac{2}{m}\sum_{i=1}^m n \| [\frac{ \bm X_i^\top\bm X_i(\bbeta-\hat{\bm \beta})}{n}]  [\frac{\bm X_i^\top W_i }{n}]^\top\|_{\max} \notag \\
    &\le \frac{2}{m}\sum_{i=1}^m n \|\frac{ \bm X_i^\top\bm X_i(\bbeta-\hat{\bm \beta})}{n}\|_{\infty} \|\frac{\bm X_i^\top W_i }{n}\|_{\max} \notag\\
    &\le \frac{2}{m}\sum_{i=1}^m n \|\frac{ \bm X_i^\top\bm X_i}{n}\|_{\max} \|\bbeta - \hat{\bbeta}\|_{1} \|\frac{\bm X_i^\top W_i }{n}\|_{\infty} \notag\\
    &\le k_0 n \cdot \frac{s^* \log d}{n_0} \notag\\
    &\le k_0 \frac{s^* \log d}{m} \label{eq:eq26}
\end{align}

We could combine the result in (\ref{eq:eq27}), (\ref{eq:eq25}), (\ref{eq:eq26}) and insert into (\ref{eq:eq18}) and into (\ref{eq:eq24}). We could finally get the first term of (\ref{eq:eq21}) has an order of $O(\sqrt{\frac{\log d}{n_0}} + \frac{s^* \log d}{n_0} + \frac{{s^*}^2 \log d}{m})$. Insert this result into (\ref{eq:eq20}), when $u$ is chosen properly, we could verify that $\sup_{\alpha \in (0,1)} |\Pro(T \le C_U(\alpha)) - \alpha| = o(1)$, which finishes the proof. 
\subsection{Proof of Theorem~\ref{thm:estimate_diffbeta}}
The proof of theorem~\ref{thm:estimate_diffbeta} is quite straight forward. We could decompose true $\bbeta_i = \bm u +\bm v_i$. Then, $\hat{\bbeta} = \hat{\bm u} +\hat{\bm v_i}$. Thus, from the result of estimation, we could get the result that:
\begin{align*}
    \|\bm u - \hat{\bm u }\|_2^2 \le  c_0 \frac{s_0 \log d}{mn} + c_2 \frac{{s_0}^2 \log d^2 \log(1/\delta)\log^3 mn}{m^2 n^2 \epsilon^2} ,
\end{align*}
and
\begin{align*}
    \|\hat{\bm v_i} - \bm v_i \|_2^2 \le  c_1\frac{s_1\log d}{n} + c_3 \frac{{s_1}^2 \log d^2 \log(1/\delta)\log^3 n}{n^2 \epsilon^2}
\end{align*}
Also, combing the above two results with the inequality that $\|\hat{\bbeta_i} - \bbeta_i\|_2 \le \|\hat{\bm v_i} - \bm v_i \|_2  + \|\bm u - \hat{\bm u }\|_2$ gives the proof of theorem~\ref{thm:estimate_diffbeta}. \hfill$\square$
\subsection{Proof of Theorem~\ref{thm:ci3}}
The proof of Theorem~\ref{thm:ci3} follows the proof of  and Theorem~\ref{thm: confidence interval} and Theorem~\ref{thm:ci2}. We follow the proof of Theorem~\ref{thm: confidence interval} until (\ref{eq:eq1}). The analysis for the second term and the third term for (\ref{eq:eq1}) stays the same. 
We will analyze the three terms in (\ref{eq:eq1}) one by one. For the first term, in the same manner, we could decompose this term as:
\begin{align}\label{eq:eq14}
    \sqrt{n} (\hat{\bm \Theta}_j^\top \hat{\bm\Sigma}-\bm e_j^\top) ( \bbeta - \hat{\bbeta^{\text{u}}}) &= \sqrt{n} (\hat{\bm \Theta}_j^\top \hat{\bm\Sigma} -{\bm \Theta}_j^\top \hat{\bm\Sigma}+{\bm \Theta}_j^\top \hat{\bm\Sigma}-\bm e_j^\top) ( \bbeta - \hat{\bbeta^{\text{u}}}) \notag\\
    &= \sqrt{n} (\hat{\bm \Theta}_j^\top \hat{\bm\Sigma} -{\bm \Theta}_j^\top \hat{\bm\Sigma}) ( \bbeta - \hat{\bbeta^{\text{u}}}) +\sqrt{n} ({\bm \Theta}_j^\top \hat{\bm\Sigma}-\bm e_j^\top) ( \bbeta - \hat{\bbeta^{\text{u}}})
\end{align}
For the first term in (\ref{eq:eq14}), we could further decompose this term from $\hat{\bm \Sigma} =\frac{1}{n}\sum_{i=1}^n \bm x_i \bm x_i^\top$:
\begin{align} \label{eq:eq15}
     \sqrt{n} (\hat{\bm \Theta}_j^\top \hat{\bm\Sigma} -{\bm \Theta}_j^\top \hat{\bm\Sigma}) ( \bbeta - \hat{\bbeta^{\text{u}}}) 
     &={\sqrt{n}} (\hat{\bm \Theta}_j^\top-{\bm \Theta}_j^\top) \hat{\Sigma} (\bbeta - \hat{\bbeta^{\text{u}}}) \notag\\
     &\le {\sqrt{n}} \lambda_{s}(\hat{\Sigma}) \|\hat{\bm \Theta}_j-{\bm \Theta}_j\|_2 |\bbeta - \hat{\bbeta^{\text{u}}}\|_2 \notag\\
     &\le o(1) + \frac{\gamma \mu_s^2}{\nu_s^2} \frac{s_1^2 \log^2 d \log (1/\delta)\log^3 mn}{m^2 n^{3/2} \epsilon^2} + \frac{\gamma \mu_s^2}{\nu_s^2} \frac{s_0^2 \log^2 d \log (1/\delta)\log^3 n}{n^{3/2} \epsilon^2} 
\end{align}
Thus, for the second term of (\ref{eq:eq14}), by Lemma~\ref{lemma1}, we have:
\begin{align}\label{eq:eq16}
  \sqrt{n} ({\bm \Theta}_j^\top \hat{\bm\Sigma}-\bm e_j^\top) ( \bbeta - \hat{\bbeta^{\text{u}}}) &\le \sqrt{n} \|{\bm \Theta}_j^\top \hat{\bm\Sigma}-\bm e_j^\top\|_{\infty} \|\bbeta - \hat{\bbeta^{\text{u}}}\|_1 \notag \\
  &\le  k \sqrt{n} \sqrt{\frac{\log d}{mn}}\sqrt{s} \|\bbeta - \hat{\bbeta^{\text{u}}}\|_2 \notag \\
  &\le k \cdot\sqrt{n}\sqrt{ \frac{s \log d}{mn} } \cdot \|\bbeta - \hat{\bbeta^{\text{u}}}\|_2\notag\\
  &\le o(1) + \frac{\gamma \mu_s^2}{\nu_s^2} \frac{s_1^2 \log^2 d \log (1/\delta) \log^3 mn}{m^2 n^{3/2} \epsilon^2} + \frac{\gamma \mu_s^2}{\nu_s^2} \frac{s_0^2 \log^2 d \log (1/\delta)\log^3 n}{n^{3/2} \epsilon^2} 
\end{align}
Then, combining (\ref{eq:eq15}) and (\ref{eq:eq16}), we have that:
\begin{align*}
    \sqrt{n} (\hat{\bm \Theta}_j^\top \hat{\bm\Sigma}-\bm e_j^\top) ( \bbeta - \hat{\bbeta^{\text{u}}})  \le \frac{2\gamma \mu_s^2}{\nu_s^2} \frac{s_1^2 \log^2 d \log (1/\delta) \log^3 mn}{m^2 n^{3/2} \epsilon^2} + \frac{2\gamma \mu_s^2}{\nu_s^2} \frac{s_0^2 \log^2 d \log (1/\delta) \log^3 n}{n^{3/2} \epsilon^2} 
\end{align*}
Then, insert the result into (\ref{eq:eq1}), we have:
\begin{align}\label{eq:eq17}
    \sqrt{n} \qty(\hat{\beta_j} - \beta_j -  \frac{2\gamma \mu_s^2}{\nu_s^2} \frac{s_1^2 \log^2 d \log (1/\delta) \log^3 mn}{m^2 n^2 \epsilon^2} - \frac{2\gamma \mu_s^2}{\nu_s^2} \frac{s_0^2 \log^2 d \log (1/\delta)\log^3 n }{n^2 \epsilon^2} ) =  \frac{1}{\sqrt{n}} \hat{\bm \Theta}_j^\top \bm X^\top \bm W +\sqrt{n}  E_3
\end{align}
From the result in (\ref{eq:eq17}), notice that the right hand side asymptotically follows a distribution of $N(0, \sigma^2 \hat{\bm \Theta}_j^\top \hat{\bm \Sigma} \hat{\bm \Theta}_j+ \frac{8\Delta_1^2 \log(1/\delta)}{n\epsilon^2})$. Also, by the concentration of Gaussian distribution, we have with high probability, $E_2 \le \frac{8\Delta_1^2 \log(1/\delta)}{n\epsilon^2})$. Thus, we have:
\begin{equation*}
     \sqrt{n} \qty(\hat{\beta_j} - \beta_j - \frac{2\gamma \mu_s^2}{\nu_s^2} \frac{s_1^2 \log^2 d \log (1/\delta)}{m^2 n^2 \epsilon^2} - \frac{2\gamma \mu_s^2}{\nu_s^2} \frac{s_0^2 \log^2 d \log (1/\delta)}{n^2 \epsilon^2} ) / \sqrt{\sigma^2 \hat{\bm \Theta}_j^\top \hat{\bm \Sigma} \hat{\bm \Theta}_j+ \frac{8\Delta_1^2 \log(1/\delta)}{n\epsilon^2}}\sim N(0,1)
\end{equation*}
Then, we could replace $\mu_s,\nu_s$ with the estimation $\hat{\mu}_s, \hat{\nu}_s$ introduced in Algorithm~\ref{algo:privateeigen}, the constant could be scaled to one given the tail bound of Laplace random variable. Also, for the estimation of $\sigma$, according to the assumption, we have $\hat{\sigma} = \sigma + o(1)$. For simplicity, we denote $a =\frac{2k \hat{\mu}_s^2}{\hat{\nu}_s^2} \frac{s_1^2 \log^2 d \log (1/\delta)\log^3 mn}{m^2 n^2 \epsilon^2} + \frac{2k \hat{\mu}_s^2}{\hat{\nu}_s^2} \frac{s_0^2 \log^2 d \log (1/\delta) \log^3 n }{n^2 \epsilon^2}$, the confidence is given by:	
\begin{align*}
   	&J_j(\alpha) = \\&[\hat{\beta_j} - a - \frac{\sigma\Phi^{-1}(1-\alpha/2) }{\sqrt{n}}\sqrt{\hat{\bm \Theta}_j^\top \hat{\bm \Sigma} \hat{\bm \Theta}_j+ \frac{8\Delta_1^2 \log(1/\delta)}{n\epsilon^2}} , \hat{\beta_j} + a+ \frac{\sigma\Phi^{-1}(1-\alpha/2) }{\sqrt{n}}\sqrt{\hat{\bm \Theta}_j^\top \hat{\bm \Sigma} \hat{\bm \Theta}_j+ \frac{8\Delta_1^2 \log(1/\delta)}{n\epsilon^2}}  ] 
\end{align*}
\subsection{Proof of Theorem~\ref{thm: bootstrap_simu_single}}
In this proof, we first need to show that our algorithm is $(\epsilon, \delta)$ private. We assume in two data sets, the adjacent data set is different in $(\bm x_{ij}, y_{ij})$ and $(\bm x'_{ij}, y'_{ij})$. Then, we have:
\begin{align*}
    \|\frac{1}{\sqrt{n}}\hat{\bm \Theta}\bm x_{ij} e_j - \frac{1}{\sqrt{n}}\hat{\bm \Theta} \bm x_{ij}' e_j\|_{\infty} &\le \frac{2}{\sqrt{n}}\|\hat{\bm \Theta}\bm x_{ij} e_j \|_{\infty}\\
    &\le \frac{2}{\sqrt{n}}\|\hat{\bm \Theta}\bm x_{ij}\|_{\infty} \|e_j\|_{\infty}\\
    &\le  \frac{2\sqrt{\log n }}{\sqrt{n}} \|\hat{\bm \Theta}\|_1 \|\bm x_{ij}\|_{\infty}\\
    &\le 2\sqrt{\frac{\log n}{n}} c_x \sqrt{s} c_1
\end{align*}
According to the choice of $B_5$, the privacy could be guaranteed. Then, let us start the proof of consistency. In this proof specifically, we define
$U' = \max_{k \in G} \hat{\bm \Theta}_k^T \frac{1}{\sqrt{n}} \sum_{i=1}^n  \bm x_{ij} e_j$ and also, we define:
\begin{equation*}
    M_0 = \max_{k \in G} \frac{1}{\sqrt{n}} \sum_{j=1}^n \xi_{jk},
\end{equation*}
where $\xi_j$ follows a Gaussian Distribution $N(0, \Theta_k^\top \Sigma \Theta_k\sigma^2)$. Also, we define the $\alpha$-quantile of $M_0$ as $U_{M_0}(\alpha)$. Then, we could start the proof.

We are aiming at proving $ \sup_{\alpha \in (0,1)} |\Pro(M \le C_U(\alpha)) - \alpha | = o(1)$. First, we could prove that $C_U(\alpha)$ and $C_{U'}$ are close to each other for any $\alpha \in (0,1)$. From the design of private max algorithm, from Lemma 3.4 in \citep{cai2019cost}, suppose $l_1$ is the element chosen from $U'$ and $l_2$ is from $U$ without noise injection, we use $w$ to represent the noise injected when we pick the largest value privately, we find that, for any $c>0$:
\begin{equation*}
    l_2^2 \le l_1^2 \le (1+c) l_2^2 + 4(1+1/c) \|w\|_{\infty}^2
\end{equation*}
From Lemma A.1 in \citep{cai2019cost}, we can verify that there exists constant $k_0, k_1$ such that $\|w\|_{\infty}^2 \le k_0 \cdot \frac{s^* \log^4 d \log n }{n}$. When we choose $c = o(1)$, e.g, $c = k_1 \frac{s^* \log d}{n}$, then from the conditions, we could claim that $l_1 = l_2 + o(1)$, also notice that the scale of noise we injected is small, it is easy to verify that $U = U' + o(1)$. The following discussions will be between $U'$ and $U_{T_0}$. 

Motivated by the proof of Theorem 3.1 and Theorem 3.2 in \citep{chernozhukov2013gaussian}, our proof will be divided into two major parts, to measure the closeness between $U'$ and $U_{M_0}$ and measure the closeness between $T$ and $T_0$. We start from the measurement between $M$ and $M_0$. From the definition that $M(\hat{\bbeta}^{(i)}) = \max_{k \in G} \sqrt{n} (\hat{\bbeta}_k^{(i)} - \bbeta_k^{(i)}) $, we notice that for each $k$ in $1,2,\dots,d$, we have:
\begin{align*}
    \sqrt{n} (\hat{\bbeta}^{(i)} - \bbeta^{(i)}) = \frac{1}{\sqrt{n}} \hat{\Theta} X_i^\top \bm W_i + (\hat{\Theta}\Sigma - I) (\hat{\bbeta}-\bbeta) +\sqrt{n} E
\end{align*}
Then, 
\begin{align}\label{eq:eq35}
    |M-M_0|\le  (\|\frac{1}{\sqrt{n}} \hat{\Theta} X_i^\top \bm W_i\|_{\infty} - \|\frac{1}{\sqrt{n}} \sum_{j=1}^n \xi_j\|_{\infty}) + \|(\hat{\Theta}\Sigma - I) (\hat{\bbeta}-\bbeta) +\sqrt{n} E\|_{\infty}
\end{align}
We analyze the two parts in (\ref{eq:eq35}) separately. For the first term in (\ref{eq:eq35}). First, from Lemma 1.1 in \citep{zhang2017simultaneous}, we could obtain the result that for any $z$, $\sup_{z} |P(\|\frac{1}{\sqrt{n}}\bm \Theta \bm X_i^\top \bm W_i\|_{\infty}\le z) - P(M_0 \le z)| \le c_0 \cdot \frac{1}{n^{c_1}}$, where $c_0$ and $c_1$ are constants. Also, 
\begin{align*}
    \|\frac{1}{\sqrt{n}}  \hat{\bm \Theta} \bm X_i^\top \bm W_i\|_{\infty} - \|\frac{1}{\sqrt{n}}\bm \Theta \bm X_i^\top \bm W_i\|_{\infty}
    &\le \frac{1}{\sqrt{n}}\| \hat{\bm \Theta} \bm X_i^\top \bm W_i - \bm \Theta \bm X_i^\top \bm W_i\|_{\infty}\\
    &\le \|\bm\hat{\Theta}-\bm \Theta\|_1 \cdot \|\frac{1}{\sqrt{n}}\bm X_i^\top \bm W_i\|_{\infty}\\
    &\le c\cdot s^* \sqrt{\frac{\log d}{n}} \sqrt{\log d} \le c\cdot s^* \log d \sqrt{\frac{1}{n}} = o(1)
\end{align*}
On the other hand, we also notice that for the second term in (\ref{eq:eq35}), following the proof in Theorem~\ref{thm: confidence interval}, we also know that the second part is $o(1)$, hence finishes the first part of the proof. In the second part of the proof. By the arguments in the proof of Theorem 3.2 in \citep{chernozhukov2013gaussian}, we have for any $v$:
\begin{align*}
    \sup_{\alpha \in (0,1)} |\Pro(T \le C_{U'}(\alpha)) - \alpha | \le c_0\frac{1}{n^{c_1}} +c_2 v^{1/3}(1 \vee \log(d/v))^{2/3}+P(\square > v),
\end{align*}
where $\square = \max_{k,l} \bm \hat{\Theta}_k^\top \bm \hat{\Sigma} \bm \hat{\bm \Theta}_l - \bm {\Theta}_k^\top \bm \Sigma \bm {\bm \Theta}_l$.Then, we have:
\begin{align*}
     &\| \hat{\bm \Theta}^\top  \hat{\bm \Sigma} \bm \hat{\bm \Theta} - \bm {\Theta}^\top \bm \Sigma \bm {\bm \Theta}\|_{\max}\\
    &\le \|\hat{\bm \Theta}^\top  \hat{\bm \Sigma} \bm \hat{\bm \Theta} - \hat{\bm \Theta}^\top  {\bm \Sigma} \bm \hat{\bm \Theta}\|_{\max} + \|\hat{\bm \Theta}\Sigma \hat{\bm \Theta} -  \bm \Theta\|_{\max} \\
    &\le \|\hat{\bm \Theta} \|_{\infty} \|\hat{\bm \Sigma}-{\bm \Sigma}\|_{\max}\|\hat{\bm \Theta} \|_1 +\|(\hat{\bm \Theta}-\bm \Theta + \bm \Theta)\Sigma (\hat{\bm \Theta}-\bm \Theta + \bm \Theta)-  \bm \Theta\|_{\max} \notag\\
    &\le L^2 \sqrt{\frac{\log d}{n_0}} + \|\hat{\bm \Theta}-\bm \Theta\|_1^2 \|\bm \Sigma\|_{\max} + 2\|\hat{\bm \Theta}-\bm \Theta\|_1\\
    &\le k_0 \frac{{s^*}^2 \log d}{n_0} + k_1 s^*\sqrt{\frac{ \log d}{n_0}},
\end{align*}
where $k_0, k_1$ are constants. Then, with a proper choice of $v$, we could claim that $ \sup_{\alpha \in (0,1)} |\Pro(T \le C_{U'}(\alpha)) - \alpha | = o(1)$, which finishes the proof.

\section{Appendix}
In this section, we will give proofs of the corollary in the main proof. We will introduce them one by one:
\subsection{Proof of corollary~\ref{corollary:1} }\label{proof:col1}
Following the proof of Theorem 1 in \citep{acharya2020general}, we can gain the upper bound of $\sum_{i=1}^k d_{TV}({\bm p_{+i}^{Z^m}},{\bm p_{-i}^{Z^m}})$, when we consider the central DP:
\begin{align*}
  \frac{1}{k} \qty(\sum_{i=1}^k d_{TV}({\bm p_{+i}^{Z^m}},{\bm p_{-i}^{Z^m}}))^2
  &\leq 7 \sum_{t=1}^m \E_{A}\qty[ \sum_{i=1}^k \int_{\mathcal{Z}} \frac{ (\E_{\bm p_{a}^{\otimes n}}[\mathcal{W}{(z \mid X)} ] - \E_{\bm p_{a^{\oplus i}}^{\otimes n}}[{\mathcal{W}(z \mid X)})]^2 }{\E_{\bm p_{a}}[{\mathcal{W}(z\mid X)}] } d{\mu}]
\end{align*}
Also notice that: 
\begin{align*}
\E_{\bm p_{a^{\oplus i}}^{\otimes n}}[{\mathcal{W}(z \mid X)})] = \E_{\bm p_{a}^{\otimes n}}\qty[\frac{d \bm p_{a^{\oplus i}}^{\otimes n}}{d \bm p_{a}^{\otimes n}}{\mathcal{W}(z \mid X)})] =\E_{\bm p_{a}}(1+q_{a,i}\phi_{a,i}(X))^n \cdot \mathcal{W}(z\mid X)
\end{align*}
The last equation is from the definition of condition~\ref{Cond:1}. Also, by the inequality that we can find constants $c_0$, $c_1$ that when $x>0$ and $x \asymp 1/n$, $(1+x)^n \le 1+ c_0 \cdot nx$ and $(1-x)^n \le 1- c_0 \cdot nx$. If we have $|q_{a,i}\phi_{a,i}(X)| \asymp 1/n$. we could find a constant $c_2$, such that: 
\begin{align*}
  \frac{1}{k}(\sum_{i=1}^k d_{TV}\qty({\bm p_{+i}^{Z^m}},{\bm p_{-i}^{Z^m}}))^2 \leq c_2 q^2 n^2 \sum_{t=1}^m \E_{A}\qty[ \sum_{i=1}^k \int_{\mathcal{Z}} \frac{ \E_{\bm p_{A}^{\otimes n}}{[\phi_{a,i}(X)\mathcal{W}(z \mid X)}]^2 }{\E_{\bm p_{A}^{\otimes n}}[{\mathcal{W}(z \mid X)}] } d{\mu} ]
\end{align*}
which finishes the proof of corollary~\ref{corollary:1}.
\subsection{Proof of Corollary~\ref{corollary:2}}\label{proof:col2}
We continue to show the proof of Corollary~\ref{corollary:2}. First, from Theorem 2 in \citep{acharya2020general}, we get a direct result when condition~\ref{Cond:2} is satisfied, given all the conditions in corollary~\ref{corollary:2} hold, we have:
\begin{align*}
\qty(\frac{1}{k}\sum_{i=1}^k d_{TV}(\bm p_{+i}^{Z^m}, \bm p_{-i}^{Z^m}))^2
\le \frac{7}{k} q^2 n^2 \sum_{t=1}^m \max_{a\in\mathcal{A}}\int_{\mathcal{Z}} \frac{\Var_{\bm p_{a}}[\mathcal{W}(z\mid X)]}{\E_{\bm p_{a}}[{\mathcal{W}(z\mid X)}]} d {\mu}
\end{align*}
Then, the focus of the proof of this corollary is on the calculation of $\int_{\mathcal{Z}} \frac{\Var_{\bm p_{a}}[\mathcal{W}(z\mid X)]}{\E_{\bm p_{a}}[{\mathcal{W}(z\mid X)}]} d {\mu}$ when the channel $\mathcal{W}$ is a privacy constraint channel $\mathcal{W}^{priv}$. For simplicity, we denote $L(\bm z, \bm X) = log \mathcal{W}^{priv}(\bm z| \bm X)$, where $\bm z \in \mathbb{R}^d$ and $\bm X \in \mathbb{R}^{n \times d}$. Then, notice that $\mathcal{W}^{priv}$ is $\epsilon$-differentially private constraint, for two adjacent dataset $\bm X$ and $\bm X'$, we have:
\[|L(\bm z, \bm X) - L(\bm z, \bm X')| \le \epsilon\]
By McDiarmid’s inequality, we could claim that $L$ is $\sqrt{n}\epsilon$- subGaussian. So we could find a constant $c$, which satisfies that:
\[\E [e^{2L}] \le c \cdot e^{2\E[L]} \cdot e^{2n \epsilon^2}\]
Then, by Jensen inequality, we have:
\[\E [e^{2L}] \le c \cdot (e^{E[L]})^2 \cdot e^{2n \epsilon^2}\]
Thus, we have:
\[\frac{\Var [ \mathcal{W}^{priv}(\bm z| \bm X)]}{\E[\mathcal{W}^{priv}(\bm z| \bm X)]^2} = \frac{\E[\mathcal{W}^{priv}(\bm z| \bm X)^2]}{(\E[\mathcal{W}^{priv}(\bm z| \bm X)])^2}-1 = \frac{\E[e^{2L}]}{(\E[e^L])^2} -1 \le e^{2n\epsilon^2}-1\]
Thus, we have:
\begin{align*}
\qty(\frac{1}{k}\sum_{i=1}^k d_{TV}(\bm p_{+i}^{Z^m}, \bm p_{-i}^{Z^m}))^2
\le &\frac{7}{k} \alpha^2 n^2 \sum_{t=1}^m \max_{a\in\mathcal{A}}\int_{\mathcal{Z}} \frac{\Var_{\bm p_{a}}[\mathcal{W}(z\mid X)]}{\E_{\bm p_{a}}[{\mathcal{W}(z\mid X)}]^2} \cdot \E_{\bm p_{a}}[{W(z\mid X)}]d {\mu}   \\
\le & \frac{7}{k} \alpha^2 n^2 ( e^{2n\epsilon^2}-1)\sum_{t=1}^m \max_{a\in\mathcal{A}}\int_{\mathcal{Z}} \E_{\bm p_{a}}[{\mathcal{W}(z\mid X)}]d {\mu}  \\
\le &\frac{7}{k} q^2 m n^2 ( e^{2n\epsilon^2}-1)
\end{align*}
which finishes the proof of corollary~\ref{corollary:2}. 
\subsection{Proof of corollary~\ref{corollary:3}}\label{proof:col3}
For an $(n, \rho)$-estimator $\hat{\theta}$ of the true parameter under $\ell_p$ loss, we define $\hat{A}$ for $A$ as 
\[
	\hat{A} = \underset{a\in\mathcal{A}}{\arg\!\min} 
	\|\theta_a-\hat{\theta}\|_p.
	\]
Then, by the triangle inequality, we have
	\[
	\norm{\theta_{A}-\theta_{\hat{A}}}_p \leq \norm{
		\theta_{A}-\hat{\theta}}_p+
	\norm{\theta_{\hat{A}}-\hat{\theta}}_p
	\leq 2\norm{\hat{\theta}-\theta_A}_p.
	\]

Because $\hat{\theta}$ is an $(n,\rho)$-estimator under $\ell_p$ loss, we have,
\begin{align}
\E_Z[\E_{\bm p_Z}[\norm{\theta_{Z}-\theta_{\hat{Z}}}_p^p]]
& \leq 2^p {\rho^p} \Pro[\bm p_Z \in \mathcal{P}_\Theta] + \max_{z \neq z'}  
\norm{\theta_{z}-\theta_{z'}}_p^p \Pro[\bm p_Z \notin \mathcal{P}_\Theta] 
\\
& \le  2^p{\rho^p} + 4^p{\rho^p} \frac{1}{\tau} \cdot \frac{\tau}{4} \label{eqn:total-error}\\
& \le \frac{3}{4} {4^p\epsilon^p} \label{eq:eq9},
\end{align}
using the fact that $\Pro[{\bm p_A \in \mathcal{P}_\Theta}] \ge 1 -  \tau/4$ and condition~\ref{Cond:4}. Also, from  condition~\ref{Cond:4}, Next, combining condition~\ref{Cond:4} and \ref{eq:eq9}, we could have: $\frac{1}{\tau k}\sum_{i=1}^k \Pro [A_i\neq \hat{A}_i] \leq \frac{3}{4}$. Also, since the Markov relation $A_i- X^m - Z^m -\hat{A}_i$ holds for all $i$, by the standard relation between total variation distance and hypothesis testing, and also the definition of $\tau$ to be less than $1/2$, we have:
\begin{align*}
\Pro[{A_i\neq \hat{A}_i} ]
&\geq \tau \Pro[{ \hat{A}_i = -1 }|{A_i = 1}] + (1-\tau) 
\Pro[{ \hat{A}_i = 1 }|{A_i = -1}] \\
&\geq \tau (\Pro[{ \hat{A}_i = -1 }|{A_i = 1} ]+ 
\Pro[{ \hat{A}_i = 1 }|{A_i = -1} ] )\\
&\geq \tau (1- d_{TV}({\bm p_{+i}^{X^m}},{\bm p_{-i}^{X^m}})) \\
&\ge \tau (1- 1/n \cdot d_{TV}({\bm p_{+i}^{Z^m}},{\bm p_{-i}^{Z^m}}) )
\end{align*}
The last inequality uses the definition of total variation, because $Z^m$ is generated by $X^m$ from the privacy constraint channel $W^{priv}$, so for each dataset $X_i$, $i = 1,2,\dots,m$ on the $i$-th machine, let $X_{ijk}$ be the dataset which changes the order of $X_{ij}$ and $X_{ik}$, then for any $z\in\mathcal{Z}$,$W^{priv}(z|X_i) = W^{priv}(z|X_{ijk})$. Thus, by the definition of total variation, we could verify that $d_{TV}({\bm p_{+i}^{X_i}},{\bm p_{-i}^{X_i}}) = 1/n \cdot d_{TV}({\bm p_{+i}^{Z_i}},{\bm p_{-i}^{Z_i}})$. 
Summing over $1\leq i\leq k$ and combining it with the previous bound, we obtain
\[
	\frac{3}{4} \geq \frac{1}{\tau k}\sum_{i=1}^k 
	\Pro[{A_i\neq \hat{A}_i}] \geq 1 - 
	\frac{1}{n k}\sum_{i=1}^k d_{TV}(\bm p_{+i}^{Z^n},\bm p_{-i}^{Z^n})
\]
which finishes the proof of corollary~\ref{corollary:3}. 
\subsection{Proof of Lemma~\ref{lemma:2}}
Proof of Lemma~\ref{lemma:2}:
First, we would like to show that our algorithm is $(\epsilon,\delta)$-differentially private. For two adjacent data sets, we have:
\begin{align*}
    &\frac{1}{mn}|(\pi_R(y_i) - x_i^T \hat{\bbeta})^2 - (\pi_R({y_i'}) - {x_i'}^T \hat{\bbeta})^2|\\
    &\le \frac{2}{mn} (\pi_R(y_i) - x_i^T \hat{\bbeta})^2\\
    &\le \frac{4}{mn} (\pi_R(y_i)^2 + (x_i^T \hat{\bbeta})^2 ) \le \frac{4}{mn}(R^2 + s c_0^2 c_x^2)
\end{align*}
From the definition of Gaussian Mechanism, we could claim that our algorithm is $(\epsilon, \delta)$-differential private. Then, for the convergence rate of our estimated $\sigma$, from our algorithm, first, we observe with the choice of $R$, we claim that with high prob, we have $\pi_R(Y) = Y$. Therefore, we have:
\begin{align*}
    |\sigma^2 - \hat{\sigma}^2| &\le |\frac{1}{mn} \|\bm X \bbeta + \bm W - \bm X\hat{\bbeta}\|_2^2 -\sigma^2|+ |E |\\
    &\le |\frac{1}{mn} \bm W^T \bm W -\sigma^2|+ (\bbeta -\hat{\bbeta})\hat{\Sigma} (\bbeta - \hat{\bbeta}) + \frac{1}{mn} (\bbeta - \hat{\bbeta}) \bm X^T \bm W + |E|
\end{align*}
For the first term, we could obtain that $|\frac{1}{mn} \bm W^T \bm W -\sigma^2| = O(\frac{1}{\sqrt{mn}})$
Also, we have:
\begin{align*}
    (\bbeta -\hat{\bbeta})\hat{\Sigma} (\bbeta - \hat{\bbeta}) &\le \lambda_s(\Sigma) \|\bbeta - \hat{\bbeta}\|_2^2\\
    &\le c L \qty(\frac{s\log d}{mn} + \frac{s^2 \log^2 d \log(1/\delta)\log^3 mn}{m^2n^2\epsilon^2})
\end{align*}
Then, from Bernstein inequality, we could obtain that:
\begin{align*}
    \|\frac{1}{mn}\bm X^T \bm W\|_{\infty} = c_1 \cdot \sqrt{\frac{\log d}{mn}}
\end{align*}
Therefore, we claim that:
\begin{align*}
    &\frac{1}{mn} (\bbeta - \hat{\bbeta}) \bm X^T \bm W \\ 
    &\le \frac{1}{mn} \|\bbeta - \hat{\bbeta}\|_1 \|\bm X^T \bm W \|_{\infty} \\
    &\le c_2  \frac{\sqrt{s}}{mn} \|\bbeta - \hat{\bbeta}\|_2 \|\bm X^T \bm W \|_{\infty} \\
    &\le c_3  \sqrt{s} \qty(\sqrt{\frac{s\log d}{mn}} + \frac{s \log d \sqrt{\log(1/\delta)}\log^{3/2} mn}{mn\epsilon}) \sqrt{\frac{\log d}{mn}}\\
    &= O_p \qty(\frac{s\log d}{mn} + \frac{s \log d \sqrt{\log(1/\delta)}\log^{3/2} mn}{mn\epsilon} \cdot \sqrt{\frac{s \log d}{mn}})\\
    & = O_p\qty(\frac{s\log d}{mn} + \frac{s^2 \log^2 d \log(1/\delta)\log^3 mn}{m^2n^2\epsilon^2})
\end{align*}
Also, from our algorithm, we have $E \sim N(0, 2B_2^2 \log(1.25/\delta)/\epsilon^2)$. Then, 
\begin{align*}
    |E| = \frac{2B_2^2\log(1/\delta)}{\epsilon^2}|N(0,1)| = c_4 \frac{R^4 + s^2 c_0^4 c_x^4 \log(1/\delta)}{m^2n^2\epsilon^2} = c_5 \cdot \frac{s^2 \log^2 d \log(1/\delta)\log^3 mn}{m^2n^2\epsilon^2}, 
\end{align*}
by observing that $c_x = O(\sqrt{\log d})$. Combining above inequalities, we have reached our conclusion. Therefore, we finish our proof. 
\subsection{Proof of Lemma~\ref{lemma:1}}
Proof of Lemma~\ref{lemma:1}:
It is not difficult to verify the privacy conditions. Then, from the theory of covering number, we could find $n_1$ vectors $\bm v_1, \bm v_2, \dots, \bm v_{n_1}$, such that for each s-sparse unit vector $\bm v$, we have $\|\bm v - \bm v_i\| \le 1/9$. Thus, we have:
\begin{align*}
    {\lambda}_s - \hat{\lambda}_s &= \bm v^* \hat{\bm \Sigma} \bm v^* - \bm v_i \hat{\bm \Sigma} \bm v_i \\
    &\le \bm v^* \hat{\bm \Sigma} (\bm v^*- \bm v_i) + (\bm v^* - \bm v_i) \hat{\bm \Sigma} \bm v_i \\
    &\le \frac{2}{9} \lambda_{2s} \quad \\
    &\le 8/9 \lambda_s
\end{align*}
Note that the second last inequality is a direct result of Cauchy inequality and the last inequality holds because let $\bm v^{*'}$ be the corresponding eigenvector of $\lambda_{2s}$, then we could break this eigenvector to two $s$-sparse vectors $\bm v_1'$ and $\bm v_2'$ such that $\bm v^{*'} = \bm v_1' +\bm v_2'$, then $\lambda_{2s} = (\bm v_1' +\bm v_2')^T \hat{\bm \Sigma} (\bm v_1' +\bm v_2') \le 4 \lambda_s$. \\
Also, notice that for the noise $\xi$, by the concentration of Laplace distribution, we could find a constant $c$ such that $\xi \le c s\log d/\sqrt{n} = o(1)$ with high probability. By the definition of $\lambda_s$, we conclude the proof. \hfill $\square$



\end{document}